\documentclass[a4paper,fleqn]{cas-dc}
\usepackage[numbers]{natbib}
\usepackage{amsmath,amssymb,amsfonts}
\usepackage{cases}
\usepackage{color,xcolor}
\usepackage{graphicx}
\usepackage{subfigure}
\usepackage{algorithm}
\usepackage{algorithmic}
\usepackage{epstopdf}
\usepackage{multirow}
\usepackage{rotating}
\usepackage{pifont}
\usepackage{bbding}
\usepackage{fontawesome}
\usepackage{makecell}
\usepackage{booktabs}
\usepackage{hyperref}
\hypersetup{
    colorlinks=true,
    linkcolor=blue,
    filecolor=blue,
    urlcolor=blue,
    citecolor=blue,
}

\graphicspath{{fig/}}

\newcommand{\red}[1]{\textcolor{red}{\textbf{#1}}}
\newcommand{\blue}[1]{\textcolor{blue}{\textbf{#1}}}

\def\tsc#1{\csdef{#1}{\textsc{\lowercase{#1}}\xspace}}
\tsc{WGM}
\tsc{QE}

\begin{document}
\let\WriteBookmarks\relax
\def\floatpagepagefraction{1}
\def\textpagefraction{.001}
\let\printorcid\relax

\shortauthors{J. Liu.et al.}

\title[mode = title]{Robust Orthogonal NMF with Label Propagation for Image Clustering}

\author[1,2]{Jingjing Liu}

\author[1]{Nian Wu}

\author[3]{Xianchao Xiu}
\ead{xcxiu@shu.edu.cn}
\cormark[1]

\author[1]{Jianhua Zhang}


\address[1]{Shanghai Key Laboratory of Chips and Systems for Intelligent Connected Vehicle, School of Microelectronics, Shanghai University, Shanghai, 200444, China}
\address[2]{State Key Laboratory of Integrated Chips and Systems, Fudan University, Shanghai, 201203, China}
\address[3]{School of Mechatronic Engineering and Automation, Shanghai University, Shanghai 200444, China}

\cortext[1]{Corresponding author}

\begin{abstract}
Non-negative matrix factorization (NMF) is a popular unsupervised learning approach widely used in image clustering. However, in real-world clustering scenarios, most existing NMF methods are highly sensitive to noise corruption and are unable to effectively leverage limited supervised information. To overcome these drawbacks, we propose a unified non-convex framework with label propagation called robust orthogonal nonnegative matrix factorization (RONMF).
This method not only considers the graph Laplacian and label propagation as regularization terms but also introduces a more effective non-convex structure to measure the reconstruction error and imposes orthogonal constraints on the basis matrix to reduce the noise corruption, thereby achieving higher robustness.
To solve RONMF, we develop an alternating direction method of multipliers (ADMM)-based optimization algorithm. In particular, all subproblems have closed-form solutions, which ensures its efficiency. Experimental evaluations on eight public image datasets demonstrate that the proposed RONMF outperforms state-of-the-art NMF methods across various standard metrics and shows excellent robustness.
The code will be available at \href{https://github.com/slinda-liu}{https://github.com/slinda-liu}.
\end{abstract}


\maketitle

\shorttitle{<Robust Orthogonal NMF with Label Propagation for Image Clustering>}

\section{Introduction}
Image clustering plays an important role in many areas of image and video processing. However, high-dimensional data often contains a large number of redundant features and noise, which makes the clustering performance unreliable. Therefore, image dimensionality reduction has become a key step in clustering. To this end, various dimensionality reduction techniques have been developed, such as principal component analysis (PCA) \cite{greenacre2022principal,li2021sparse}, canonical correlation analysis (CCA) \cite{yang2019survey,xiu2022efficient}, and non-negative matrix factorization (NMF) \cite{li2025one,jia2024semi}. It is worth noting that compared with PCA and CCA, NMF not only retains the inter-class structure of the original data but also considers the non-negativity of the actual problem \cite{lee1999learning}. At present, NMF has been extensively applied in multi-view clustering \cite{dou2025learning}, biomedical science \cite{kriebel2022uinmf}, hyperspectral unmixing \cite{li2025deep}, community detection \cite{bai2024dual}. Interested readers can refer to \cite{gillis2020nonnegative}.

During the past few decades, numerous NMF variants have been proposed. For example, to deal with data distribution,
Wang \textit{et al.} \cite{wang2006label} introduced a novel graph-based transductive classification technique \cite{xia2021graph} to NMF termed linear neighborhood propagation (LNP). Subsequently, Cai \textit{et al.} \cite{cai2010graph} proposed graph Laplacian regularized NMF (GNMF) by representing geometric data through the nearest neighbor graph. In addition, Guan \textit{et al.} \cite{guan2017truncated} proposed a truncated Cauchy NMF framework and solved the non-quadratic objective through semi-quadratic programming. Leng \textit{et al.} \cite{leng2019graph} developed the graph Laplacian regularized smooth NMF (GSNMF), which combined $\ell_p$-norm  smoothing constraints with graph regularization. Note that here $1<p<2$, and it is obviously a convex function between $\ell_1$-norm and $\ell_2$-norm. Recently, Wang \textit{et al.} \cite{wang2020robust} considered a robust random graph regularized matrix factorization method to achieve more effective data clustering.

In practical scenarios, the high costs associated with sample labeling for high-dimensional data present a significant challenge. To address this issue, one can utilize a small portion of labeled data to improve image clustering performance \cite{peng2023multiview}. Liu \textit{et al.} \cite{liu2011constrained} proposed to enhance discrimination by incorporating label information into NMF, leading to the development of constrained NMF (CNMF). Li \textit{et al.} \cite{li2016class} introduced class-driven concept decomposition (CDCF), which incorporates class-driven constraints to link the class labels of data points with their feature representations. Building on this idea, Jia \textit{et al.} \cite{jia2019semi} proposed integrating complementary regularizers into the traditional NMF framework, resulting in a method called SNMFDSR. Later, Lan \textit{et al.} \cite{lan2020label} combined semi-supervised learning with NMF to create an image clustering method that leverages label propagation. Extending these advancements, Liu \textit{et al.} \cite{liu2023constrained} introduced label propagation-constrained NMF (LpCNMF), an efficient variant that incorporates label information from both labeled and unlabeled data. This method also uses the intrinsic geometric structure of the data as a regularization term, effectively constraining the model.



Note that the aforementioned methods typically use the Frobenius norm to measure reconstruction error, which can be sensitive to noise.
To solve the above problems, Kong \textit{et al.} \cite{kong2011robust} adopted $\ell_{2,1}$-norm \cite{nie2010efficient} to measure the reconstruction error to enhance the robustness. Similar ideas are also considered in \cite{wu2018manifold,huang2020robust}. Furthermore, Li \textit{et al.} \cite{li2017robust} introduced robust structured NMF (RSNMF) to explore block diagonal structures by replacing $\ell_{2,1}$-norm with $\ell_{2,p}$-norm $(0<p<1)$.
In fact, compared with convex functions, non-convex functions can achieve better sparsity, thereby achieving better feature extraction and data dimensionality reduction.  Thanks to the rapid development of sparse optimization \cite{lu2015nonconvex,zhang2022structured,sun2023learning}, one can employ various non-convex functions such as capped-$\ell_1$ \cite{zhang2010analysis}, minimax concave penalty (MCP) \cite{zhang2010nearly}, smoothly clipped absolute deviation (SCAD) \cite{fan2001variable}, and exponential-type penalty (ETP) \cite{gao2011feasible}. Unfortunately, we find that the structured non-convex sparsity is rarely considered to measure the reconstruction error in NMF. \textit{This observation raises the question: can we leverage the strengths of the graph regularization, label propagation, and these non-convex functions to enhance the robustness and effectiveness of image clustering?}

\begin{figure}[t]
	\centering
	\includegraphics[width=3.5 in]{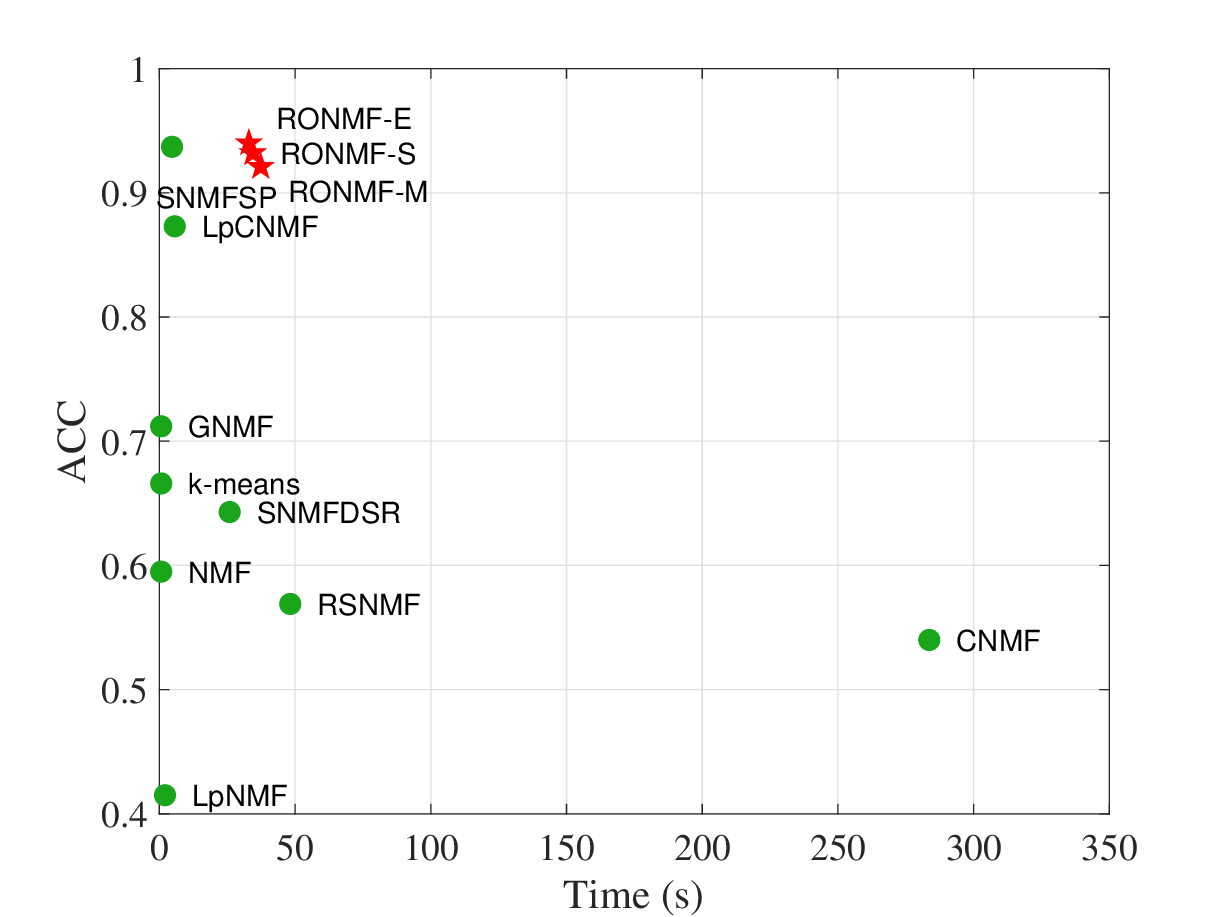}
	 \vskip-0.1cm
	\caption{ACC performance of our proposed RONMF (including RONMF-M, RONMF-S, and RONMF-E) and competitors on MNIST.}
	\label{post1}
\end{figure}

Motivated by the above studies, in this paper, we propose a robust orthogonal NMF method, termed RONMF. Note that the orthogonal constraints are imposed on the basis matrix to improve the discriminability of features and further improve the robustness \cite{li2019orthogonal,tabealhojeh2023rmaml}. The primary contributions of this paper can be described in the following three aspects.

\begin{enumerate}[1)]
\item We propose a unified NMF framework that utilizes non-convex functions as the error loss for better feature selection. To the best of our knowledge, this non-convex NMF method has not been investigated in the literature.
\item We enforce orthogonal constraints on the basic matrix to enhance sparsity, thereby reducing the correlation among basis vectors and improving the generalization performance. Besides, we consider the label information of labeled and unlabeled data with different weights, where graph Laplacian regularization explores the internal geometric structure of high-dimensional data and label propagation obtains the predicted membership values of each category.
\item We propose an optimization algorithm based on the alternating direction method of multipliers (ADMM), whose subproblems admit explicit solutions. Extensive experiments demonstrate the superiority and robustness of the proposed method in various indicators. As shown in Figure  \ref{post1}, RONME with different non-convex functions (marked by red stars) can achieve high accuracy in an acceptable runtime.
\end{enumerate}

The structure of this paper is as follows. Section \ref{sec2} briefly reviews NMF basics and variants. Section \ref{sec3} presents the proposed method and optimization algorithm. Section \ref{sec4} provides the experimental results and discussions. Finally, Section \ref{sec5} summarizes this paper.

\section{Related Work}\label{sec2}

This section first introduces NMF basics, followed by some popular NMF variants.

\subsection{NMF Basics}
NMF is designed to approximate the high-dimensional data matrix $X \in \mathbb{R}^{d\times n}$ by decomposing it into a basis matrix $U\in \mathbb{R}^{d\times r}$ and a coefficient matrix $V\in \mathbb{R}^{n\times r}$, where $d$ is the sample dimension and $n$ is the sample number. Mathematically, the classical NMF can be represented by
\begin{equation}\label{nmf}
\begin{aligned}
\min_{U,V}~&\|X-UV^\top\|_\textrm{F}^2\\
\textrm{s.t.}~~&U\geq0,~V\geq0.
\end{aligned}
\end{equation}

To enhance the robustness to noise and outliers, Kong \textit{et al.} \cite{kong2011robust} introduced $\ell_{2,1}$-norm into the objective function of \eqref{nmf} and proposed the following RNMF model
\begin{equation}\label{rnmf}
\begin{aligned}
\min_{U,V}~&\|X-UV^\top\|_{2,1}\\
\textrm{s.t.}~~&U\geq0,~V\geq0,
\end{aligned}
\end{equation}
where $\ell_{2,1}$-norm is defined as
\begin{equation}
\|Y\|_{2,1}=\sum_{i=1}^d \|Y_i\|_2
\end{equation}
with $Y_i$ being the $i$-th row of $Y$. It is verified that RNMF is more robust than NMF because $\ell_{2,1}$-norm can effectively remove noise and outliers row-wisely.

To ensure the underlying geometric structure of the data, Cai \textit{et al.} \cite{cai2010graph} introduced GNMF as follows
\begin{equation}\label{gnmf}
\begin{aligned}
\min_{U,V}~&\|X-UV^\top\|_\textrm{F}^2+\lambda\textrm{Tr}(V^\top LV)\\
\textrm{s.t.}~~&U\geq0,~V\geq0,
\end{aligned}
\end{equation}
where $\lambda>0$ is the trade-off parameter and $\textrm{Tr}(\cdot)$ denotes the matrix trace. Besides, $L=D-W\in \mathbb{R}^{n\times n}$ is the Laplacian matrix with $D$ being the degree matrix and $W$ being the weight matrix \cite{pang2017graph}. However, GNMF heavily relies on the quality and structure of graph information.

\begin{table*}[t]
	\caption{Non-convex functions and related proximal operators.}\label{nonconvex}
	\setlength\tabcolsep{6mm}{{\renewcommand\baselinestretch{1}\selectfont
			{\centering\begin{tabular}{|c| c| c|}\hline
					Penalty &$\phi_\sigma(x)$&$\textrm{Prox}_{\phi,\sigma}(v)$\\\hline\hline
					MCP ($\tau> 1$)&
					$\left\{
					\begin{array}{ll}
						\sigma |x|-\frac{x^2}{2\tau},  &\textrm{if}~|x|\leq \sigma \tau \\
						\frac{\sigma^2\tau}{2},  &\textrm{if}~|x| > \sigma \tau\\
					\end{array}
					\right.$
					&$\left\{
					\begin{array}{ll}
						0,  &\textrm{if}~|v|\leq \sigma \\
						\textrm{sgn}(v)\frac{\tau(|v|-\sigma)}{\tau-1},&\textrm{if}~\sigma \leq |v|\leq \sigma\tau \\
						v,  &\textrm{if}~|v| > \sigma \tau\\
					\end{array}
					\right.$\\\hline
					SCAD ($\tau> 2$)&
					$\left\{
					\begin{array}{ll}
						\sigma |x|,  &\textrm{if}~|x|\leq \sigma\\
						\frac{\sigma \tau|x|-\frac{1}{2}(|x|^2+\sigma^2)}{\tau-1},  &\textrm{if}~\sigma\leq |x| \leq \sigma\tau\\
						\frac{\sigma^2(\tau+1)}{2},  &\textrm{if}~|x| \geq \sigma \tau\\
					\end{array}
					\right.$
					&$\left\{
					\begin{array}{ll}
						0,  &\textrm{if}~|v|\leq \sigma\\
						\textrm{sgn}(v)(|v|-\sigma),  &\textrm{if}~\sigma \leq |v|\leq 2\sigma \\
						\textrm{sgn}(v)\frac{(\tau-1)|v|-\sigma \tau}{\tau-2},  &\textrm{if}~2\sigma \leq |v|\leq \sigma \tau\\
						v,  &\textrm{if}~|v| > \sigma \tau\\
					\end{array}
					\right.$\\\hline
					ETP &
					$\frac{\sigma}{1 - \exp(-\gamma)} \left(1 - \exp(-\gamma |x|)\right)$
					&$\left\{
					\begin{array}{ll}
						0 & \text{if } |v| \leq \sigma \\
						\text{sgn}(v)(|v| - \frac{\sigma}{\gamma}) & \text{if } \sigma < |v| \leq \sigma(1 + \frac{1}{\gamma}) \\
						v & \text{if } |v| > \sigma(1 + \frac{1}{\gamma})\\
					\end{array}
					\right.$\\
					\hline
				\end{tabular}\par}}}
\end{table*}

\subsection{Constrained NMF}
Assume that only the first $p$ samples have label information, which can be divided into $c$ classes.
Define the label matrix as the form of
\begin{equation}
	A=\begin{bmatrix}
	 C & 0  \\
	 0 & I
	 \end{bmatrix}\in \mathbb{R}^{n\times (n+c-p)},
\end{equation}
where $C$ is an indicator matrix of size $p\times c$ defined as
    \begin{equation}
    C_{ij}=\begin{cases}
    1,~\textrm{if $X_i$ belongs to the $j$-th class},\\
    0,~\textrm{otherwise},
    \end{cases}
	\end{equation}
and $I$ is the identity matrix of size $(n-p)\times (n-p)$.

According to this label information, Liu \textit{et al.} \cite{liu2011constrained} reformulated NMF as the following CNMF model
\begin{equation}\label{cnmf}
\begin{aligned}
\min_{U,Z}~&\|X-UZ^\top A^\top\|_\textrm{F}^2\\
\textrm{s.t.}~~&U\geq0,~Z\geq0,
\end{aligned}
\end{equation}
in which the auxiliary matrix $Z \in \mathbb{R}^{(n+c-p) \times r}$ satisfies the equation $V = AZ$, ensuring that samples from the same class share a similar representation. However, the above CNMF does not have any constraints on unlabeled samples, which may bring failures when there is only one labeled sample per class.

Very recently,  Liu \textit{et al.} \cite{liu2023constrained}  introduced the label propagation technique to infer label information for unlabeled samples based on their proximity to labeled samples \cite{lan2020label}.
LpCNMF is expressed as
\begin{equation}\label{lpcnmf}
\begin{aligned}
\min_{U,A,Z}~&\|X-UZ^\top A^\top\|_\textrm{F}^2+\lambda (\textrm{Tr}(A^\top LA)\\
&+\textrm{Tr}((A-Y)^\top S (A-Y)))\\
\textrm{s.t.}~~~&U\geq0,~A\geq0,~Z\geq0,
\end{aligned}
\end{equation}
where $S\in \mathbb{R}^{n \times n}$ is the label indicator defined as
    \begin{equation}
    S_{ii}=\begin{cases}
    1,~\textrm{if $X_i$ is a label sample},\\
    0,~\textrm{otherwise},
    \end{cases}
	\end{equation}
and
\begin{equation}
	Y=\begin{bmatrix}
	 C   \\
	 0
	 \end{bmatrix}\in \mathbb{R}^{n \times c}.
\end{equation}
Unlike CNMF, $A\in \mathbb{R}^{n \times c}$ in \eqref{lpcnmf} is learned during optimization, which is more friendly to unlabeled samples.

\section{Robust Orthogonal NMF}\label{sec3}

In this section, we first present our new model and then discuss its optimization algoirhm.

\subsection{Model Formulation}
From GNMF and LpCNMF, it is concluded that local geometric structures can be effectively captured through nearest-neighbor graphs on discrete data points~\cite{belkin2001laplacian}, while utilizing models to generate pseudo-labels is a widely employed approach in semi-supervised learning.
The inherent information of the image and the prediction of pseudo labels can be attached to the NMF model.

To improve the robustness, in this paper, we construct robust orthogonal NMF (RONMF), which is given by
\begin{equation}\label{ronmf}
\begin{aligned}
\min_{U,A,Z}~&\|X-UZ^\top A^\top\|_{2,\phi}+\lambda \textrm{Tr}(A^\top LA)\\
&+\mu \textrm{Tr}((A-Y)^\top S (A-Y))\\
\textrm{s.t.}~~~&U\geq0,~A\geq0,~Z\geq0,~U^\top U=I,
\end{aligned}
\end{equation}
where $I\in \mathbb{R}^{r\times r}$ is an identity matrix. Moreover, the combination of the non-negative constraint and the orthogonal constraint ensures that each row of the matrix contains at most one positive element, theoretically guaranteeing the sparsity of the matrix and preventing overlap between the basis vectors \cite{jiang2023exact}. Note that $\|\cdot\|_{2,\phi}$ is a generalized structured non-convex function, defined as
\begin{equation}
\|Y\|_{2,\phi}=\sum_{i=1}^d\phi\left(\|Y_i\|_2\right),
\end{equation}
where $\phi$ covers these functions mentioned in \cite{zhang2022structured}. Here, we take MCP, SCAD, and ETP as representatives, see Table \ref{nonconvex}.

Obviously, the main differences compared to LpCNMF in \eqref{lpcnmf} can be summarized as three aspects.
\begin{itemize}
  \item It introduces $\|\cdot\|_{2,\phi}$ to measure $X-UZ^\top A^\top$, which is more robust than F-norm.
  \item It enforces $U^\top U=I$ for the basis matrix $U$ to make feature selection discriminative.
  \item It utilizes different weights for label regularization terms, i.e., $\lambda$ for $\textrm{Tr}(A^\top LA)$ and $\mu$ for $\textrm{Tr}((A-Y)^\top S (A-Y))$. We verify the different effects of these two regularizations in ablation experiments.
\end{itemize}

Therefore, the proposed RONMF can not only take advantage of LpCNMF  but also be robust to noise and outliers.

\subsection{Optimization Algorithm}

Due to the introduction of non-convex functions and orthogonal constraints, problem  \eqref{ronmf}  is not easy to be calculated directly. Hence, we first reformulate  \eqref{ronmf} as
\begin{equation}\label{rnmf-v}
\begin{aligned}
\min_{U,A,Z,E}~&\|E\|_{2,\phi}+\lambda \textrm{Tr}(A^\top LA)\\
&+\mu \textrm{Tr}((A-Y)^\top S (A-Y))\\
\textrm{s.t.}~~~&U\geq0,~A\geq0,~Z\geq0,~U^\top U=I,\\
&X-UZ^\top A^\top=E.
\end{aligned}
\end{equation}
The augmented Lagrange function is
\begin{equation}
\begin{aligned}
\mathcal{L}_\beta (U,A,Z,E,\Lambda)&=\|E\|_{2,\phi}+\lambda \textrm{Tr}(A^\top LA)\\
&+\mu \textrm{Tr}((A-Y)^\top S (A-Y))\\
&-\langle\Lambda, X-UZ^\top A^\top-E \rangle\\
& +\frac{\beta}{2}\|X-UZ^\top A^\top-E\|_\textrm{F}^2
\end{aligned}
\end{equation}
with constraints \( U \geq 0 \), \( A \geq 0 \), \( Z \geq 0 \), and \( U^\top U = I \), where \( \Lambda \) denotes the Lagrange multiplier and \( \beta > 0 \) is the penalty parameter.
Following the ADMM \cite{zhang2023linear}, all variables can be updated iteratively using a Gauss-Seidel fashion, with each iteration providing a closed-form solution.

\subsubsection{Fix $A,Z,E,\Lambda$, update $U$ via}
\begin{equation}\label{sub-u}
\begin{aligned}
\min_{U}~&\frac{1}{2}\|X-UZ^{k\top} A^{k\top}-E^k-\Lambda^k/\beta\|_\textrm{F}^2\\
\textrm{s.t.}~~&U\geq0,~U^\top U=I.\\
\end{aligned}
\end{equation}
In fact, it is an optimization problem with both non-negative and orthogonal constraints. Denote
\begin{equation}
f(U)=\frac{1}{2}\|W-UZ^{k\top} A^{k\top}\|_\textrm{F}^2,
\end{equation}
where $W=X-E^k-\Lambda^k/\beta$. According to \cite{jiang2023exact}, it obtains the exact penalty function of \eqref{sub-u} as
\begin{equation}
f_{\sigma}(U)=f(U)+\sigma(\|Uv\|^{2}_{2}-1),
\end{equation}
where $v=e/\sqrt{r}$ with $e\in \mathbb{R}^{r}$ being the real vector whose elements are all $1$, and $\sigma>0$ is a penalty parameter.
It then follows that the Riemannian gradient of $f_{\sigma}(U)$ at point $U$ is
\begin{equation}
\textrm{grad}~f_{\sigma}(U)=\nabla f_{\sigma}(U)-U \textrm{Diag}(U^\top\nabla f_{\sigma}(U)).
\end{equation}

Overall, the iterative scheme for updating $U$ in \eqref{sub-u} is detailed in Algorithm \ref{sol:U}.

\begin{algorithm}[t]
\caption{Update $U$}\label{sol:U}
\textbf{Input:} Compute $W$, parameters $\sigma, \epsilon_1, \epsilon_2$\\
\textbf{Initialize:} $U^{0}=U^{k}$, set $t=0$\\
\textbf{While} not converged \textbf{do}	
		\begin{algorithmic}[1]
            \STATE  Find $U^{t+1}$ by the projection gradient method such that
		\begin{equation}\label{eq:stopw}	
        \begin{aligned}
        \nonumber
			&\|\min (U^{t+1},\textrm{grad}~f_{\sigma}(U^{t+1}))\|_\textrm{F}\leq \epsilon_1\\
			&f_{\sigma}(U^{t+1})\leq f_{\sigma}(U^{t})
        \end{aligned}
		\end{equation}
            \STATE  Check convergence: $\|U^{t+1}v\|^{2}_{2}-1 \leq \epsilon_2$
		\end{algorithmic}
\textbf{End While}\\
\textbf{Output:} $U^{k+1}=U^{t+1}$
\end{algorithm}

\subsubsection{Fix $U,Z,E,\Lambda$, update $A$ via}
\begin{equation}\label{sub-a}
\begin{aligned}
\min_{A\geq0}~&\frac{\beta}{2}\|W^\top-A Z^{k} U^{k+1\top}\|_\textrm{F}^2+\lambda \textrm{Tr}(A^\top LA)\\
&+\mu \textrm{Tr}((A-Y)^\top S (A-Y))).
\end{aligned}
\end{equation}
This, together with the fact that $U^{k+1\top}U^{k+1}=I$, derives the optimality condition of \eqref{sub-a} as
\begin{equation}
\begin{aligned}
0=&-\beta W^\top U^{k+1}Z^{k\top}-A Z^{k} Z^{k\top}\\
&+(\lambda L+\mu S)A-\mu SY,\\
\end{aligned}
\end{equation}
which can be easily obtained by Lyapunov solvers, and the solution is denoted by $\bar{A}$.
Therefore, \eqref{sub-a} admits the following closed-form solution
\begin{equation}\label{sol-a}
A^{k+1}=\max\left\{\bar{A},0\right\}.
\end{equation}

\subsubsection{Fix $U,A,E,\Lambda$, update $Z$ via}
\begin{equation}\label{sub-z}
\min_{Z\geq0}~\frac{1}{2}\|W^\top -A^{k+1}Z U^{k+1\top}\|_\textrm{F}^2
\end{equation}
The optimality condition of \eqref{sub-z} is
\begin{equation}
\begin{aligned}
0=-A^{k+1\top}W^\top U^{k+1} + A^{k+1\top} A^{k+1}Z.
\end{aligned}
\end{equation}
Let $\bar{Z}=(A^{k+1\top} A^{k+1})^{-1}A^{k+1\top}W^\top U^{k+1}$. Then, the closed-form solution of \eqref{sub-z} is
\begin{equation}\label{sol-z}
Z^{k+1}=\max\left\{\bar{Z},0\right\}.
\end{equation}

\subsubsection{Fix $U,A,Z,\Lambda$, update $E$ via}
\begin{equation}\label{sub-E}
\min_{E}~\frac{\beta}{2}\|E-V\|_\textrm{F}^2+\|E\|_{2,\phi},
\end{equation}
where $V=X-U^{k+1}Z^{k+1 \top} A^{k+1 \top}-\Lambda^k/\beta$.
Then, the closed-form solution of \eqref{sub-E} can be given by
\begin{equation}\label{sol-e}
E^{k+1}_i=\frac{V_i}{\|V_i\|_2}\circ \textrm{Prox}_{\phi,1/\beta}\left(\|V_i\|_2\right),
\end{equation}
where $\circ$ denotes the element-wise computation, and proximal operators are provided in  Table \ref{nonconvex}.

\subsubsection{Fix $U,A,Z,E$, update $\Lambda$ via}
\begin{equation}\label{sol-d}
\begin{aligned}
\Lambda^{k+1}&=\Lambda^k-\beta(X-U^{k+1}Z^{k+1 \top} A^{k+1 \top}-E^{k+1}).
\end{aligned}
\end{equation}

Therefore, the overall optimization framework is summarized in Algorithm \ref{ADMM}.

\begin{algorithm}[t]
\caption{Optimization Algorithm} \label{ADMM}
\textbf{Input:} Given matrices $X, L, Y$,~parameters $\lambda, \mu$\\
\textbf{Initialize:} $(U^0,A^0,Z^0,E^0,\Lambda^0)$, set $k=0$\\
\textbf{While} not converged \textbf{do}	
		\begin{algorithmic}[1]
            \STATE  Update $U^{k+1}$ by Algorithm \ref{sol:U}
            \STATE  Update $A^{k+1}$ by \eqref{sol-a}
            \STATE  Update $Z^{k+1}$ by \eqref{sol-z}
			\STATE  Update $E^{k+1}$ by \eqref{sol-e}
            \STATE  Update $\Lambda^{k+1}$ by \eqref{sol-d}
		\end{algorithmic}
\textbf{End While}\\
\textbf{Output:} $(U,A,Z)$
\end{algorithm}

\subsubsection{Convergence and Complexity }

\begin{table}[h]
\centering
\caption{Updated hierarchical complexity analysis.} \label{complexity}
\begin{tabular}{|c|c|c|}
\hline
\setlength{\tabcolsep}{1pt}
\renewcommand{\arraystretch}{1.1}
Variables & Complexity & Dominant Operation  \\
\hline
$U$ & $\mathcal{O}(T \cdot (d^2r + dnr))$ & Convergence iterations \\
$A$ & $\mathcal{O}(n^3 + n^2r)$ & Lyapunov equation \\
$Z$ & $\mathcal{O}(r^3 + dr^2)$ & Matrix inversion \\
$E$ & $\mathcal{O}(dn)$ & Matrix multiplication \\
$\Lambda$ & $\mathcal{O}(dn)$ & Element-wise operations\\
\hline
\end{tabular}
\end{table}

Under the ADMM framework, variables are updated sequentially through

\begin{equation} \label{eq:ADMM_update}
\begin{cases}
 \mathcal{L}_\beta \left(U^{k+1},A^k,Z^k,E^k,\Lambda^k\right) \\ ~~~~~\leq \mathcal{L}_\beta \left(U^k,A^k,Z^k,E^k,\Lambda^k\right), \\
 \mathcal{L}_\beta \left(U^{k+1},A^{k+1},Z^k,E^k,\Lambda^k\right) \\ ~~~~~\leq \mathcal{L}_\beta \left(U^{k+1},A^k,Z^k,E^k,\Lambda^k\right), \\
 \mathcal{L}_\beta \left(U^{k+1},A^{k+1},Z^{k+1},E^k,\Lambda^k\right) \\ ~~~~~\leq \mathcal{L}_\beta \left(U^{k+1},A^{k+1},Z^k,E^k,\Lambda^k\right), \\
 \mathcal{L}_\beta \left(U^{k+1},A^{k+1},Z^{k+1},E^{k+1},\Lambda^k\right) \\ ~~~~~\leq \mathcal{L}_\beta \left(U^{k+1},A^{k+1},Z^{k+1},E^k,\Lambda^k\right), \\
  \mathcal{L}_\beta \left(U^{k+1},A^{k+1},Z^{k+1},E^{k+1},\Lambda^{k+1}\right) \\ ~~~~~\leq \mathcal{L}_\beta \left(U^{k+1},A^{k+1},Z^{k+1},E^{k+1},\Lambda^k\right).
\end{cases}
\end{equation}

Therefore, the following conclusion holds
\begin{equation} \label{eq:ADMM_update}
\begin{aligned}
&  \mathcal{L}_\beta \left(U^{k+1},A^{k+1},Z^{k+1},E^{k+1},\Lambda^{k+1}\right) \\
&~~~~~\leq \mathcal{L}_\beta \left(U^k,A^k,Z^k,E^k,\Lambda^k\right).
\end{aligned}
\end{equation}

For computational complexity, Table \ref{complexity} provides the hierarchical analysis.
Let $X \in \mathbb{R}^{d \times n}$ be the input matrix with latent dimension $r \ll \min(d,n)$, and $T$ denote the number of iterations.
The first step is to update $U$, which is dominated by the manifold projection operation ($\mathcal{O}(d^2r)$) and the Frobenius norm reconstruction term ($\mathcal{O}(dnr)$).
The second step is to update $A$, which requires solving the Lyapunov equation and performing graph regularization.
The dense matrix inversion dominates with a complexity of $\mathcal{O}(n^3)$, while the graph Laplacian term $\text{Tr}(A^\top LA)$ can be accelerated to $\mathcal{O}(n^2r)$ by sparse matrices.
The third step is to update $Z$, which is implemented by matrix inversion $(A^\top A)^{-1}$ and multiplication chain $A^\top W^\top U$ , with a complexity of $\mathcal{O}(r^3 + dr^2)$.
In addition, the computational complexity of updating $E$ and $\Lambda$ is $\mathcal{O}(dn)$.
Therefore, the total complexity can be summarized as $\mathcal{O}\left(T(d^2r + dnr) + n^3 + dn)\right)$.

\begin{table}[t]
\centering
\caption{Dataset information.}\label{datasets}
\setlength{\tabcolsep}{8pt}
\renewcommand{\arraystretch}{1.1}
\begin{tabular}{|c c|c|c|c|}
\hline
\multicolumn{2}{|c|}{Datasets} & Size  & Classes & Samples \\ \hline
\hline
\multirow{2}{*}[-0.5ex]{Type-\uppercase\expandafter{\romannumeral1}}
&UMIST   & 644   & 20            & 575           \\
&YALE    & 1024 & 15            & 165           \\\hline
\multirow{2}{*}[-0.5ex]{Type-\uppercase\expandafter{\romannumeral2}}&COIL20  & 1024 & 20            & 1440          \\
&COIL100 & 1024 & 100           & 7200          \\\hline
\multirow{2}{*}[-0.5ex]{Type-\uppercase\expandafter{\romannumeral3}}&USPS    & 256 & 10            & 9298         \\
&MNIST   & 784   & 10            & 20000         \\ \hline
\multirow{2}{*}[-0.5ex]{Type-\uppercase\expandafter{\romannumeral4}}&ORL10P   & 10304   & 10          & 100         \\
&CAL101   & 60000   & 101            & 9000         \\ \hline
\end{tabular}
\end{table}

\begin{table*}[ht!]
	\centering
	\caption{Clustering (mean±std) results of compared methods on the Type-\uppercase\expandafter{\romannumeral1} datasets. The best and second-best results are marked in \textcolor{red}{red} and \textcolor{blue}{blue}, respectively.}\label{tabclean1}
	\setlength{\tabcolsep}{1.5pt}
		\scalebox{0.98}{
	\renewcommand{\arraystretch}{1.1}
	\begin{tabular}{|c|c|ccccccccc|ccc|}
		\hline
		\multicolumn{2}{|c|}{Type-\uppercase\expandafter{\romannumeral1}} & \makecell{ k-means \\ \cite{liu2008reducing}} & \makecell{ NMF \\ \cite{lee1999learning}} & \makecell{ GNMF \\ \cite{cai2010graph}} & \makecell{ RSNMF \\ \cite{li2017robust}} & \makecell{ SNMFDSR \\ \cite{jia2019semi}} & \makecell{ CNMF \\ \cite{liu2011constrained} }&\makecell{  LpNMF \\ \cite{lan2020label}} & \makecell{ LpCNMF \\ \cite{liu2023constrained}} & \makecell{ SNMFSP \\ \cite{jing2025semi}} & RONMF-M & RONMF-S & RONMF-E \\ \hline \hline
		\multirow{4}{*}[-3.5ex]{\makebox[0.05\textwidth][c]{UMIST}}
		& ACC & \makecell{0.716 \\ \scriptsize ($\pm$0.070)} & \makecell{0.671 \\ \scriptsize ($\pm$0.101)} & \makecell{0.786 \\ \scriptsize ($\pm$0.152)} & \makecell{0.694 \\ \scriptsize ($\pm$0.100)} & \makecell{0.711 \\ \scriptsize ($\pm$0.133)} & \makecell{0.761 \\ \scriptsize ($\pm$0.074)} & \makecell{0.569 \\ \scriptsize ($\pm$0.092)} & \makecell{0.823 \\ \scriptsize ($\pm$0.123)} & \makecell{0.687 \\ \scriptsize ($\pm$0.016)} & \makecell{\blue{0.945} \\ \scriptsize \blue{($\pm$0.030)}} & \makecell{0.930 \\ \scriptsize ($\pm$0.015)}& \makecell{\red{0.954} \\ \scriptsize \red{($\pm$0.016)}} \\
& F1 & \makecell{0.727 \\ \scriptsize ($\pm$0.059)} & \makecell{0.690 \\ \scriptsize ($\pm$0.078)} & \makecell{0.811 \\ \scriptsize ($\pm$0.110)} & \makecell{0.704 \\ \scriptsize ($\pm$0.077)} & \makecell{0.753 \\ \scriptsize ($\pm$0.111)} & \makecell{0.784 \\ \scriptsize ($\pm$0.034)} & \makecell{0.607 \\ \scriptsize ($\pm$0.091)} & \makecell{0.859 \\ \scriptsize ($\pm$0.088)} & \makecell{0.744 \\ \scriptsize ($\pm$0.019)} & \makecell{0.915 \\ \scriptsize ($\pm$0.028)} & \makecell{\blue{0.916} \\ \scriptsize \blue{($\pm$0.014)}}& \makecell{\red{0.977} \\ \scriptsize \red{($\pm$0.029)}}\\
& NMI & \makecell{0.700 \\ \scriptsize ($\pm$0.066)} & \makecell{0.660 \\ \scriptsize ($\pm$0.084)} & \makecell{0.805 \\ \scriptsize ($\pm$0.100)} & \makecell{0.690 \\ \scriptsize ($\pm$0.088)} & \makecell{0.741 \\ \scriptsize ($\pm$0.120)} & \makecell{0.783 \\ \scriptsize ($\pm$0.053)} & \makecell{0.564 \\ \scriptsize ($\pm$0.097)} & \makecell{0.836 \\ \scriptsize ($\pm$0.080)} & \makecell{0.726 \\ \scriptsize ($\pm$0.019)}  & \makecell{0.919 \\ \scriptsize ($\pm$0.022)} & \makecell{\blue{0.930} \\ \scriptsize \blue{($\pm$0.012)}}& \makecell{\red{0.966} \\ \scriptsize \red{($\pm$0.023)}}\\
& PUR & \makecell{0.732 \\ \scriptsize ($\pm$0.049)} & \makecell{0.691 \\ \scriptsize ($\pm$0.075)} & \makecell{0.812 \\ \scriptsize ($\pm$0.106)} & \makecell{0.732 \\ \scriptsize ($\pm$0.073)} & \makecell{0.754 \\ \scriptsize ($\pm$0.102)} & \makecell{0.789 \\ \scriptsize ($\pm$0.033)} & \makecell{0.605 \\ \scriptsize ($\pm$0.082)} & \makecell{0.858 \\ \scriptsize ($\pm$0.091)} & \makecell{0.687 \\ \scriptsize ($\pm$0.016)} & \makecell{0.907 \\ \scriptsize ($\pm$0.033)}  & \makecell{\blue{0.929} \\ \scriptsize \blue{($\pm$0.012)}}& \makecell{\red{0.956} \\ \scriptsize \red{($\pm$0.020)}}\\ \hline
		\multirow{4}{*}[-3.5ex]{YALE}
		& ACC & \makecell{0.461 \\ \scriptsize ($\pm$0.063)} & \makecell{0.471 \\ \scriptsize ($\pm$0.090)} & \makecell{0.420 \\ \scriptsize ($\pm$0.050)} & \makecell{0.515 \\ \scriptsize ($\pm$0.065)} & \makecell{0.527 \\ \scriptsize ($\pm$0.112)} & \makecell{0.398 \\ \scriptsize ($\pm$0.084)} & \makecell{0.420 \\ \scriptsize ($\pm$0.079)} & \makecell{0.433 \\ \scriptsize ($\pm$0.076)} & \makecell{0.680 \\ \scriptsize ($\pm$0.080)} & \makecell{0.781 \\ \scriptsize ($\pm$0.079)} & \makecell{\red{0.875} \\ \scriptsize \red{($\pm$0.063)}}& \makecell{\blue{0.871} \\ \scriptsize \blue{($\pm$0.047)}} \\
& F1 & \makecell{0.514 \\ \scriptsize ($\pm$0.064)} & \makecell{0.516 \\ \scriptsize ($\pm$0.078)} & \makecell{0.468 \\ \scriptsize ($\pm$0.045)} & \makecell{0.558 \\ \scriptsize ($\pm$0.065)} & \makecell{0.556 \\ \scriptsize ($\pm$0.087)} & \makecell{0.440 \\ \scriptsize ($\pm$0.076)} & \makecell{0.462 \\ \scriptsize ($\pm$0.073)} & \makecell{0.540 \\ \scriptsize ($\pm$0.079)} & \makecell{0.721 \\ \scriptsize ($\pm$0.060)} & \makecell{0.868 \\ \scriptsize ($\pm$0.031)} & \makecell{\blue{0.875} \\ \scriptsize \blue{($\pm$0.061)}}& \makecell{\red{0.909} \\ \scriptsize \red{($\pm$0.075)}}\\
& NMI & \makecell{0.354 \\ \scriptsize ($\pm$0.069)} & \makecell{0.351 \\ \scriptsize ($\pm$0.097)} & \makecell{0.288 \\ \scriptsize ($\pm$0.067)} & \makecell{0.465 \\ \scriptsize ($\pm$0.076)} & \makecell{0.433 \\ \scriptsize ($\pm$0.115)} & \makecell{0.321 \\ \scriptsize ($\pm$0.100)} & \makecell{0.307 \\ \scriptsize ($\pm$0.078)} & \makecell{0.395 \\ \scriptsize ($\pm$0.079)} & \makecell{0.566 \\ \scriptsize ($\pm$0.052)} & \makecell{0.818 \\ \scriptsize ($\pm$0.054)} & \makecell{\blue{0.873} \\ \scriptsize \blue{($\pm$0.054)}}& \makecell{\red{0.920} \\ \scriptsize \red{($\pm$0.040)}}\\
& PUR & \makecell{0.468 \\ \scriptsize ($\pm$0.063)} & \makecell{0.473 \\ \scriptsize ($\pm$0.075)} & \makecell{0.424 \\ \scriptsize ($\pm$0.044)} & \makecell{0.515 \\ \scriptsize ($\pm$0.066)} & \makecell{0.530 \\ \scriptsize ($\pm$0.092)} & \makecell{0.406 \\ \scriptsize ($\pm$0.078)} & \makecell{0.432 \\ \scriptsize ($\pm$0.080)} & \makecell{0.505 \\ \scriptsize ($\pm$0.086)} & \makecell{0.698 \\ \scriptsize ($\pm$0.065)} & \makecell{0.825 \\ \scriptsize ($\pm$0.058)} & \makecell{\blue{0.874} \\ \scriptsize \blue{($\pm$0.056)}}& \makecell{\red{0.909} \\ \scriptsize \red{($\pm$0.048)}}\\ \hline
\hline
\multicolumn{2}{|c|}{Average} & \makecell{ 0.584 \\ \scriptsize ($\pm$0.063)} & \makecell{0.565 \\ \scriptsize ($\pm$0.085)} & \makecell{ 0.602 \\ \scriptsize ($\pm$0.084)} & \makecell{ 0.609 \\ \scriptsize ($\pm$0.076)} & \makecell{ 0.626 \\ \scriptsize ($\pm$0.109)} & \makecell{ 0.585 \\ \scriptsize ($\pm$0.067)} & \makecell{ 0.496 \\ \scriptsize ($\pm$0.084)} & \makecell{0.689 \\ \scriptsize ($\pm$0.041)} & \makecell{ 0.689 \\ \scriptsize ($\pm$0.041)} & \makecell{ 0.872 \\ \scriptsize ($\pm$0.042)} & \makecell{ \blue{0.900} \\ \scriptsize \blue{($\pm$0.036)}} &  \makecell{ \red{0.933} \\ \scriptsize \red{($\pm$0.037)}}  \\ \hline
	\end{tabular}}
\end{table*}

\begin{table*}[ht]
	\centering
	\caption{Clustering (mean±std) results of compared methods on the Type-\uppercase\expandafter{\romannumeral2} datasets. The best and second-best results are marked in \red{red} and \blue{blue}, respectively.}\label{tabclean2}
	\setlength{\tabcolsep}{1.25pt}
		\scalebox{0.98}{
	\renewcommand{\arraystretch}{1.1}
	\begin{tabular}{|c|c|ccccccccc|ccc|}
		\hline
		\multicolumn{2}{|c|}{Type-\uppercase\expandafter{\romannumeral2}} & \makecell{ k-means \\ \cite{liu2008reducing}} & \makecell{ NMF \\ \cite{lee1999learning}} & \makecell{ GNMF \\ \cite{cai2010graph}} & \makecell{ RSNMF \\ \cite{li2017robust}} & \makecell{ SNMFDSR \\ \cite{jia2019semi}} & \makecell{ CNMF \\ \cite{liu2011constrained} }&\makecell{  LpNMF \\ \cite{lan2020label}} & \makecell{ LpCNMF \\ \cite{liu2023constrained}} & \makecell{ SNMFSP \\ \cite{jing2025semi}} & RONMF-M & RONMF-S & RONMF-E \\ \hline \hline
		\multirow{4}{*}[-3.5ex]{\makebox[0.04\textwidth][c]{COIL20}}
         & ACC & \makecell{0.701 \\ \scriptsize ($\pm$0.083)} & \makecell{0.668 \\ \scriptsize ($\pm$0.102)} & \makecell{0.893 \\ \scriptsize ($\pm$0.154)} & \makecell{0.692 \\ \scriptsize ($\pm$0.120)} & \makecell{0.727 \\ \scriptsize ($\pm$0.126)} & \makecell{0.481 \\ \scriptsize ($\pm$0.038)} & \makecell{0.810 \\ \scriptsize ($\pm$0.100)} & \makecell{0.885 \\ \scriptsize ($\pm$0.073)} & \makecell{\red{0.953} \\ \scriptsize \red{($\pm$0.099)}} & \makecell{0.847 \\ \scriptsize ($\pm$0.012)} & \makecell{0.899 \\ \scriptsize ($\pm$0.037)} & \makecell{\blue{0.922} \\ \scriptsize \blue{($\pm$0.013)}}\\
& F1 & \makecell{0.772 \\ \scriptsize ($\pm$0.089)} & \makecell{0.774 \\ \scriptsize ($\pm$0.044)} & \makecell{0.894 \\ \scriptsize ($\pm$0.079)} & \makecell{0.777 \\ \scriptsize ($\pm$0.099)} & \makecell{0.713 \\ \scriptsize ($\pm$0.056)} & \makecell{0.445 \\ \scriptsize ($\pm$0.116)} & \makecell{0.719 \\ \scriptsize ($\pm$0.111)} & \makecell{\blue{0.949} \\ \scriptsize \blue{($\pm$0.060)}} & \makecell{\red{0.968} \\ \scriptsize \red{($\pm$0.067)}} & \makecell{0.908 \\ \scriptsize ($\pm$0.011)} & \makecell{0.914 \\ \scriptsize ($\pm$0.010)} & \makecell{0.934 \\ \scriptsize ($\pm$0.024)} \\
& NMI & \makecell{0.692 \\ \scriptsize ($\pm$0.103)} & \makecell{0.656 \\ \scriptsize ($\pm$0.105)} & \makecell{0.897 \\ \scriptsize ($\pm$0.131)} & \makecell{0.705 \\ \scriptsize ($\pm$0.112)} & \makecell{0.717 \\ \scriptsize ($\pm$0.148)} & \makecell{0.398 \\ \scriptsize ($\pm$0.045)} & \makecell{0.801 \\ \scriptsize ($\pm$0.099)} & \makecell{0.921 \\ \scriptsize ($\pm$0.120)} & \makecell{\red{0.974} \\ \scriptsize \red{($\pm$0.054)}} & \makecell{0.823 \\ \scriptsize ($\pm$0.011)} & \makecell{\blue{0.949} \\ \scriptsize \blue{($\pm$0.009)}} & \makecell{0.942 \\ \scriptsize ($\pm$0.007)} \\
& PUR & \makecell{0.723 \\ \scriptsize ($\pm$0.086)} & \makecell{0.701 \\ \scriptsize ($\pm$0.090)} & \makecell{0.901 \\ \scriptsize ($\pm$0.134)} & \makecell{0.727 \\ \scriptsize ($\pm$0.096)} & \makecell{0.748 \\ \scriptsize ($\pm$0.115)} & \makecell{0.518 \\ \scriptsize ($\pm$0.043)} & \makecell{0.825 \\ \scriptsize ($\pm$0.090)} & \makecell{0.926 \\ \scriptsize ($\pm$0.112)} & \makecell{\red{0.967} \\ \scriptsize \red{($\pm$0.070)}} & \makecell{0.855 \\ \scriptsize ($\pm$0.012)} & \makecell{\blue{0.939} \\ \scriptsize \blue{($\pm$0.011)}} & \makecell{\red{0.941} \\ \scriptsize \red{($\pm$0.012)}} \\ \hline
		\multirow{4}{*}[-3.5ex]{COIL100}
		& ACC & \makecell{0.765 \\ \scriptsize ($\pm$0.097)} & \makecell{0.766 \\ \scriptsize ($\pm$0.049)} & \makecell{0.886 \\ \scriptsize ($\pm$0.088)} & \makecell{0.750 \\ \scriptsize ($\pm$0.120)} & \makecell{0.699 \\ \scriptsize ($\pm$0.079)} & \makecell{0.408 \\ \scriptsize ($\pm$0.143)} & \makecell{0.697 \\ \scriptsize ($\pm$0.127)} & \makecell{\blue{0.941} \\ \scriptsize\blue{($\pm$0.062)}} & \makecell{\red{0.952} \\ \scriptsize\red{($\pm$0.102)}} & \makecell{0.881 \\ \scriptsize ($\pm$0.017)} & \makecell{0.921 \\ \scriptsize ($\pm$0.019)} & \makecell{0.924 \\ \scriptsize ($\pm$0.020)} \\
& F1 & \makecell{0.772 \\ \scriptsize ($\pm$0.089)} & \makecell{0.774 \\ \scriptsize ($\pm$0.044)} & \makecell{0.894 \\ \scriptsize ($\pm$0.079)} & \makecell{0.777 \\ \scriptsize ($\pm$0.099)} & \makecell{0.713 \\ \scriptsize ($\pm$0.056)} & \makecell{0.445 \\ \scriptsize ($\pm$0.116)} & \makecell{0.719 \\ \scriptsize ($\pm$0.111)} & \makecell{\blue{0.949} \\ \scriptsize\blue{($\pm$0.060)}} & \makecell{\red{0.968} \\ \scriptsize\red{($\pm$0.067)}} & \makecell{0.908 \\ \scriptsize ($\pm$0.011)} & \makecell{0.914 \\ \scriptsize ($\pm$0.010)} & \makecell{0.934 \\ \scriptsize ($\pm$0.024)} \\
& NMI & \makecell{0.735 \\ \scriptsize ($\pm$0.092)} & \makecell{0.735 \\ \scriptsize ($\pm$0.040)} & \makecell{0.879 \\ \scriptsize ($\pm$0.071)} & \makecell{0.741 \\ \scriptsize ($\pm$0.108)} & \makecell{0.682 \\ \scriptsize ($\pm$0.057)} & \makecell{0.289 \\ \scriptsize ($\pm$0.131)} & \makecell{0.682 \\ \scriptsize ($\pm$0.121)} & \makecell{0.934 \\ \scriptsize ($\pm$0.065)} & \makecell{\red{0.974} \\ \scriptsize\red{($\pm$0.054)}} & \makecell{0.918 \\ \scriptsize ($\pm$0.009)} & \makecell{0.923 \\ \scriptsize ($\pm$0.029)} & \makecell{\blue{0.937} \\ \scriptsize\blue{($\pm$0.024)}} \\
& PUR & \makecell{0.788 \\ \scriptsize ($\pm$0.091)} & \makecell{0.784 \\ \scriptsize ($\pm$0.044)} & \makecell{0.894 \\ \scriptsize ($\pm$0.076)} & \makecell{0.784 \\ \scriptsize ($\pm$0.103)} & \makecell{0.715 \\ \scriptsize ($\pm$0.066)} & \makecell{0.437 \\ \scriptsize ($\pm$0.120)} & \makecell{0.723 \\ \scriptsize ($\pm$0.110)} & \makecell{\blue{0.956} \\ \scriptsize\blue{($\pm$0.056)}} & \makecell{\red{0.967} \\ \scriptsize\red{($\pm$0.070)}} & \makecell{0.889 \\ \scriptsize ($\pm$0.010)} & \makecell{0.921 \\ \scriptsize ($\pm$0.021)} & \makecell{0.922 \\ \scriptsize ($\pm$0.042)} \\
\hline \hline
\multicolumn{2}{|c|}{Average} & \makecell{0.738 \\ \scriptsize ($\pm$0.090)} & \makecell{0.722 \\ \scriptsize ($\pm$0.070)} & \makecell{0.893 \\ \scriptsize ($\pm$0.108)} & \makecell{0.739 \\ \scriptsize ($\pm$0.108)} & \makecell{0.719 \\ \scriptsize ($\pm$0.096)} & \makecell{0.439 \\ \scriptsize ($\pm$0.084)} & \makecell{0.760 \\ \scriptsize ($\pm$0.105)} & \makecell{\red{0.966} \\ \scriptsize\red{($\pm$0.073)}} & \makecell{\blue{0.965} \\ \scriptsize\blue{($\pm$0.073)}} & \makecell{0.870 \\ \scriptsize ($\pm$0.012)} & \makecell{0.915 \\ \scriptsize ($\pm$0.018)} & \makecell{0.934 \\ \scriptsize ($\pm$0.020)} \\ \hline
	\end{tabular}}
\end{table*}

\begin{table*}[ht!]
\centering
\caption{Clustering (mean±std) results of compared methods on the Type-\uppercase\expandafter{\romannumeral3} datasets. The best and second-best results are marked in \red{red} and \blue{blue}, respectively.}\label{tabclean3}
\setlength{\tabcolsep}{1.25pt}
	\scalebox{0.98}{
\renewcommand{\arraystretch}{1.1}
\begin{tabular}{|c|c|ccccccccc|ccc|}
\hline

\multicolumn{2}{|c|}{Type-\uppercase\expandafter{\romannumeral3}} & \makecell{ k-means \\ \cite{liu2008reducing}} & \makecell{ NMF \\ \cite{lee1999learning}} & \makecell{ GNMF \\ \cite{cai2010graph}} & \makecell{ RSNMF \\ \cite{li2017robust}} & \makecell{ SNMFDSR \\ \cite{jia2019semi}} & \makecell{ CNMF \\ \cite{liu2011constrained} }&\makecell{  LpNMF \\ \cite{lan2020label}} & \makecell{ LpCNMF \\ \cite{liu2023constrained}}& \makecell{ SNMFSP \\ \cite{jing2025semi}} & RONMF-M & RONMF-S & RONMF-E \\ \hline \hline
\multirow{4}{*}[-3.5ex]{\makebox[0.05\textwidth][c]{USPS}}
       & ACC & \makecell{0.600 \\ \scriptsize ($\pm$0.076)} & \makecell{0.564 \\ \scriptsize ($\pm$0.055)} & \makecell{0.620 \\ \scriptsize ($\pm$0.086)} & \makecell{0.563 \\ \scriptsize ($\pm$0.061)} & \makecell{0.623 \\ \scriptsize ($\pm$0.005)} & \makecell{0.438 \\ \scriptsize ($\pm$0.051)} & \makecell{0.745 \\ \scriptsize ($\pm$0.062)} & \makecell{0.924 \\ \scriptsize ($\pm$0.067)} & \makecell{0.692 \\ \scriptsize ($\pm$0.086)} & \makecell{0.902 \\ \scriptsize ($\pm$0.016)} & \makecell{\red{0.941} \\ \scriptsize\red{($\pm$0.018)}} & \makecell{\blue{0.933} \\ \scriptsize\blue{($\pm$0.014)}} \\
& F1  & \makecell{0.531 \\ \scriptsize ($\pm$0.062)} & \makecell{0.484 \\ \scriptsize ($\pm$0.049)} & \makecell{0.648 \\ \scriptsize ($\pm$0.057)} & \makecell{0.491 \\ \scriptsize ($\pm$0.053)} & \makecell{0.523 \\ \scriptsize ($\pm$0.004)} & \makecell{0.269 \\ \scriptsize ($\pm$0.040)} & \makecell{0.752 \\ \scriptsize ($\pm$0.051)} & \makecell{0.836 \\ \scriptsize ($\pm$0.055)} & \makecell{0.759 \\ \scriptsize ($\pm$0.045)} & \makecell{0.907 \\ \scriptsize ($\pm$0.009)} & \makecell{\blue{0.928} \\ \scriptsize\blue{($\pm$0.015)}} & \makecell{\red{0.930} \\ \scriptsize\red{($\pm$0.018)}}\\
& NMI & \makecell{0.643 \\ \scriptsize ($\pm$0.062)} & \makecell{0.601 \\ \scriptsize ($\pm$0.047)} & \makecell{0.692 \\ \scriptsize ($\pm$0.056)} & \makecell{0.610 \\ \scriptsize ($\pm$0.048)} & \makecell{0.659 \\ \scriptsize ($\pm$0.013)} & \makecell{0.478 \\ \scriptsize ($\pm$0.036)} & \makecell{0.615 \\ \scriptsize ($\pm$0.061)} & \makecell{\blue{0.929} \\ \scriptsize\blue{($\pm$0.074)}} & \makecell{0.641 \\ \scriptsize ($\pm$0.041)} & \makecell{0.915 \\ \scriptsize ($\pm$0.022)} & \makecell{\red{0.930} \\ \scriptsize\red{($\pm$0.019)}} & \makecell{\red{0.930} \\ \scriptsize\red{($\pm$0.026)}}\\
& PUR & \makecell{0.659 \\ \scriptsize ($\pm$0.056)} & \makecell{0.621 \\ \scriptsize ($\pm$0.048)} & \makecell{0.689 \\ \scriptsize ($\pm$0.068)} & \makecell{0.620 \\ \scriptsize ($\pm$0.049)} & \makecell{0.629 \\ \scriptsize ($\pm$0.004)} & \makecell{0.458 \\ \scriptsize ($\pm$0.039)} & \makecell{0.772 \\ \scriptsize ($\pm$0.054)} & \makecell{\red{0.930} \\ \scriptsize\red{($\pm$0.068)}} & \makecell{0.755 \\ \scriptsize ($\pm$0.059)} & \makecell{0.889 \\ \scriptsize ($\pm$0.015)} & \makecell{0.917 \\ \scriptsize ($\pm$0.010)} & \makecell{\blue{0.929} \\ \scriptsize\blue{($\pm$0.035)}}\\ \hline
\multirow{4}{*}[-3.5ex]{MNIST}
       & ACC & \makecell{0.666 \\ \scriptsize ($\pm$0.072)} & \makecell{0.595 \\ \scriptsize ($\pm$0.024)} & \makecell{0.712 \\ \scriptsize ($\pm$0.138)} & \makecell{0.569 \\ \scriptsize ($\pm$0.091)} & \makecell{0.643 \\ \scriptsize ($\pm$0.045)} & \makecell{0.540 \\ \scriptsize ($\pm$0.010)} & \makecell{0.415 \\ \scriptsize ($\pm$0.013)} & \makecell{0.873 \\ \scriptsize ($\pm$0.095)} & \makecell{\blue{0.937} \\ \scriptsize\blue{($\pm$0.102)}} & \makecell{0.901 \\ \scriptsize ($\pm$0.014)} & \makecell{0.926 \\ \scriptsize ($\pm$0.013)} & \makecell{\red{0.940} \\ \scriptsize\red{($\pm$0.027)}} \\
& F1  & \makecell{0.697 \\ \scriptsize ($\pm$0.057)} & \makecell{0.630 \\ \scriptsize ($\pm$0.026)} & \makecell{0.766 \\ \scriptsize ($\pm$0.026)} & \makecell{0.604 \\ \scriptsize ($\pm$0.078)} & \makecell{0.676 \\ \scriptsize ($\pm$0.027)} & \makecell{0.584 \\ \scriptsize ($\pm$0.013)} & \makecell{0.437 \\ \scriptsize ($\pm$0.018)} & \makecell{\blue{0.942} \\ \scriptsize\blue{($\pm$0.072)}} & \makecell{0.956 \\ \scriptsize ($\pm$0.063)} & \makecell{0.929 \\ \scriptsize ($\pm$0.013)} & \makecell{\red{0.943} \\ \scriptsize\red{($\pm$0.011)}} & \makecell{0.933 \\ \scriptsize ($\pm$0.017)} \\
& NMI & \makecell{0.571 \\ \scriptsize ($\pm$0.001)} & \makecell{0.488 \\ \scriptsize ($\pm$0.027)} & \makecell{0.713 \\ \scriptsize ($\pm$0.114)} & \makecell{0.476 \\ \scriptsize ($\pm$0.065)} & \makecell{0.551 \\ \scriptsize ($\pm$0.030)} & \makecell{0.420 \\ \scriptsize ($\pm$0.005)} & \makecell{0.230 \\ \scriptsize ($\pm$0.008)} & \makecell{0.861 \\ \scriptsize ($\pm$0.057)} & \makecell{\red{0.980} \\ \scriptsize\red{($\pm$0.044)}} & \makecell{0.920 \\ \scriptsize ($\pm$0.014)} & \makecell{0.932 \\ \scriptsize ($\pm$0.016)} & \makecell{\blue{0.940} \\ \scriptsize\blue{($\pm$0.012)}} \\
& PUR & \makecell{0.709 \\ \scriptsize ($\pm$0.003)} & \makecell{0.648 \\ \scriptsize ($\pm$0.017)} & \makecell{0.766 \\ \scriptsize ($\pm$0.118)} & \makecell{0.618 \\ \scriptsize ($\pm$0.070)} & \makecell{0.650 \\ \scriptsize ($\pm$0.043)} & \makecell{0.571 \\ \scriptsize ($\pm$0.011)} & \makecell{0.431 \\ \scriptsize ($\pm$0.014)} & \makecell{0.941 \\ \scriptsize ($\pm$0.064)} & \makecell{\red{0.950} \\ \scriptsize\red{($\pm$0.072)}} & \makecell{0.933 \\ \scriptsize ($\pm$0.013)} & \makecell{0.922 \\ \scriptsize ($\pm$0.012)} & \makecell{\blue{0.945} \\ \scriptsize\blue{($\pm$0.015)}} \\ \hline \hline
\multicolumn{2}{|c|}{Average} & \makecell{0.635 \\ \scriptsize ($\pm$0.049)} & \makecell{0.579 \\ \scriptsize ($\pm$0.037)} & \makecell{0.701 \\ \scriptsize ($\pm$0.083)} & \makecell{0.569 \\ \scriptsize ($\pm$0.064)} & \makecell{0.619 \\ \scriptsize ($\pm$0.021)} & \makecell{0.470 \\ \scriptsize ($\pm$0.026)} & \makecell{0.550 \\ \scriptsize ($\pm$0.035)} & \makecell{0.904 \\ \scriptsize ($\pm$0.069)} & \makecell{0.822 \\ \scriptsize ($\pm$0.064)} & \makecell{0.912 \\ \scriptsize ($\pm$0.015)} & \makecell{\blue{0.930} \\ \scriptsize\blue{($\pm$0.014)}} & \makecell{\red{0.935} \\ \scriptsize\red{($\pm$0.021)}} \\ \hline
\end{tabular}}
\end{table*}

\begin{table*}[ht]
\centering
\caption{Clustering (mean±std) results of compared methods on the Type-\uppercase\expandafter{\romannumeral4} datasets. The best and second-best results are marked in \red{red} and \blue{blue}, respectively.}\label{tabclean4}
\setlength{\tabcolsep}{1.25pt}
	\scalebox{0.98}{
\renewcommand{\arraystretch}{1.1}
\begin{tabular}{|c|c|ccccccccc|ccc|}
\hline

\multicolumn{2}{|c|}{Type-\uppercase\expandafter{\romannumeral4}} & \makecell{ k-means \\ \cite{liu2008reducing}} & \makecell{ NMF \\ \cite{lee1999learning}} & \makecell{ GNMF \\ \cite{cai2010graph}} & \makecell{ RSNMF \\ \cite{li2017robust}} & \makecell{ SNMFDSR \\ \cite{jia2019semi}} & \makecell{ CNMF \\ \cite{liu2011constrained} }&\makecell{  LpNMF \\ \cite{lan2020label}} & \makecell{ LpCNMF \\ \cite{liu2023constrained}}& \makecell{ SNMFSP \\ \cite{jing2025semi}} & RONMF-M & RONMF-S & RONMF-E \\ \hline \hline
\multirow{4}{*}[-3.5ex]{\makebox[0.05\textwidth][c]{ORL10P}}
        & ACC & \makecell{0.833 \\ \scriptsize ($\pm$0.079)} & \makecell{0.810 \\ \scriptsize ($\pm$0.087)} & \makecell{0.820 \\ \scriptsize ($\pm$0.082)} & \makecell{0.869 \\ \scriptsize ($\pm$0.060)} & \makecell{0.750 \\ \scriptsize ($\pm$0.118)} & \makecell{0.671 \\ \scriptsize ($\pm$0.086)} & \makecell{0.807 \\ \scriptsize ($\pm$0.056)} & \makecell{0.855 \\ \scriptsize ($\pm$0.064)} & \makecell{0.935 \\ \scriptsize ($\pm$0.005)} & \makecell{\blue{0.967} \\ \scriptsize\blue{($\pm$0.070)}} & \makecell{\red{0.983} \\ \scriptsize\red{($\pm$0.053)}} & \makecell{\red{0.983} \\ \scriptsize\red{($\pm$0.053)}} \\
& F1 & \makecell{0.842 \\ \scriptsize ($\pm$0.069)} & \makecell{0.825 \\ \scriptsize ($\pm$0.070)} & \makecell{0.837 \\ \scriptsize ($\pm$0.063)} & \makecell{0.869 \\ \scriptsize ($\pm$0.044)} & \makecell{0.776 \\ \scriptsize ($\pm$0.076)} & \makecell{0.699 \\ \scriptsize ($\pm$0.060)} & \makecell{0.818 \\ \scriptsize ($\pm$0.041)} & \makecell{0.879 \\ \scriptsize ($\pm$0.040)} & \makecell{0.935 \\ \scriptsize ($\pm$0.005)} & \makecell{\blue{0.941} \\ \scriptsize\blue{($\pm$0.105)}} & \makecell{\red{0.986} \\ \scriptsize\red{($\pm$0.044)}} & \makecell{\red{0.986} \\ \scriptsize\red{($\pm$0.044)}} \\
& NMI & \makecell{0.827 \\ \scriptsize ($\pm$0.068)} & \makecell{0.806 \\ \scriptsize ($\pm$0.070)} & \makecell{0.809 \\ \scriptsize ($\pm$0.065)} & \makecell{0.858 \\ \scriptsize ($\pm$0.053)} & \makecell{0.759 \\ \scriptsize ($\pm$0.096)} & \makecell{0.590 \\ \scriptsize ($\pm$0.083)} & \makecell{0.817 \\ \scriptsize ($\pm$0.053)} & \makecell{0.849 \\ \scriptsize ($\pm$0.061)} & \makecell{\blue{0.976} \\ \scriptsize\blue{($\pm$0.010)}} & \makecell{0.948 \\ \scriptsize ($\pm$0.095)} & \makecell{\red{0.988} \\ \scriptsize\red{($\pm$0.038)}} & \makecell{\red{0.988} \\ \scriptsize\red{($\pm$0.038)}} \\
& PUR & \makecell{0.848 \\ \scriptsize ($\pm$0.064)} & \makecell{0.822 \\ \scriptsize ($\pm$0.076)} & \makecell{0.835 \\ \scriptsize ($\pm$0.068)} & \makecell{0.872 \\ \scriptsize ($\pm$0.044)} & \makecell{0.772 \\ \scriptsize ($\pm$0.089)} & \makecell{0.685 \\ \scriptsize ($\pm$0.065)} & \makecell{0.815 \\ \scriptsize ($\pm$0.036)} & \makecell{0.877 \\ \scriptsize ($\pm$0.046)} & \makecell{\blue{0.945} \\ \scriptsize\blue{($\pm$0.005)}} & \makecell{0.933 \\ \scriptsize ($\pm$0.117)} & \makecell{\red{0.983} \\ \scriptsize\red{($\pm$0.053)}} & \makecell{\red{0.983} \\ \scriptsize\red{($\pm$0.053)}} \\ \hline
\multirow{4}{*}[-3.5ex]{CAL101}
       & ACC & \makecell{0.386 \\ \scriptsize ($\pm$0.089)} & \makecell{0.376 \\ \scriptsize ($\pm$0.071)} & \makecell{0.336 \\ \scriptsize ($\pm$0.070)} & \makecell{0.398 \\ \scriptsize ($\pm$0.076)} & \makecell{0.374 \\ \scriptsize ($\pm$0.038)} & \makecell{0.275 \\ \scriptsize ($\pm$0.043)} & \makecell{0.344 \\ \scriptsize ($\pm$0.071)} & \makecell{0.310 \\ \scriptsize ($\pm$0.089)} & \makecell{0.709 \\ \scriptsize ($\pm$0.080)} & \makecell{\blue{0.867} \\ \scriptsize\blue{($\pm$0.182)}} & \makecell{\red{0.932} \\ \scriptsize\red{($\pm$0.090)}} & \makecell{\red{0.932} \\ \scriptsize\red{($\pm$0.090)}} \\
& F1 & \makecell{0.407 \\ \scriptsize ($\pm$0.090)} & \makecell{0.394 \\ \scriptsize ($\pm$0.071)} & \makecell{0.365 \\ \scriptsize ($\pm$0.072)} & \makecell{0.412 \\ \scriptsize ($\pm$0.073)} & \makecell{0.405 \\ \scriptsize ($\pm$0.054)} & \makecell{0.295 \\ \scriptsize ($\pm$0.042)} & \makecell{0.368 \\ \scriptsize ($\pm$0.070)} & \makecell{0.438 \\ \scriptsize ($\pm$0.072)} & \makecell{0.725 \\ \scriptsize ($\pm$0.061)} & \makecell{\blue{0.893} \\ \scriptsize\blue{($\pm$0.147)}} & \makecell{\red{0.948} \\ \scriptsize\red{($\pm$0.067)}} & \makecell{\red{0.948} \\ \scriptsize\red{($\pm$0.067)}} \\
& NMI & \makecell{0.208 \\ \scriptsize ($\pm$0.111)} & \makecell{0.196 \\ \scriptsize ($\pm$0.095)} & \makecell{0.165 \\ \scriptsize ($\pm$0.088)} & \makecell{0.237 \\ \scriptsize ($\pm$0.098)} & \makecell{0.234 \\ \scriptsize ($\pm$0.077)} & \makecell{0.098 \\ \scriptsize ($\pm$0.037)} & \makecell{0.180 \\ \scriptsize ($\pm$0.082)} & \makecell{0.212 \\ \scriptsize ($\pm$0.097)} & \makecell{0.584 \\ \scriptsize ($\pm$0.041)} & \makecell{\blue{0.895} \\ \scriptsize\blue{($\pm$0.144)}} & \makecell{\red{0.950} \\ \scriptsize\red{($\pm$0.064)}} & \makecell{\red{0.950} \\ \scriptsize\red{($\pm$0.064)}} \\
& PUR & \makecell{0.395 \\ \scriptsize ($\pm$0.087)} & \makecell{0.383 \\ \scriptsize ($\pm$0.069)} & \makecell{0.358 \\ \scriptsize ($\pm$0.070)} & \makecell{0.409 \\ \scriptsize ($\pm$0.070)} & \makecell{0.397 \\ \scriptsize ($\pm$0.049)} & \makecell{0.289 \\ \scriptsize ($\pm$0.044)} & \makecell{0.368 \\ \scriptsize ($\pm$0.073)} & \makecell{0.439 \\ \scriptsize ($\pm$0.070)} & \makecell{0.760 \\ \scriptsize ($\pm$0.053)} & \makecell{\blue{0.883} \\ \scriptsize\blue{($\pm$0.158)}} & \makecell{\red{0.937} \\ \scriptsize\red{($\pm$0.082)}} & \makecell{\red{0.937} \\ \scriptsize\red{($\pm$0.082)}} \\ \hline \hline
\multicolumn{2}{|c|}{Average} & \makecell{0.593 \\ \scriptsize ($\pm$0.082)} & \makecell{0.577 \\ \scriptsize ($\pm$0.076)} & \makecell{0.566 \\ \scriptsize ($\pm$0.072)} & \makecell{0.615 \\ \scriptsize ($\pm$0.065)} & \makecell{0.558 \\ \scriptsize ($\pm$0.075)} & \makecell{0.450 \\ \scriptsize ($\pm$0.057)} & \makecell{0.565 \\ \scriptsize ($\pm$0.060)} & \makecell{0.607 \\ \scriptsize ($\pm$0.067)} & \makecell{0.821 \\ \scriptsize ($\pm$0.033)} & \makecell{\blue{0.916} \\ \scriptsize\blue{($\pm$0.177)}} & \makecell{\red{0.963} \\ \scriptsize\red{($\pm$0.126)}} & \makecell{\red{0.963} \\ \scriptsize\red{($\pm$0.125)}} \\ \hline
\end{tabular}}
\end{table*}

\begin{figure*}[htbp!]
\makeatletter
\renewcommand{\@thesubfigure}{\hskip\subfiglabelskip}
  \centering
\subfigure[~~~~~~~~(a) ACC (UMIST)]{
    \includegraphics[width=1.8 in]{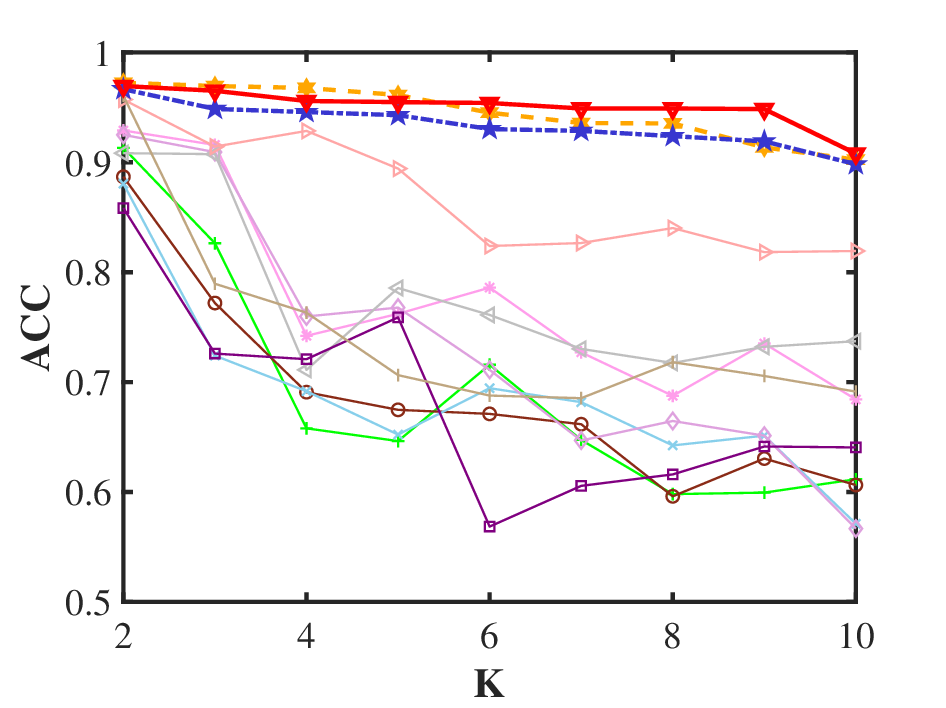}\hspace{-5mm}}
\subfigure[~~~~~~~~(b) FSC (UMIST)]{
    \includegraphics[width=1.8 in]{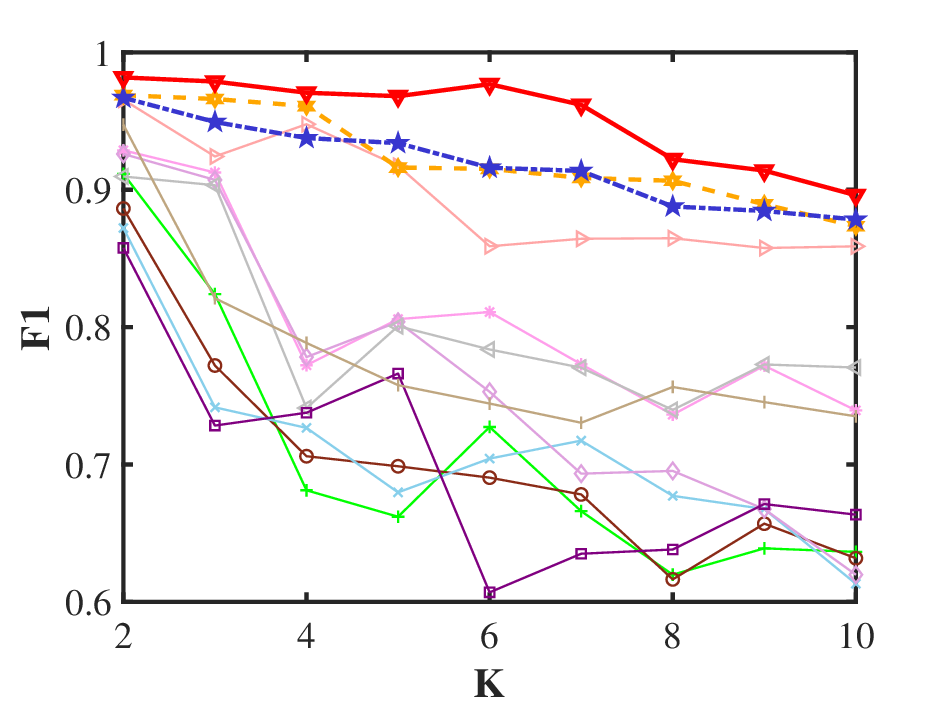}\hspace{-5mm}}
\subfigure[~~~~~~~~(c) NMI (UMIST)]{
    \includegraphics[width=1.8 in]{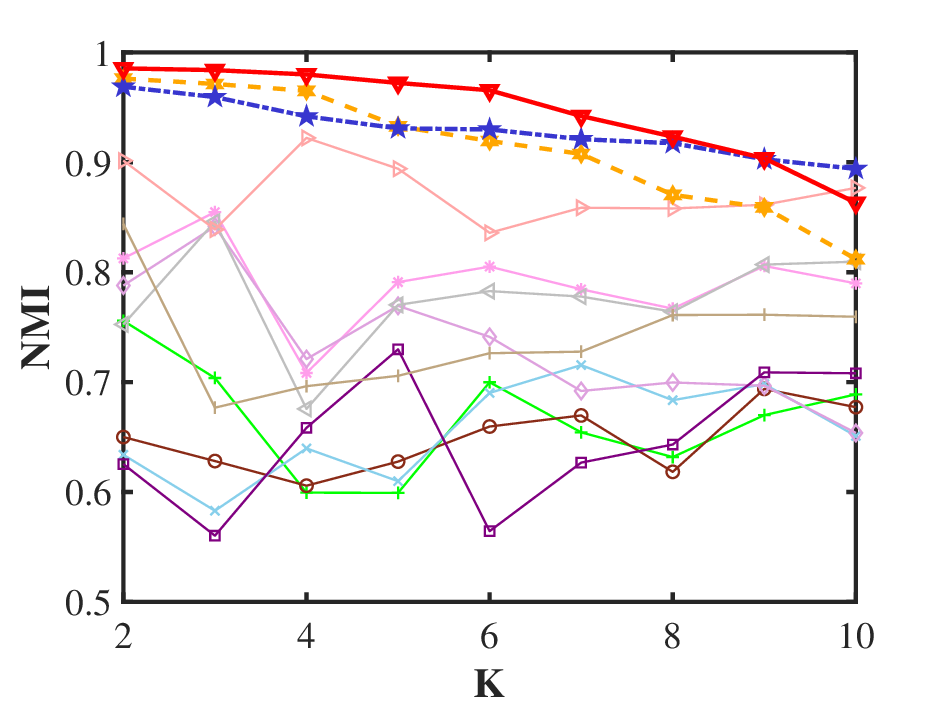}\hspace{-5mm}}
\subfigure[~~~~~~~~(d) PUR (UMIST)]{
    \includegraphics[width=1.8 in]{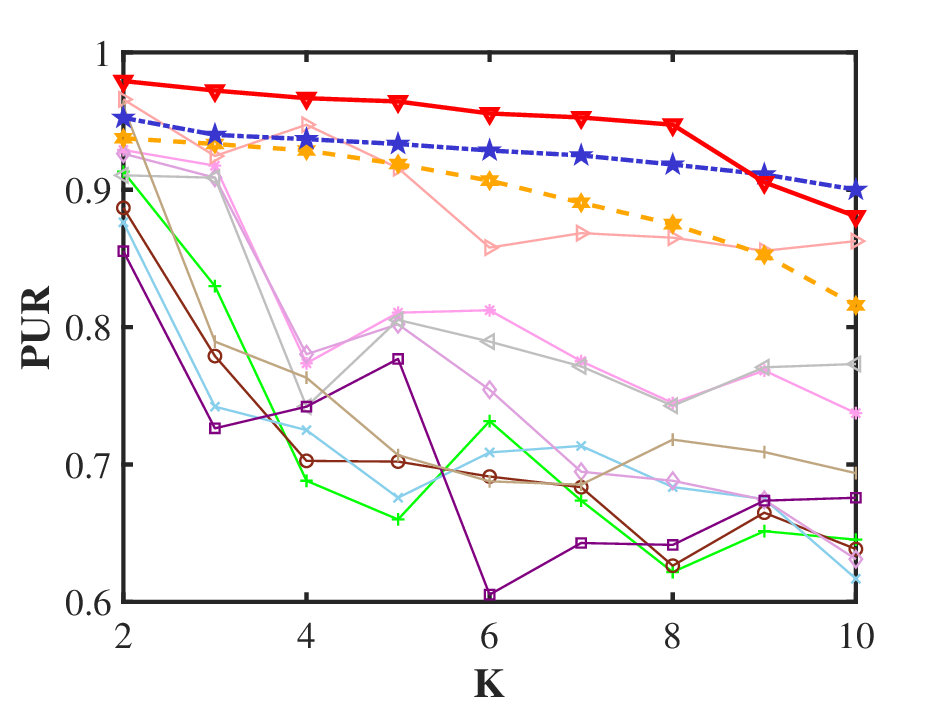}\hspace{-5mm}}
  \subfigure[~~~~~~~~(e) ACC (YALE)]{
    \includegraphics[width=1.8 in]{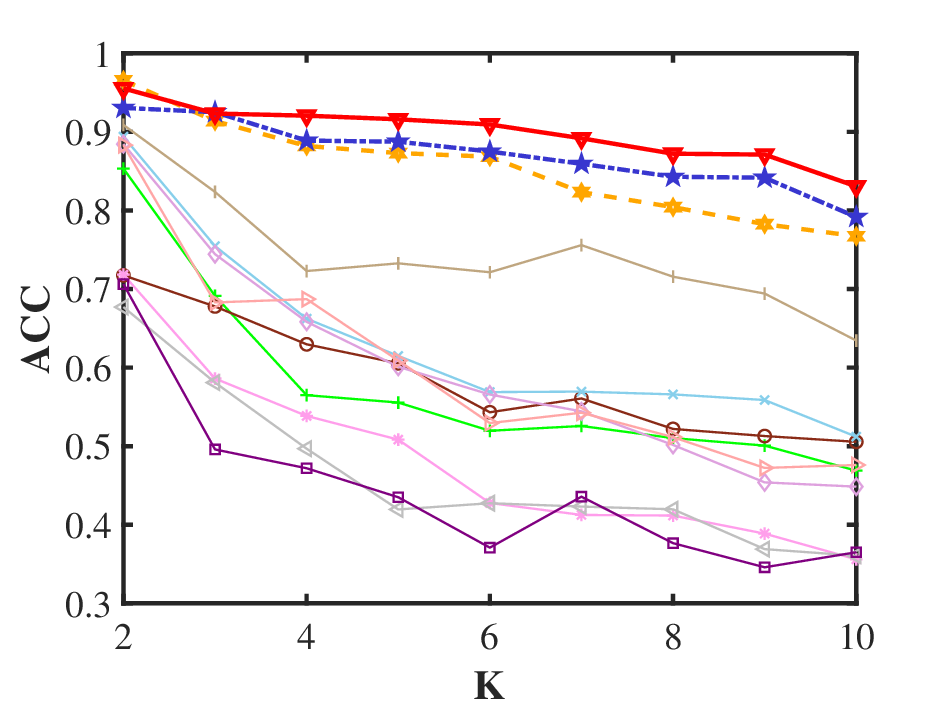}\hspace{-5mm}}
\subfigure[~~~~~~~~(f) FSC (YALE)]{
    \includegraphics[width=1.8 in]{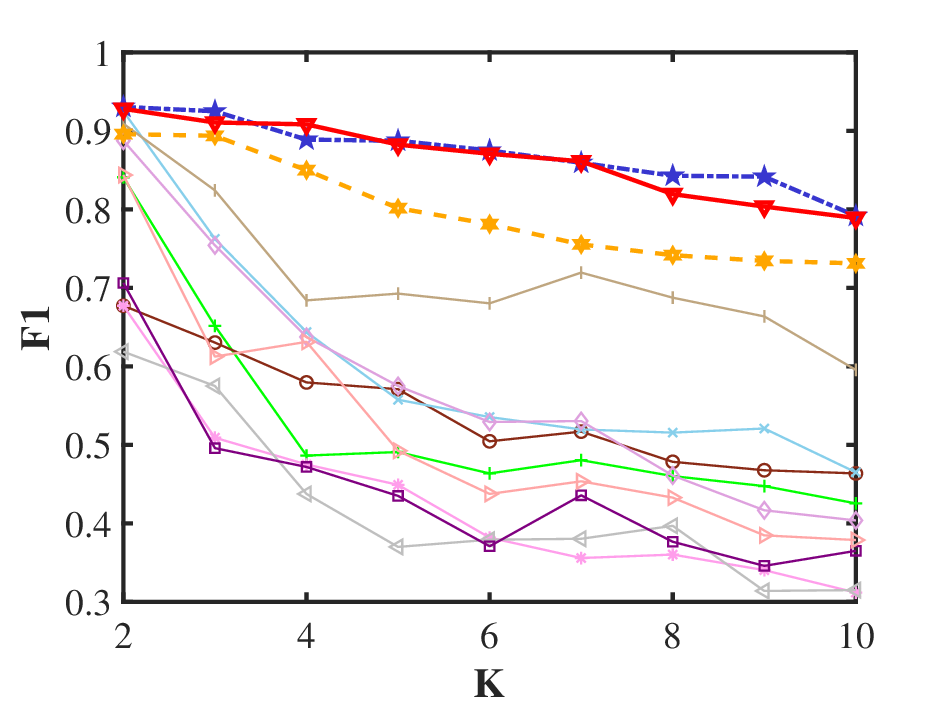}\hspace{-5mm}}
\subfigure[~~~~~~~~(g) NMI (YALE)]{
    \includegraphics[width=1.8 in]{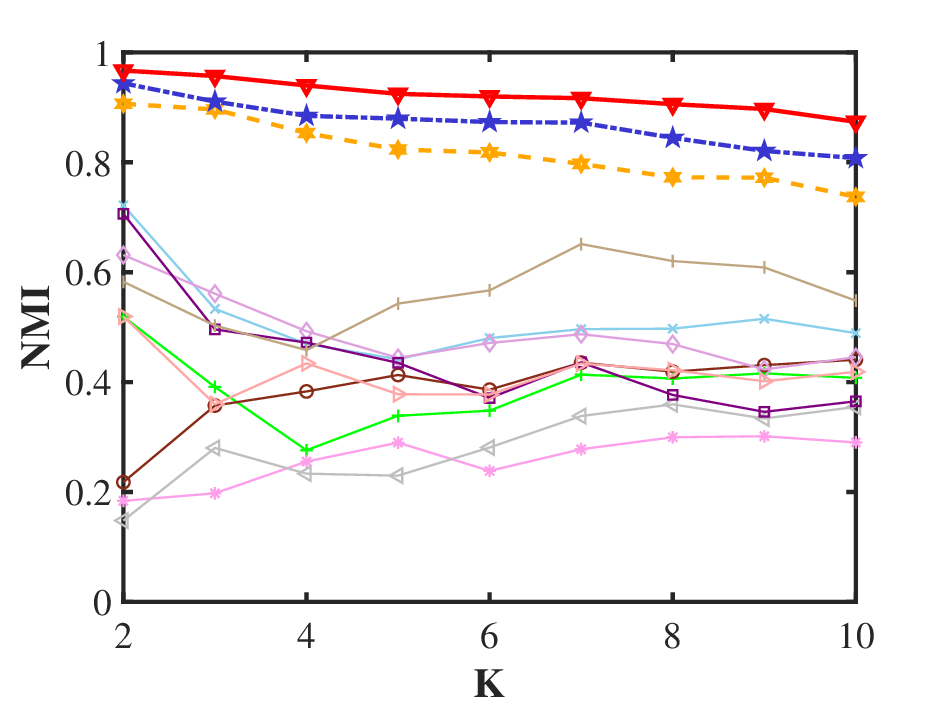}\hspace{-5mm}}
\subfigure[~~~~~~~~(h) PUR (YALE)]{
    \includegraphics[width=1.8 in]{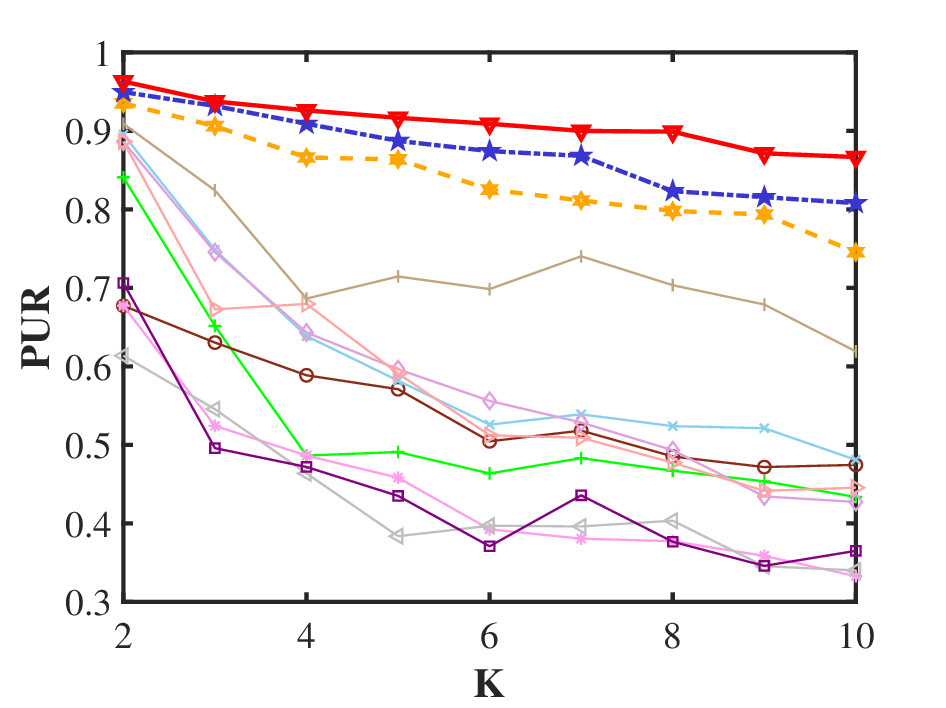}\hspace{-5mm}}
  \subfigure[~~~~~~~~(i) ACC (COIL20)]{
    \includegraphics[width=1.8 in]{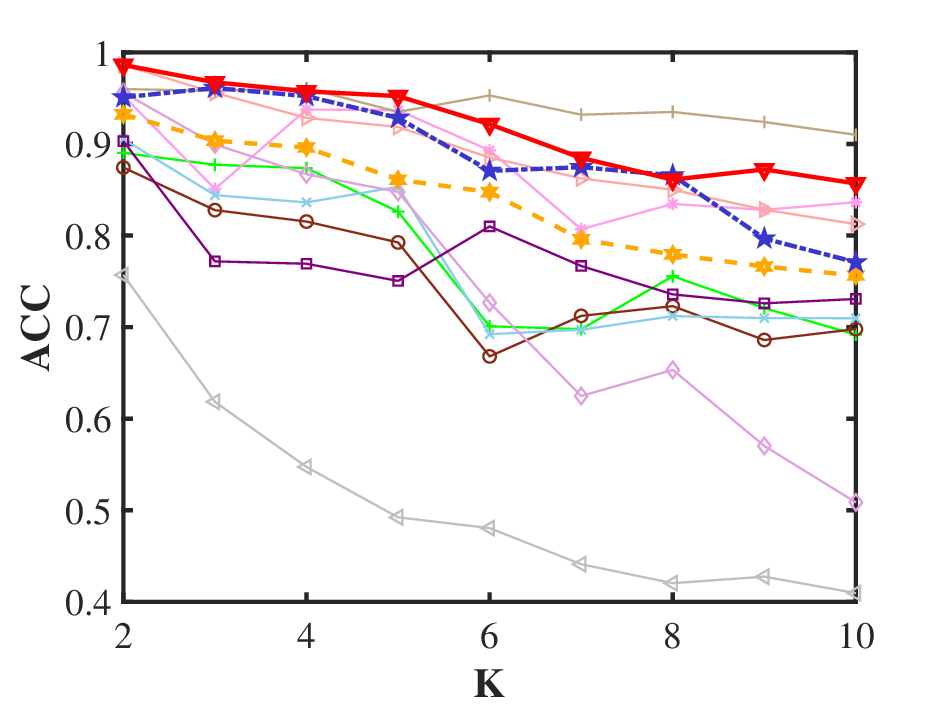}\hspace{-5mm}}
\subfigure[~~~~~~~~(j) FSC (COIL20)]{
    \includegraphics[width=1.8 in]{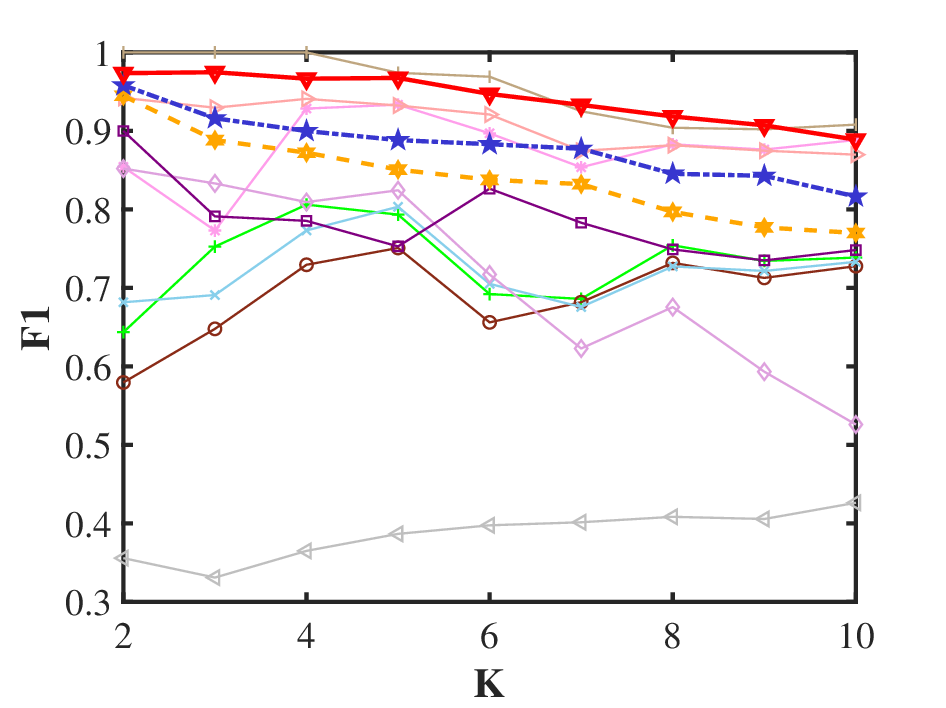}\hspace{-5mm}}
\subfigure[~~~~~~~~(k) NMI (COIL20)]{
    \includegraphics[width=1.8 in]{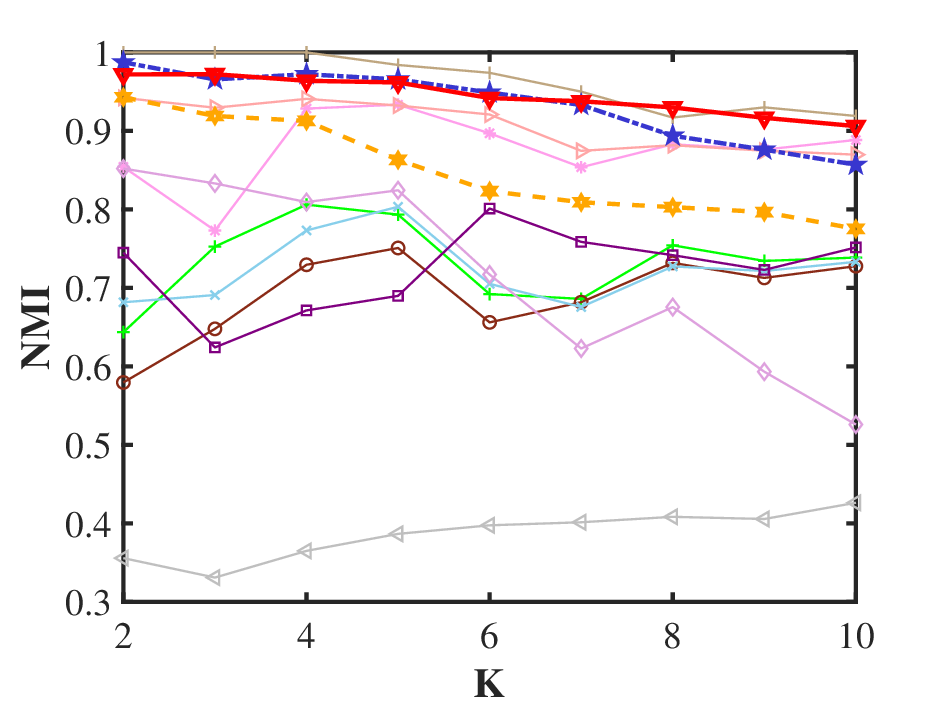}\hspace{-5mm}}
\subfigure[~~~~~~~~(l) PUR (COIL20)]{
    \includegraphics[width=1.8 in]{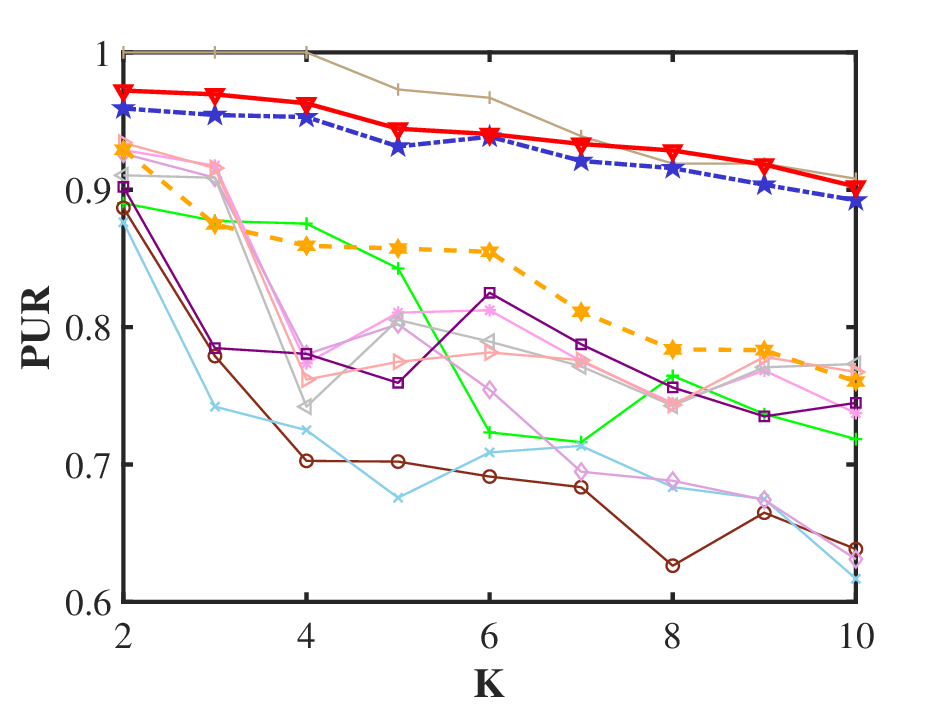}\hspace{-5mm}}
  \subfigure[~~~~~~~~(m) ACC (COIL100)]{
    \includegraphics[width=1.8 in]{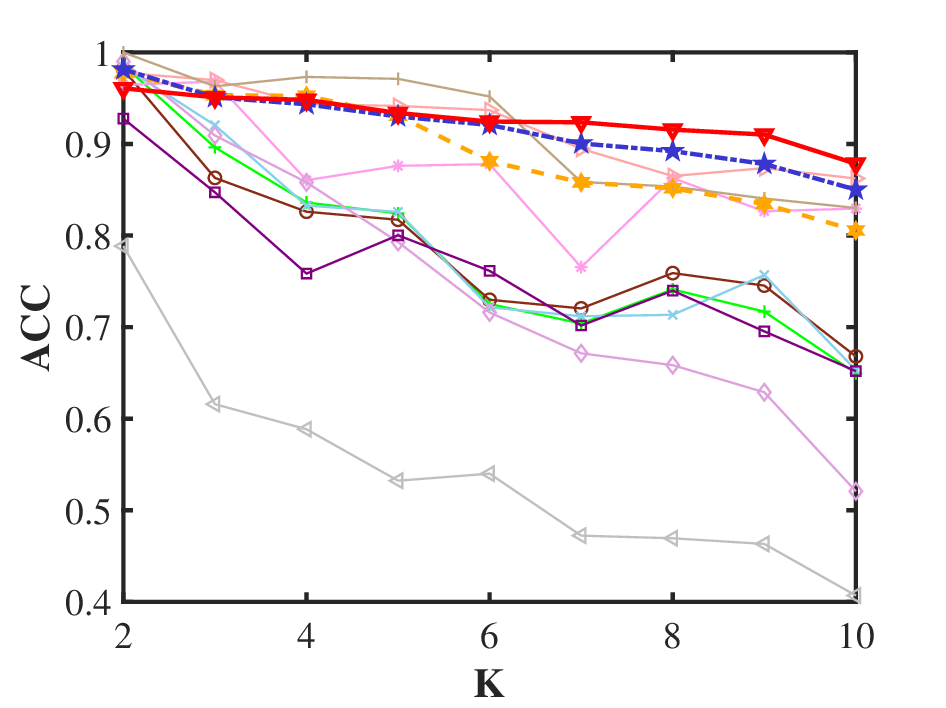}\hspace{-5mm}}
\subfigure[~~~~~~~~(n) FSC (COIL100)]{
    \includegraphics[width=1.8 in]{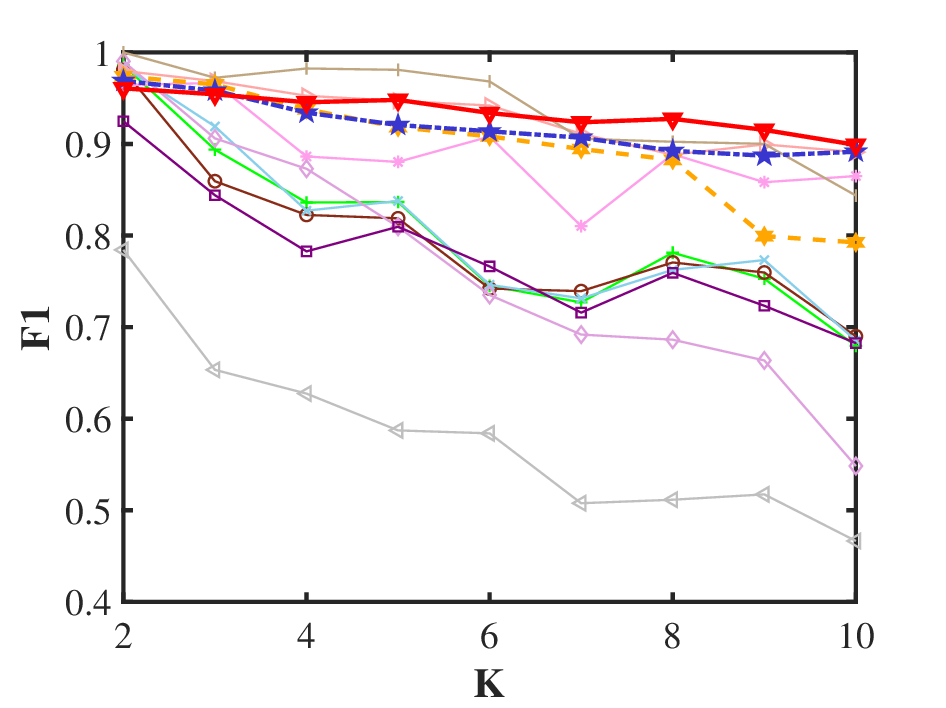}\hspace{-5mm}}
\subfigure[~~~~~~~~(o) NMI (COIL100)]{
    \includegraphics[width=1.8 in]{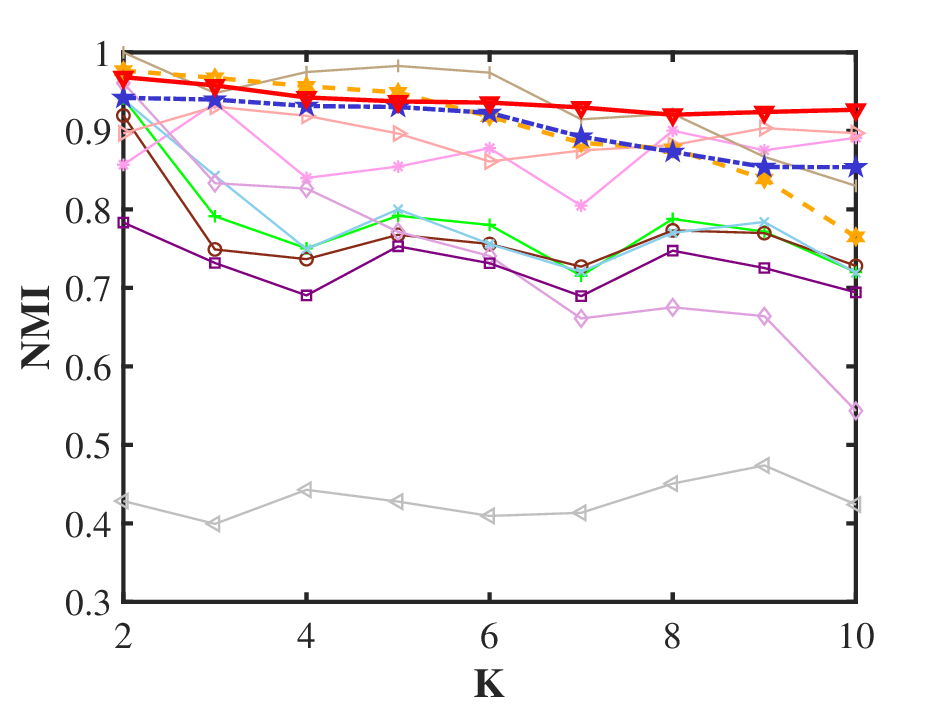}\hspace{-5mm}}
\subfigure[~~~~~~~~(p) PUR (COIL100)]{
    \includegraphics[width=1.8 in]{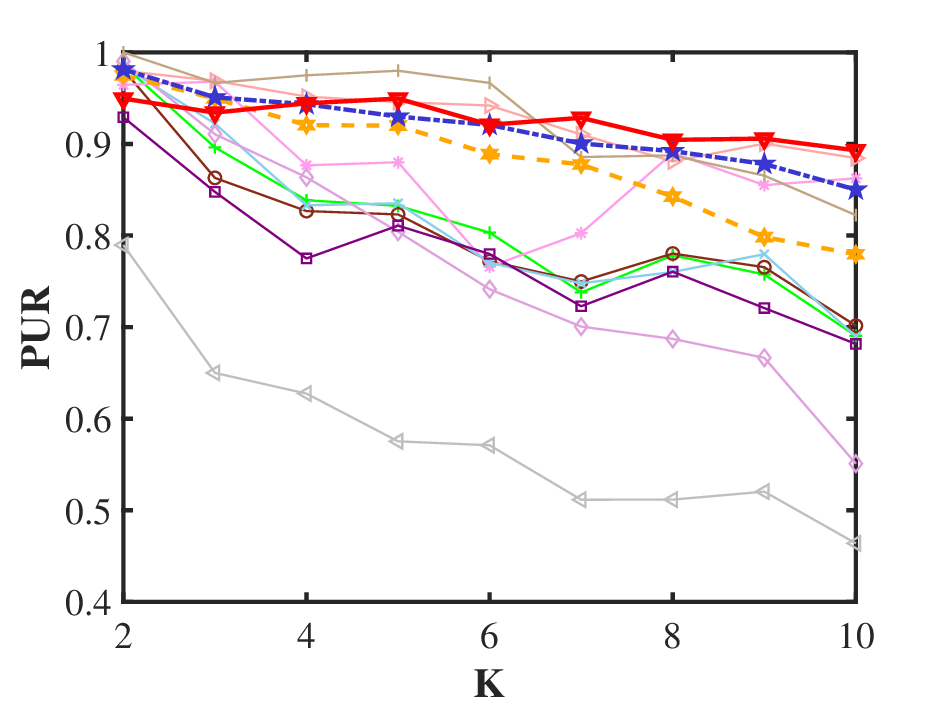}\hspace{-5mm}}
   \subfigure[]{\label{}
    \includegraphics[width=6 in]{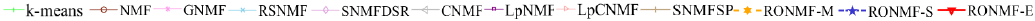}}
  \vskip-0.2cm
  \caption{All metrics for different clustering numbers $k$ on Type-\uppercase\expandafter{\romannumeral1} and Type-\uppercase\expandafter{\romannumeral2} datasets, where the horizontal axis denotes $k$, and the vertical axis denotes the value of indicators.}
  \label{zxtclean1}
\end{figure*}

\begin{figure*}[htbp!]
\makeatletter
\renewcommand{\@thesubfigure}{\hskip\subfiglabelskip}
\makeatother
\centering
\subfigure[~~~~~~~~(a) ACC (USPS)]{
    \includegraphics[width=1.8 in]{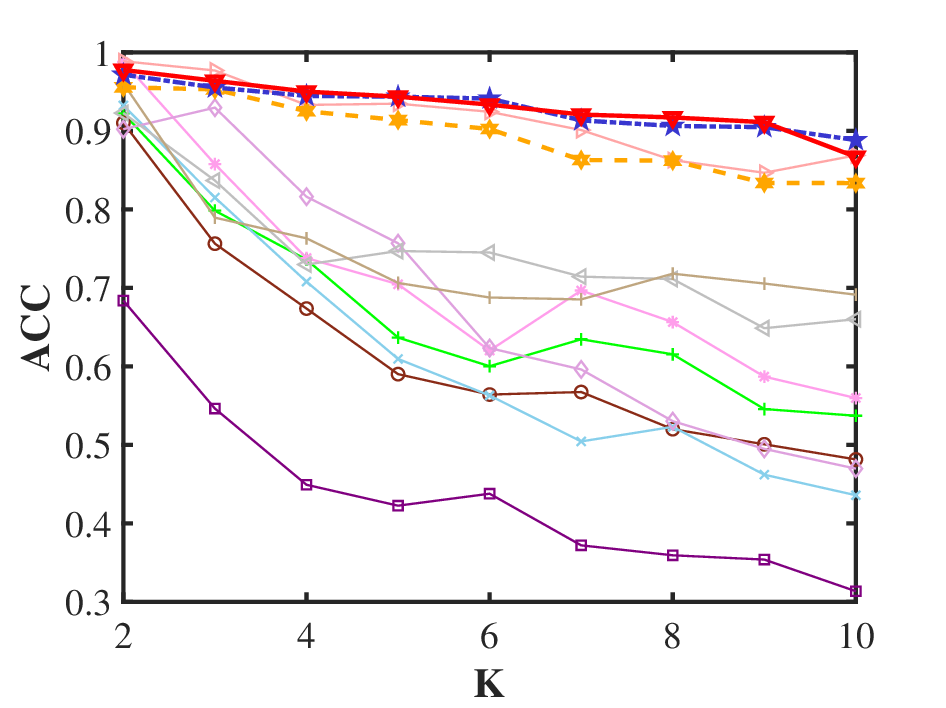}\hspace{-5mm}}
\subfigure[~~~~~~~~(b) F1 (USPS)]{
    \includegraphics[width=1.8 in]{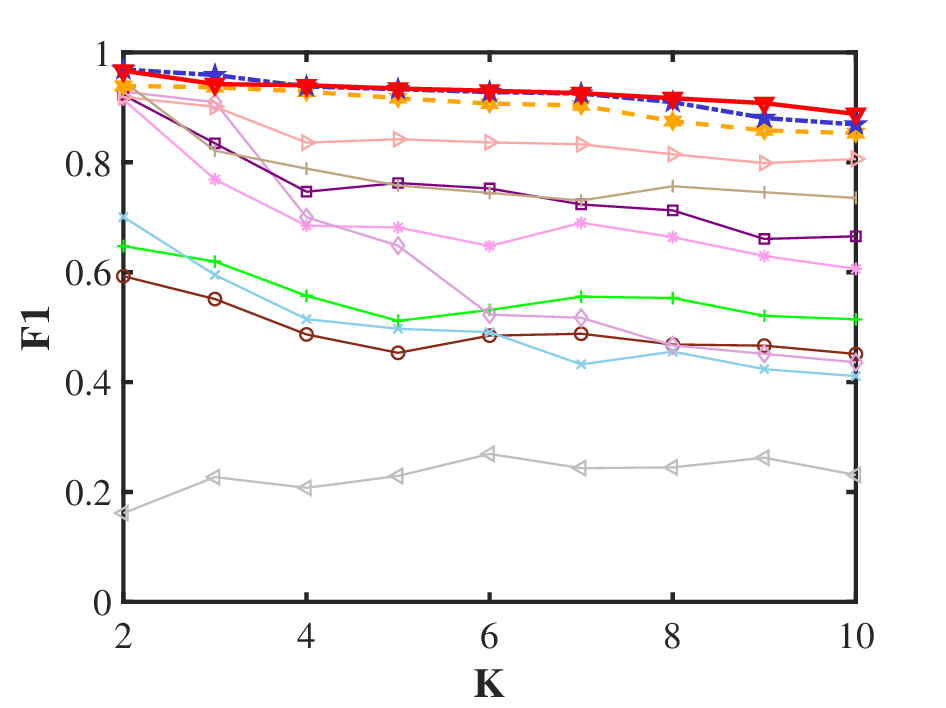}\hspace{-5mm}}
\subfigure[~~~~~~~~(c) NMI (USPS)]{
    \includegraphics[width=1.8 in]{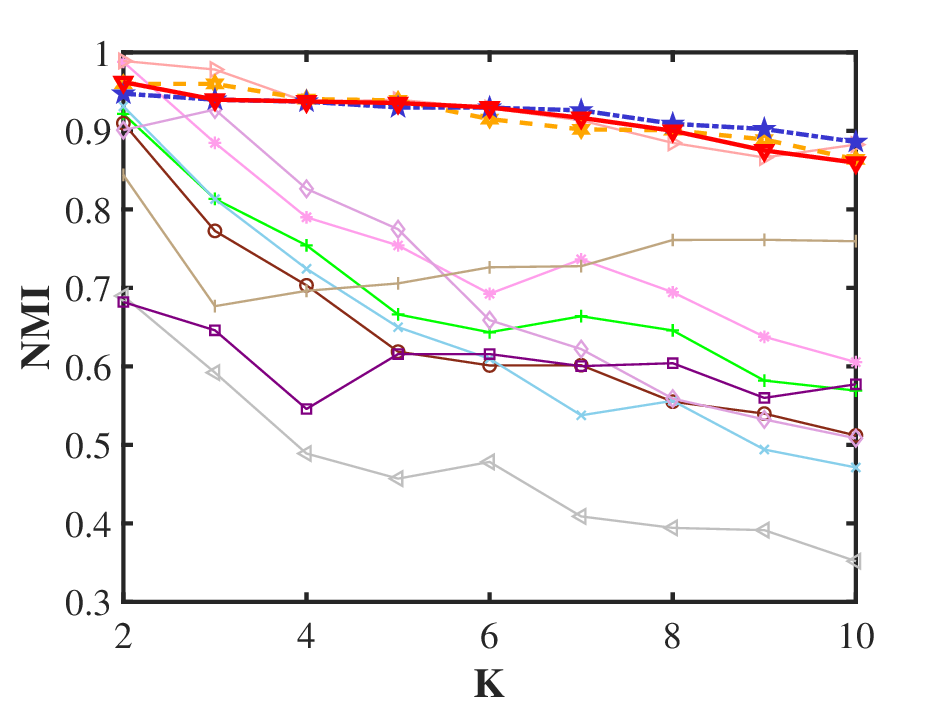}\hspace{-5mm}}
\subfigure[~~~~~~~~(d) PUR (USPS)]{
    \includegraphics[width=1.8 in]{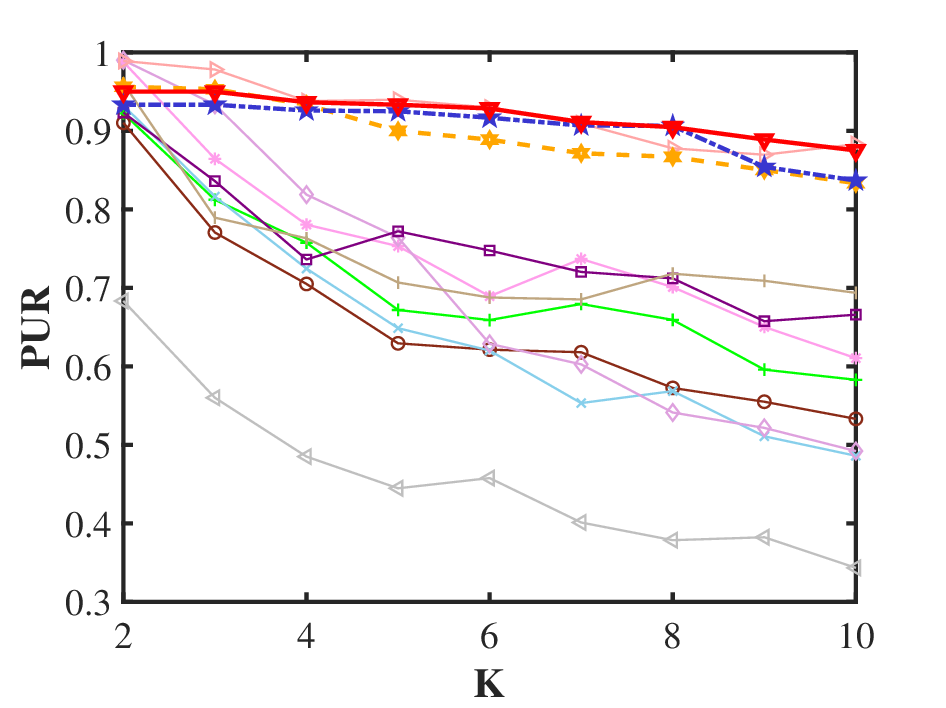}\hspace{-5mm}}
  \subfigure[~~~~~~~~(e) ACC (MNIST)]{
    \includegraphics[width=1.8 in]{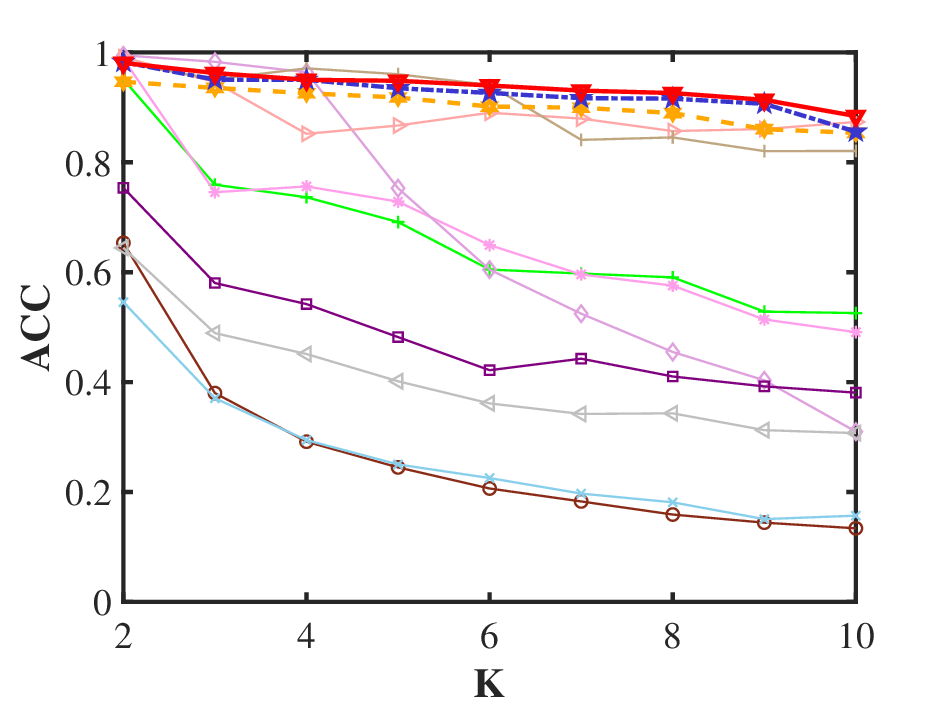}\hspace{-5mm}}
\subfigure[~~~~~~~~(f) F1 (MNIST)]{
    \includegraphics[width=1.8 in]{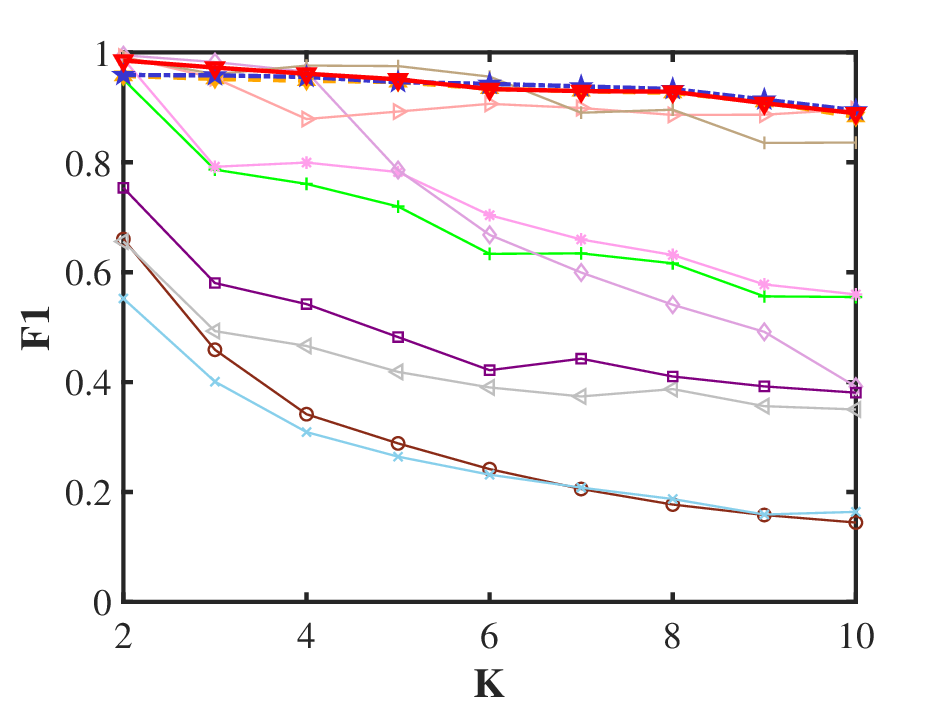}\hspace{-5mm}}
\subfigure[~~~~~~~~(g) NMI (MNIST)]{
    \includegraphics[width=1.8 in]{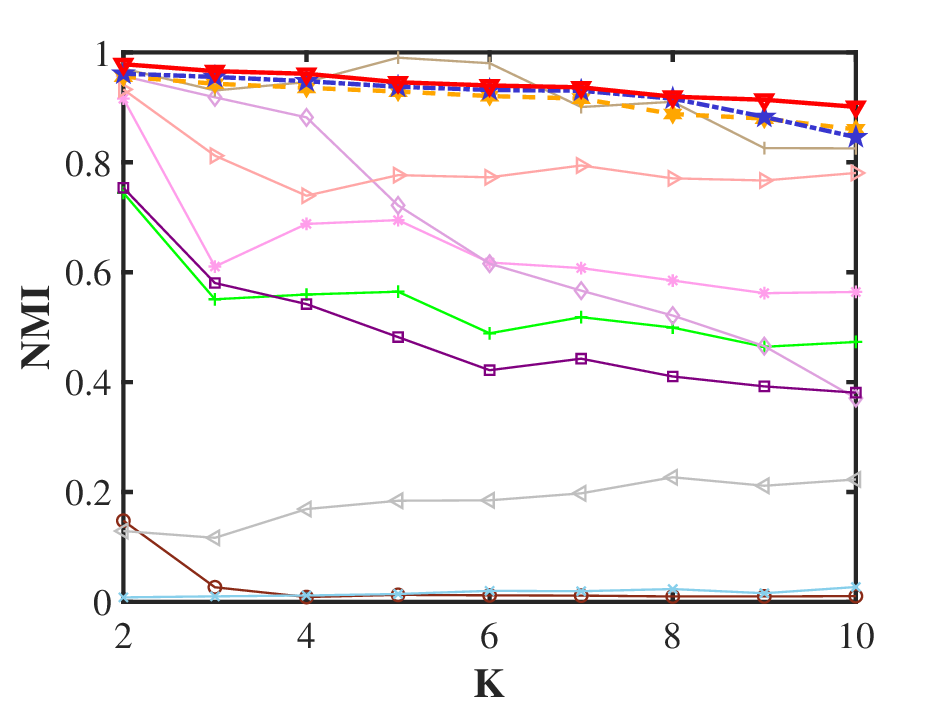}\hspace{-5mm}}
\subfigure[~~~~~~~~(h) PUR (MNIST)]{
    \includegraphics[width=1.8 in]{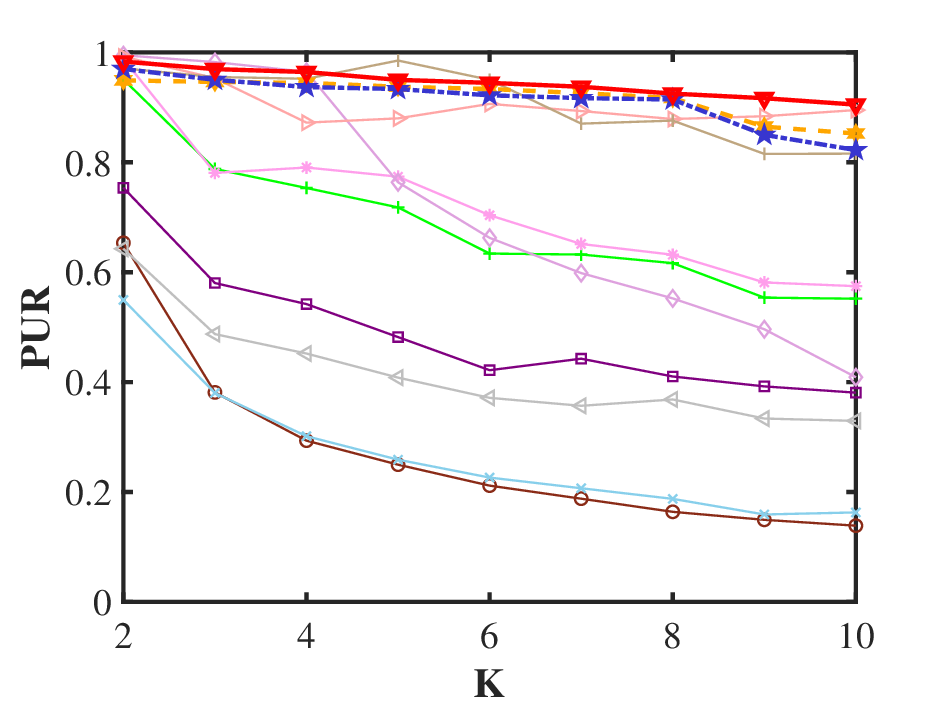}\hspace{-5mm}}
  \subfigure[~~~~~~~~(i) ACC (ORL10P)]{
    \includegraphics[width=1.8 in]{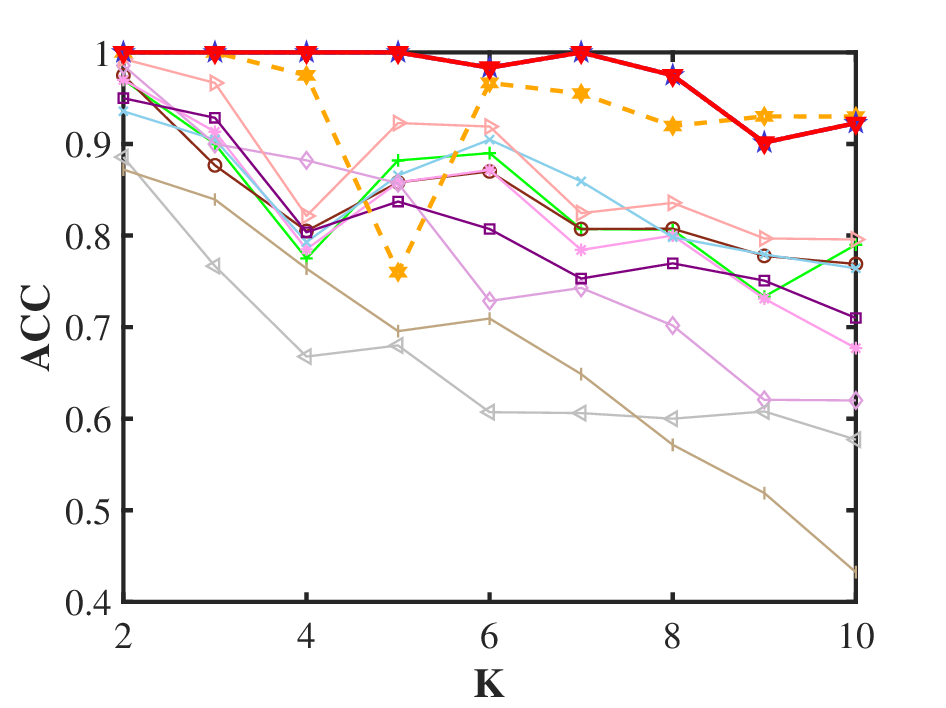}\hspace{-5mm}}
\subfigure[~~~~~~~~(j) F1 (ORL10P)]{
    \includegraphics[width=1.8 in]{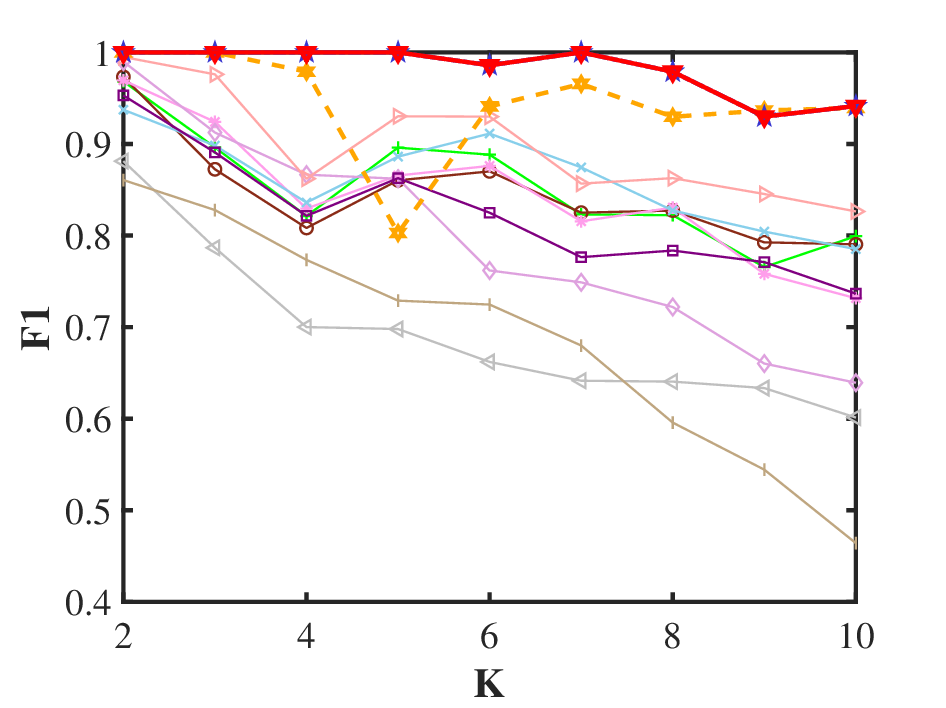}\hspace{-5mm}}
\subfigure[~~~~~~~~(k) NMI (ORL10P)]{
    \includegraphics[width=1.8 in]{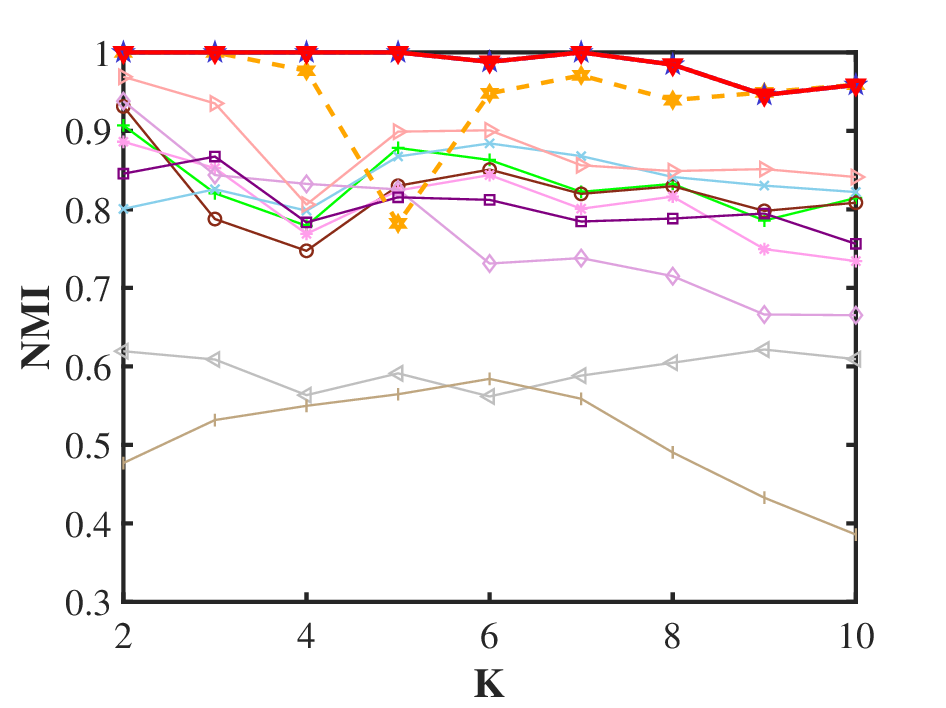}\hspace{-5mm}}
\subfigure[~~~~~~~~(l) PUR (ORL10P)]{
    \includegraphics[width=1.8 in]{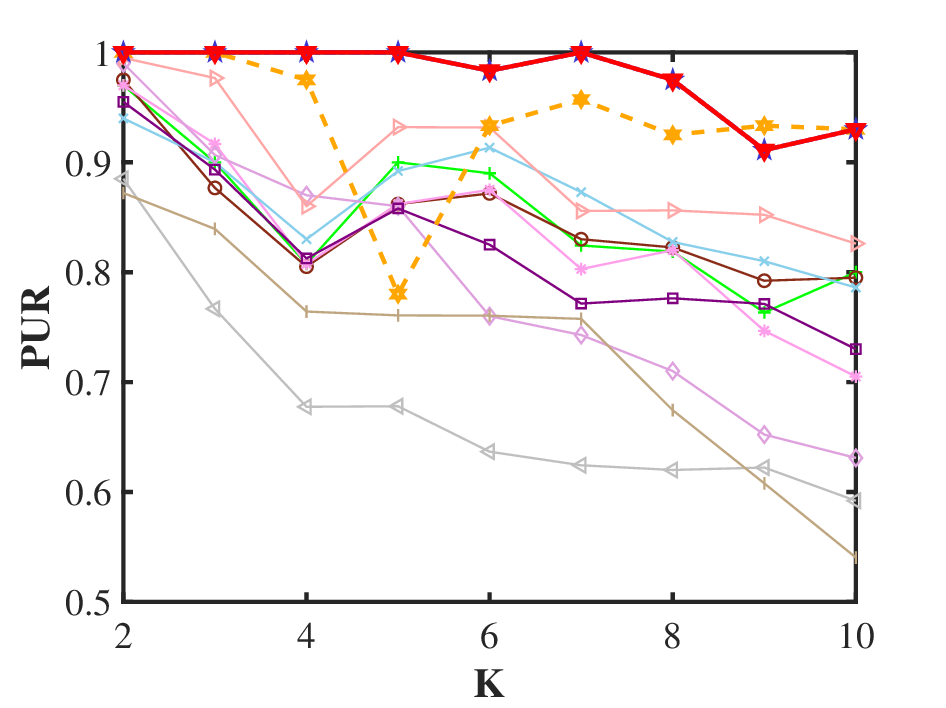}\hspace{-5mm}}
  \subfigure[~~~~~~~~(m) ACC (CAL101)]{
    \includegraphics[width=1.8 in]{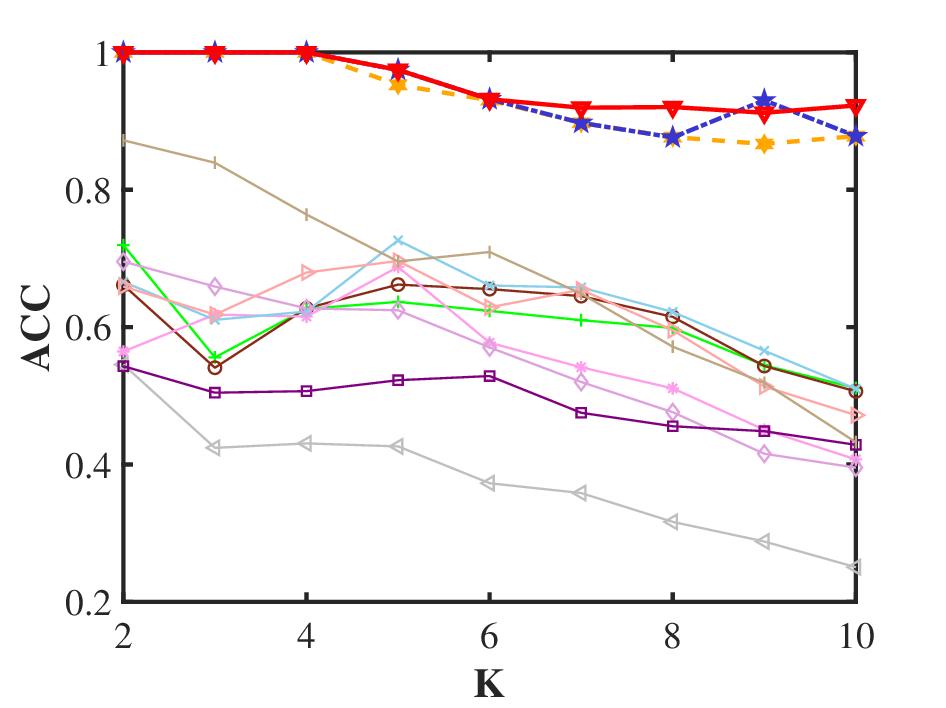}\hspace{-5mm}}
\subfigure[~~~~~~~~(n) F1 (CAL101)]{
    \includegraphics[width=1.8 in]{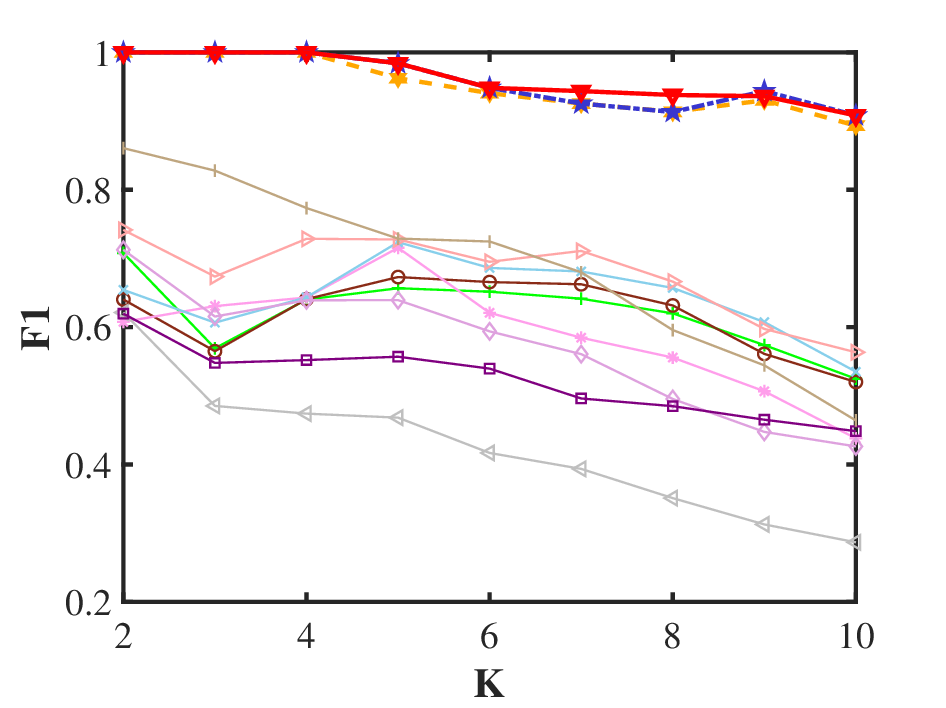}\hspace{-5mm}}
\subfigure[~~~~~~~~(o) NMI (CAL101)]{
    \includegraphics[width=1.8 in]{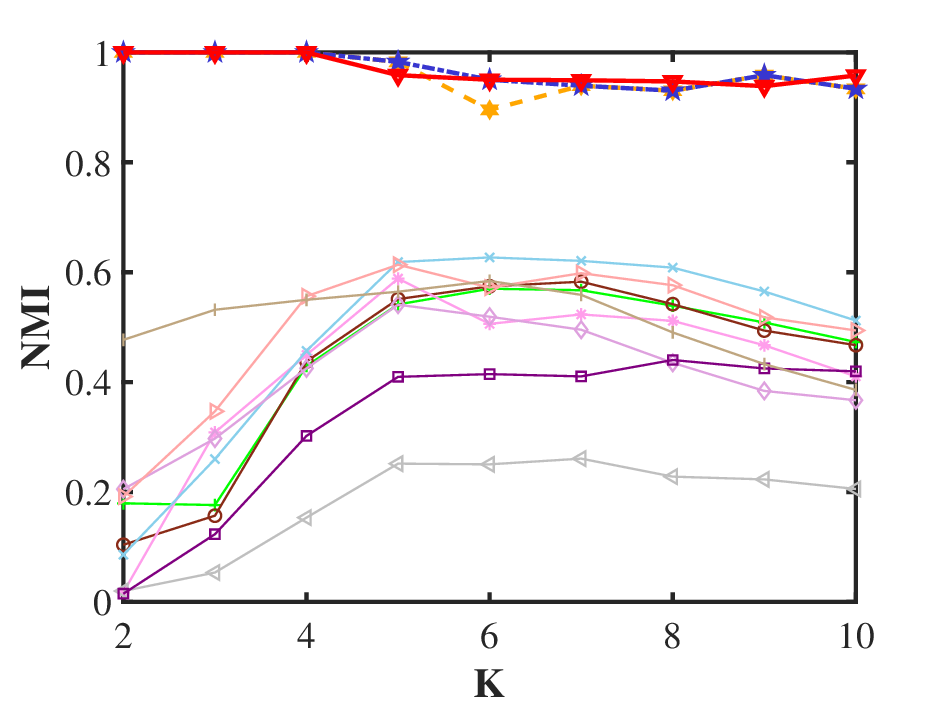}\hspace{-5mm}}
\subfigure[~~~~~~~~(p) PUR (CAL101)]{
    \includegraphics[width=1.8 in]{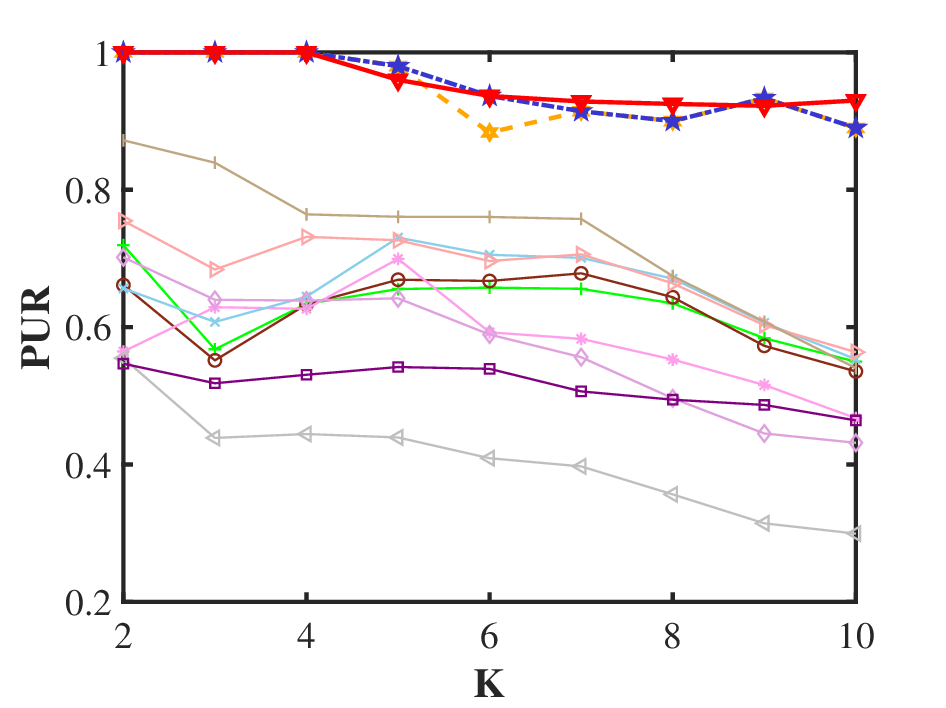}\hspace{-5mm}}
   \subfigure[]{\label{}
    \includegraphics[width=6 in]{cleantl.pdf}}
  \vskip-0.2cm
  \caption{All metrics for different clustering numbers $k$ on Type-\uppercase\expandafter{\romannumeral3} and Type-\uppercase\expandafter{\romannumeral4} datasets, where the horizontal axis denotes $k$, and the vertical axis denotes the value of indicators.}
  \label{zxtclean2}
\end{figure*}

\section{Numerical Experiments}\label{sec4}
This section compares the proposed RONMF with k-means \cite{liu2008reducing}, NMF \cite{lee1999learning}, GNMF \cite{cai2010graph}, RSNMF \cite{li2017robust}, SNMFDSR \cite{jia2019semi}, CNMF \cite{liu2011constrained},  LpNMF \cite{lan2020label}, LpCNMF \cite{liu2023constrained} and SNMFSP \cite{jing2025semi}.
Table \ref{datasets} provides details of the four types of datasets used to evaluate the performance of these methods.
\begin{itemize}
\item Type-\uppercase\expandafter{\romannumeral1}: This category includes UMIST\footnote{\href{https://www.visioneng.org.uk/datasets/}{https://www.visioneng.org.uk/datasets/}} and YALE\footnote{\href{http://cvc.cs.yale.edu/cvc/projects/yalefaces/yalefaces.html}{http://cvc.cs.yale.edu/cvc/projects/yalefaces/yalefaces.html}} face datasets, covering various races, genders, and appearances.
Facial images encompass intricate details and exhibit highly correlated structures, with features often demonstrating interdependencies.
\item Type-\uppercase\expandafter{\romannumeral2}: This category contains  COIL20\footnote{\href{http://www.cs.columbia.edu/CAVE/software/softlib/coil-20.php}{http://www.cs.columbia.edu/CAVE/software/softlib/coil-20.php}} and COIL100\footnote{\href{http://www.cs.columbia.edu/CAVE/software/softlib/coil-100.php}{http://www.cs.columbia.edu/CAVE/software/softlib/coil-100.php}} object datasets, which consist of images of various objects captured from different angles during a 360° rotation, including both the objects and the background. Object datasets typically exhibit well-defined structures and prominent edge features, and may also contain some redundant background information.
\item Type-\uppercase\expandafter{\romannumeral3}: This category includes USPS\footnote{\href{https://paperswithcode.com/dataset/usps}{https://paperswithcode.com/dataset/usps}} and MNIST\footnote{\href{https://yann.lecun.com/exdb/mnist/}{https://yann.lecun.com/exdb/mnist/}} handwritten datasets, which consist of images representing handwritten digits from 0 to 9. Handwritten numerical data exhibits a certain degree of variability due to the diverse writing styles of different individuals, however, the overall structure is still relatively simple and consistent.
\item Type-\uppercase\expandafter{\romannumeral4}:
This category includes orlraws10P\footnote{\url{http://www.vision.caltech.edu/Image_Datasets/Caltech101/}}  and CALTECH101\footnote{\href{https://www.cl.cam.ac.uk/research/dtg/attarchive/facedatabase.html}{https://www.cl.cam.ac.uk/research/dtg/attarchive/facedatabase.html}}.
In contrast to the other three types of datasets, this particular dataset is notable for its rich image features, with each individual image having a size exceeding 10,000. By incorporating these two datasets, insights into the performance of the NMF clustering method when handling large\-scale datasets can be effectively gained.

\end{itemize}

\begin{figure*}[ht!]
  \centering
    \makeatletter
    \renewcommand{\@thesubfigure}{\hskip\subfiglabelskip}
    \makeatother
    \subfigure[(a) ACC (YALE)]{\label{noisey}
    \includegraphics[width=1.6 in]{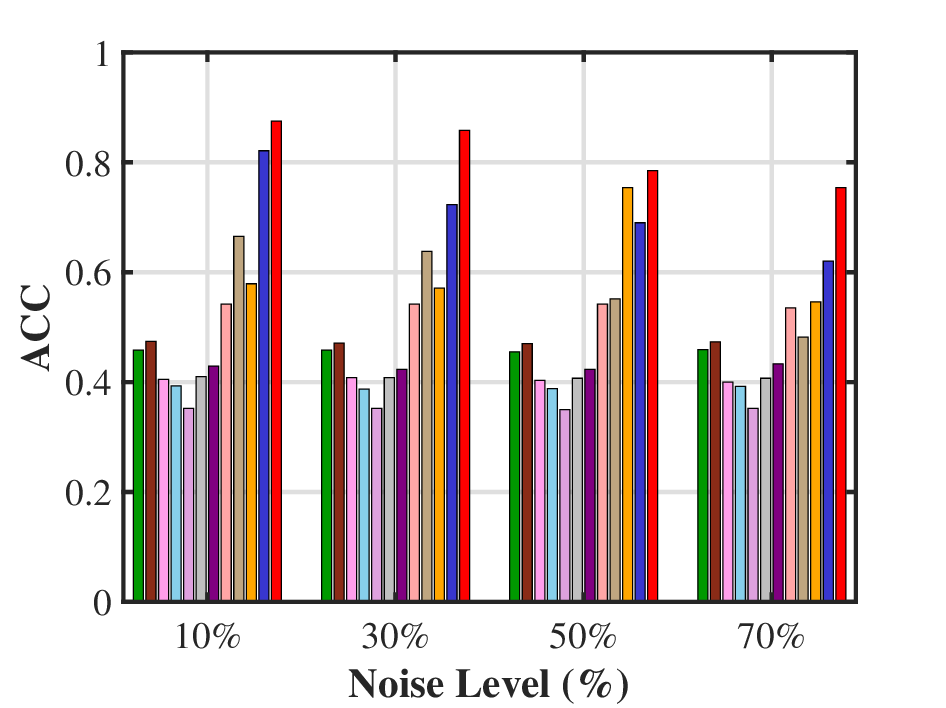}}
    \subfigure[(b) F1 (YALE)]{
    \includegraphics[width=1.6 in]{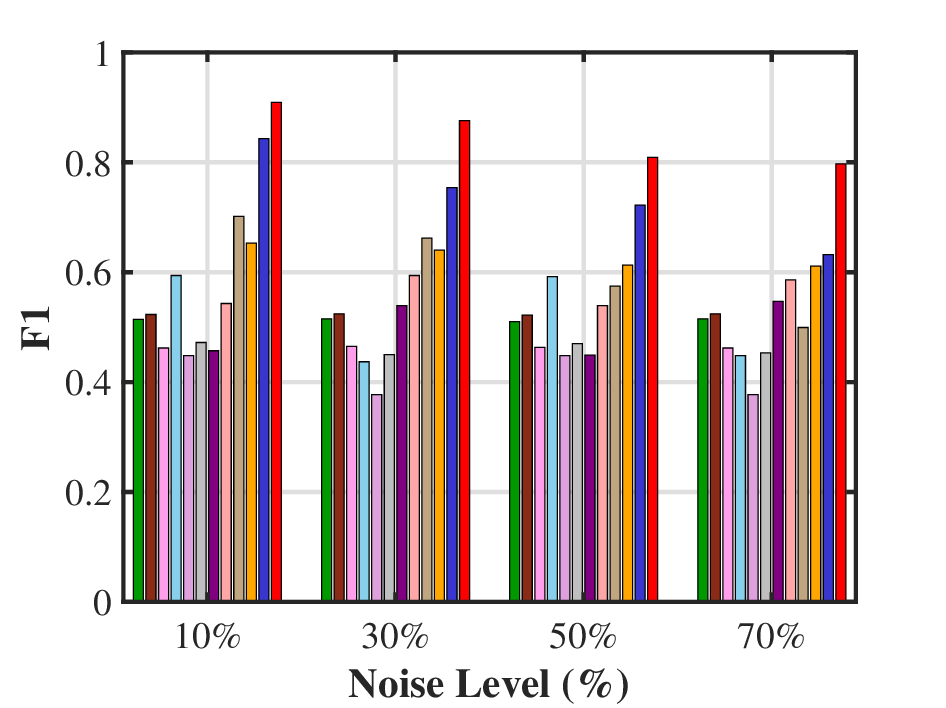}}
    \subfigure[(c) NMI (YALE)]{
    \includegraphics[width=1.6 in]{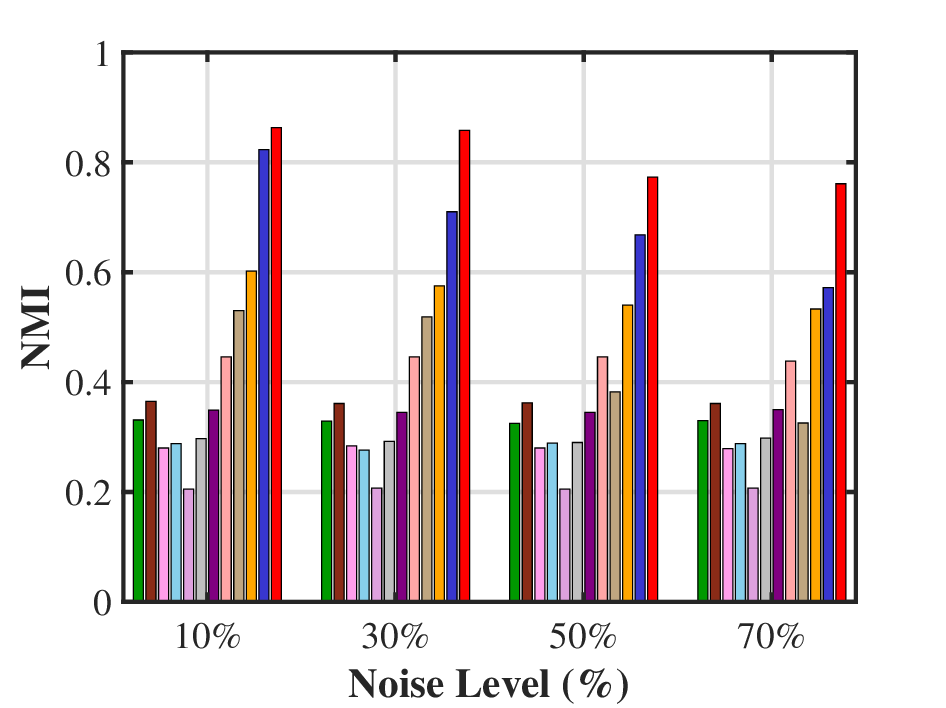}}
    \subfigure[(d) PUR (YALE)]{
    \includegraphics[width=1.6 in]{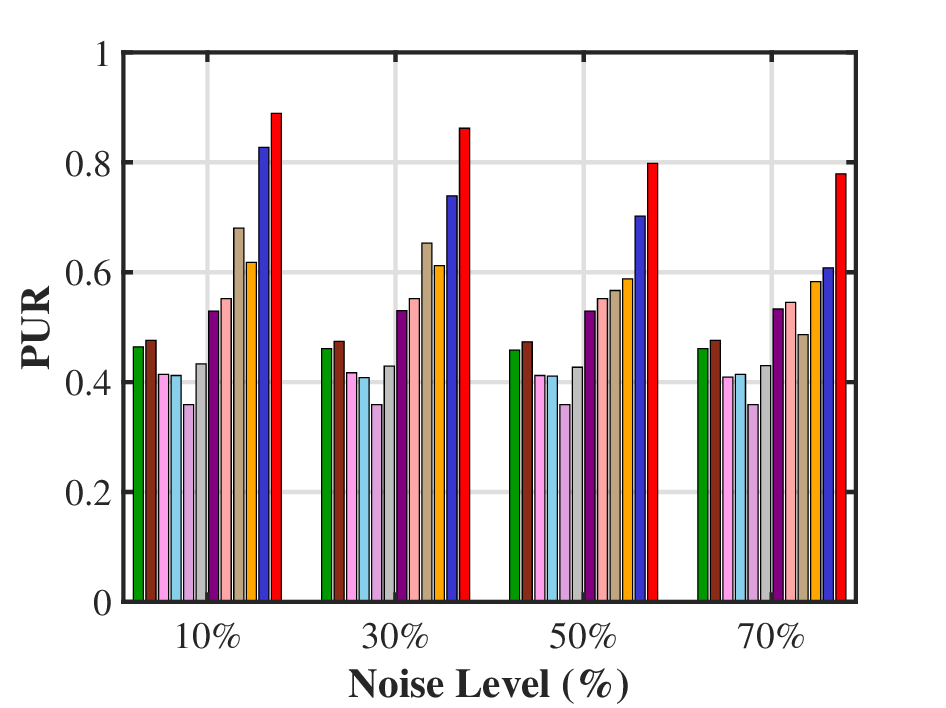}}
    \subfigure[(e) ACC (COIL20)]{\label{noisec}
    \includegraphics[width=1.6 in]{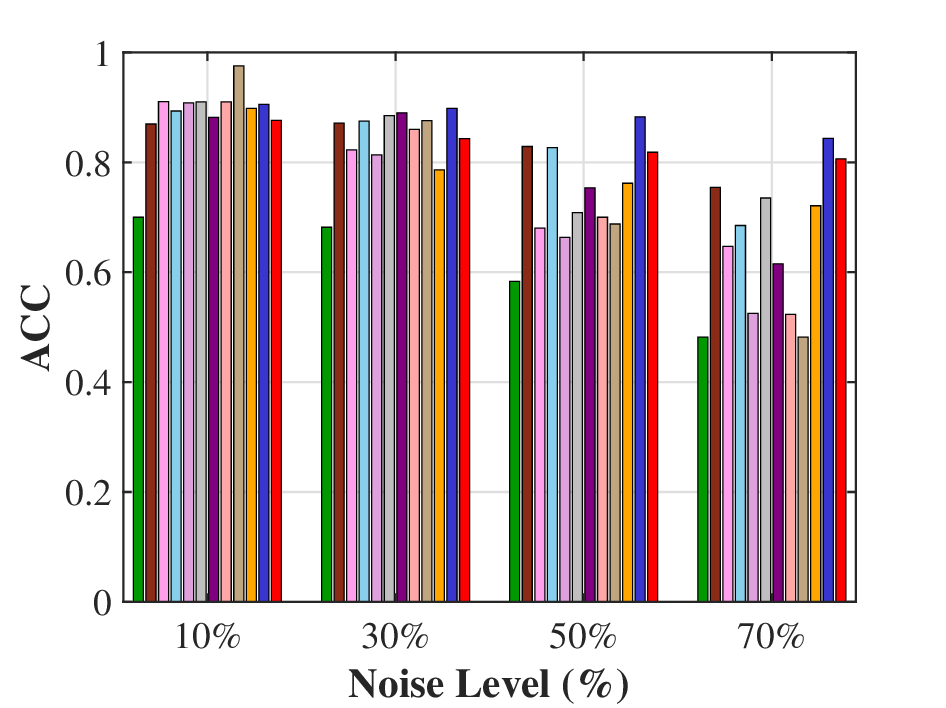}}
    \subfigure[(f) F1 (COIL20)]{
    \includegraphics[width=1.6 in]{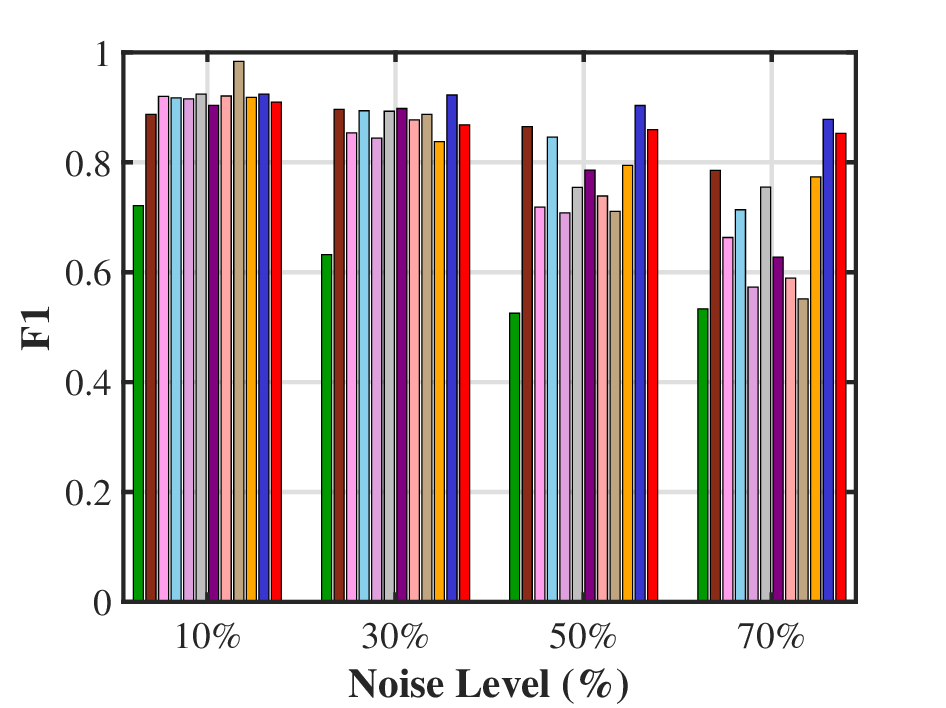}}
    \subfigure[(g) NMI (COIL20)]{
    \includegraphics[width=1.6 in]{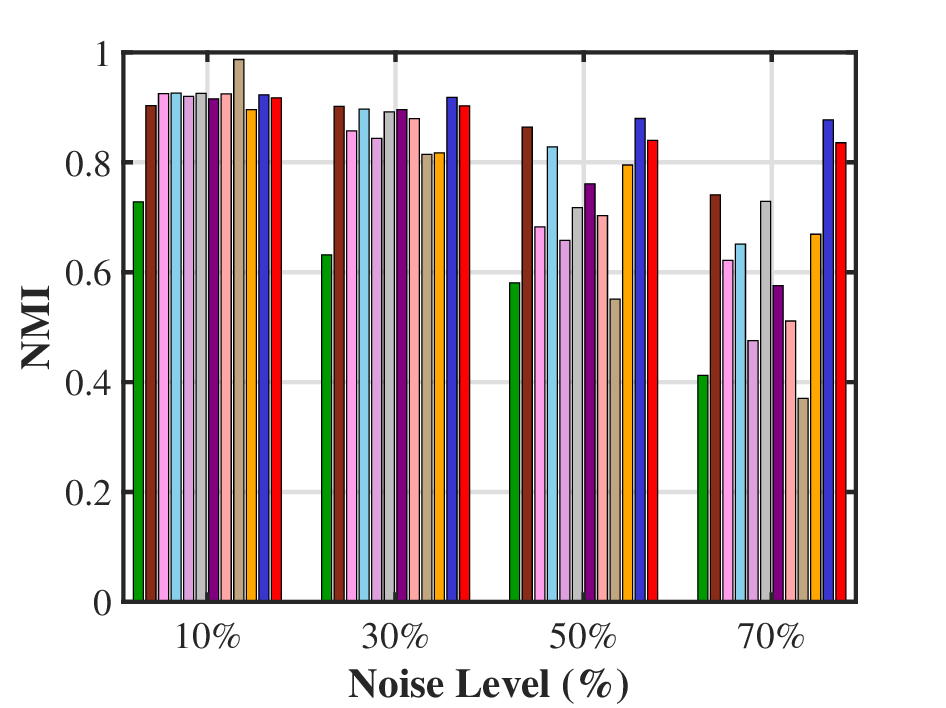}}
    \subfigure[(h) PUR (COIL20)]{
    \includegraphics[width=1.6 in]{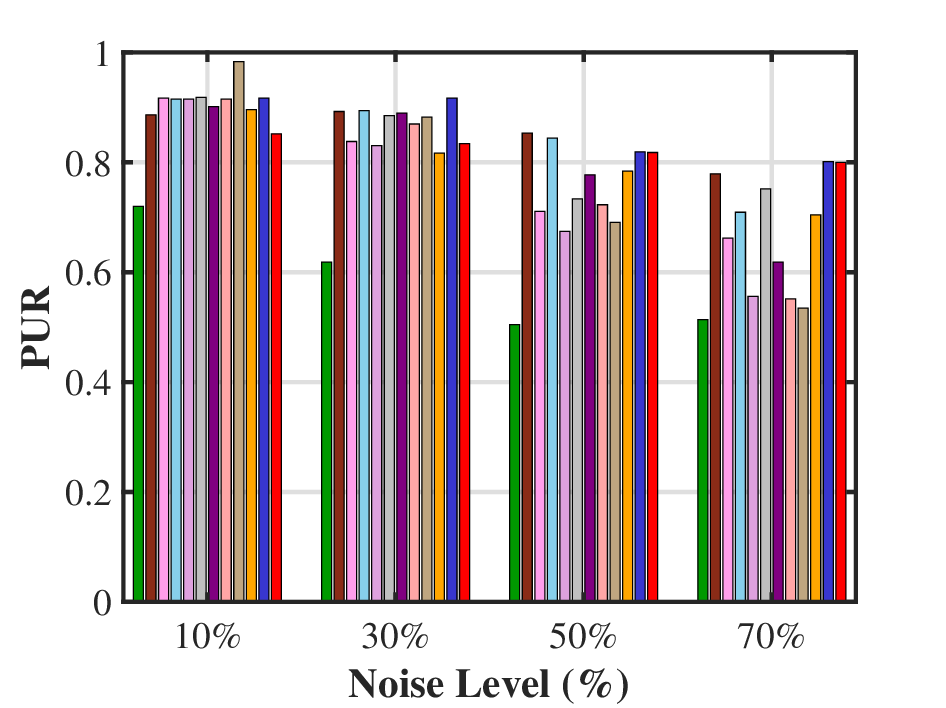}}
    \subfigure[(i) ACC (USPS)]{\label{noiseus}
    \includegraphics[width=1.6 in]{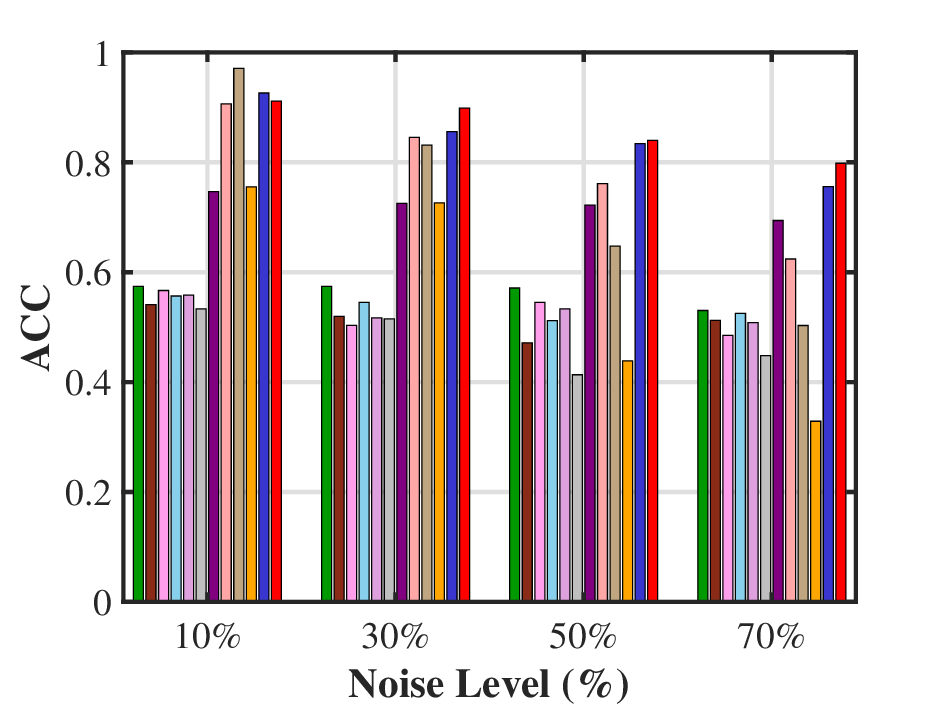}}
    \subfigure[(j) F1 (USPS)]{
    \includegraphics[width=1.6 in]{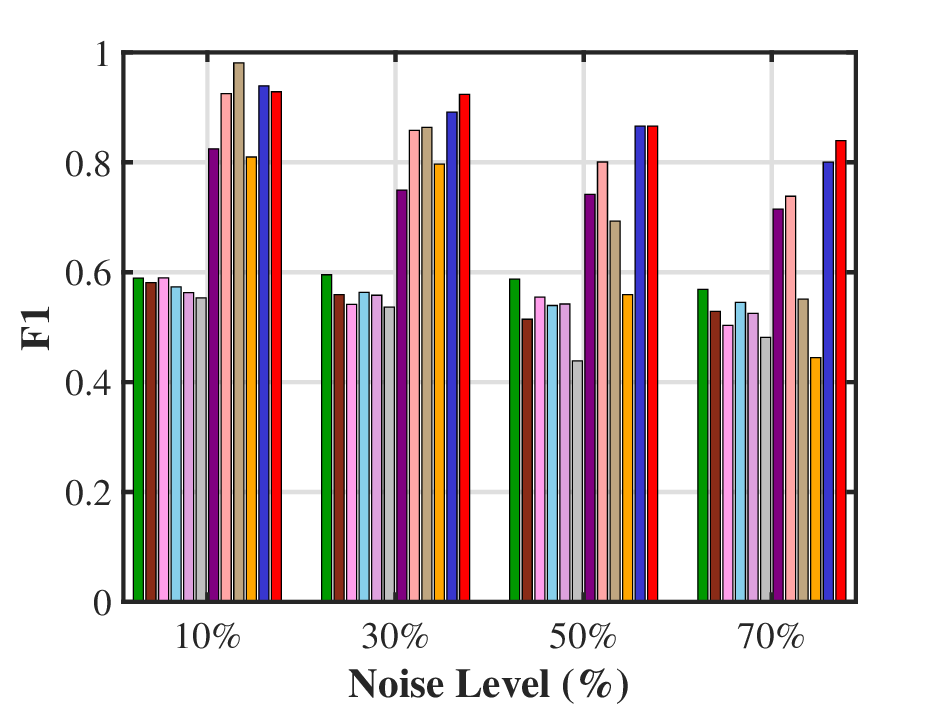}}
    \subfigure[(k) NMI (USPS)]{
    \includegraphics[width=1.6 in]{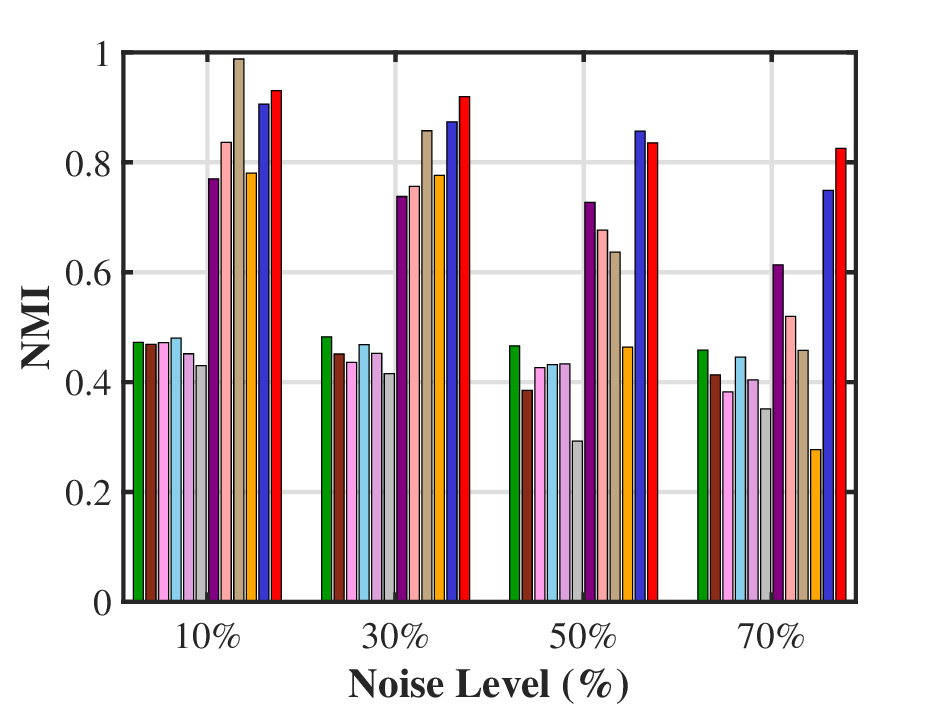}}
    \subfigure[(l) PUR (USPS)]{
    \includegraphics[width=1.6 in]{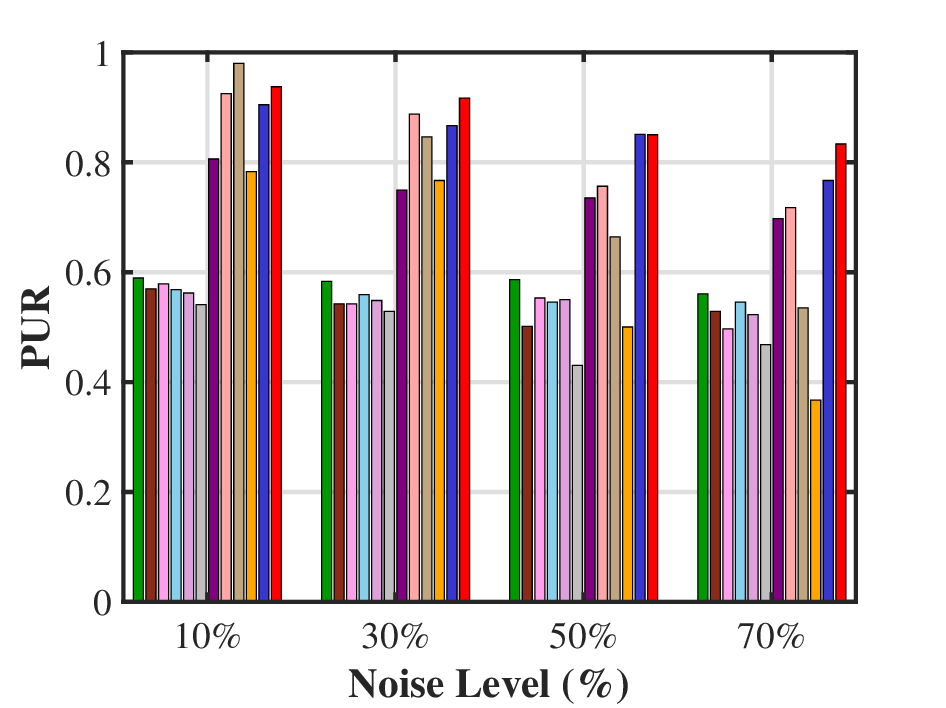}}
    \subfigure[]{\label{}
    \includegraphics[width=6 in]{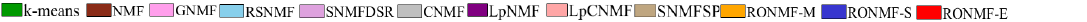}}
    \vskip-0.2cm
  \caption{{Performance of all methods on YALE, COIL20, and USPS  with 10\%, 30\%, 50\%, and 70\% noise.}}\label{noise}
\end{figure*}

\begin{figure*}[ht!]
  \centering
      \makeatletter
    \renewcommand{\@thesubfigure}{\hskip\subfiglabelskip}
    \makeatother
    \subfigure[~~~~~~(a) ACC (YALE)]{\label{yale}
    \includegraphics[width=1.8 in]{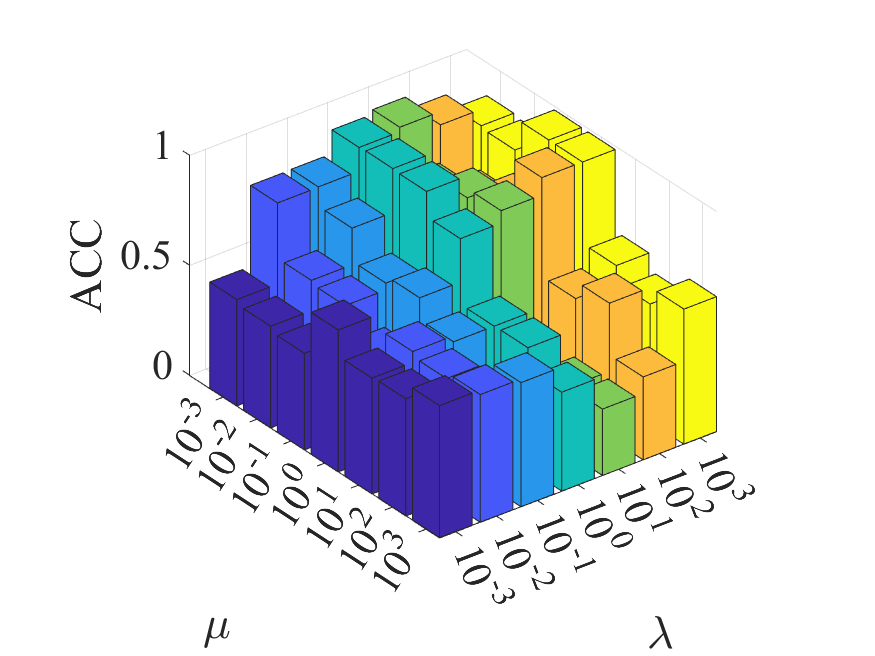}}\hspace{-6mm}
     \subfigure[~~~~~~(b) F1 (YALE)]{
    \includegraphics[width=1.8 in]{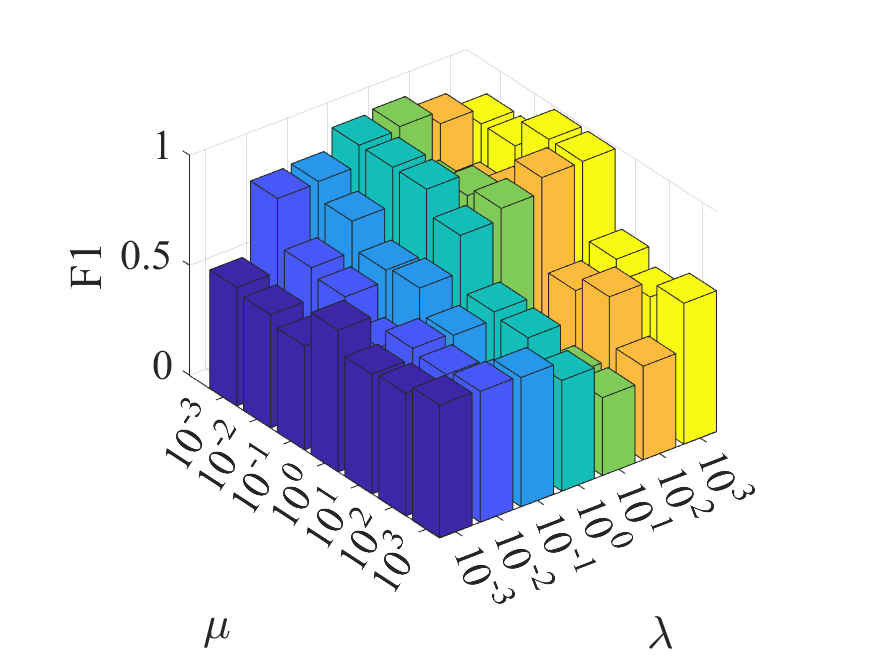}}\hspace{-6mm}
     \subfigure[~~~~~~(c) NMI (YALE)]{
    \includegraphics[width=1.8 in]{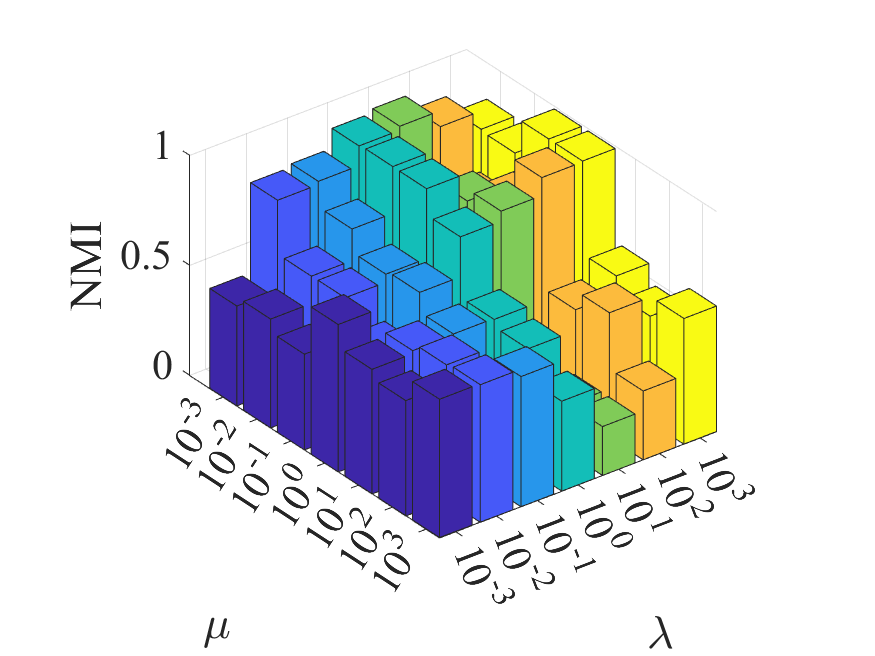}}\hspace{-6mm}
     \subfigure[~~~~~~(d) PUR (YALE)]{
    \includegraphics[width=1.8 in]{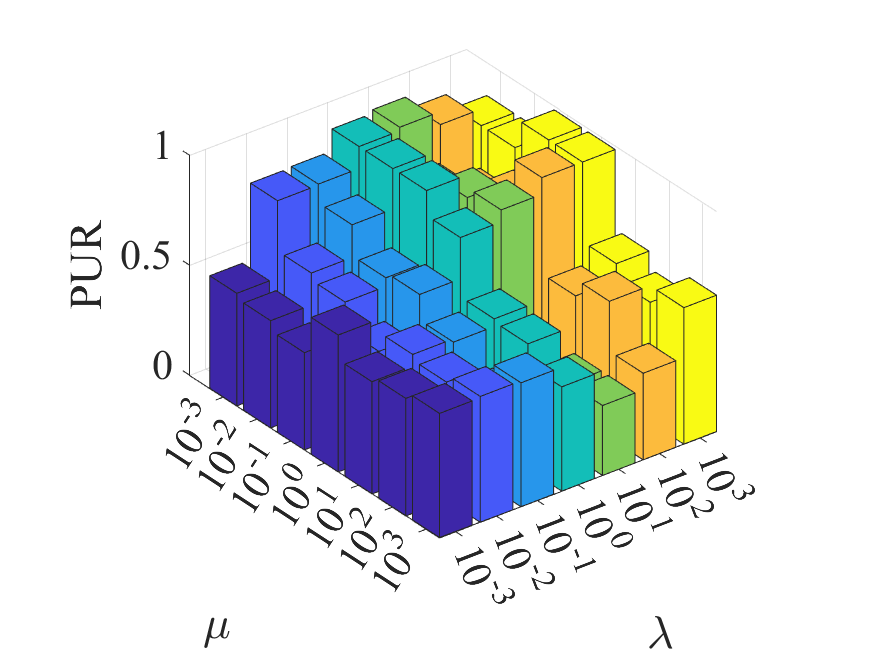}}\hspace{-6mm}
    \subfigure[~~~~~~(e) ACC (COIL20)]{\label{coil}
    \includegraphics[width=1.8 in]{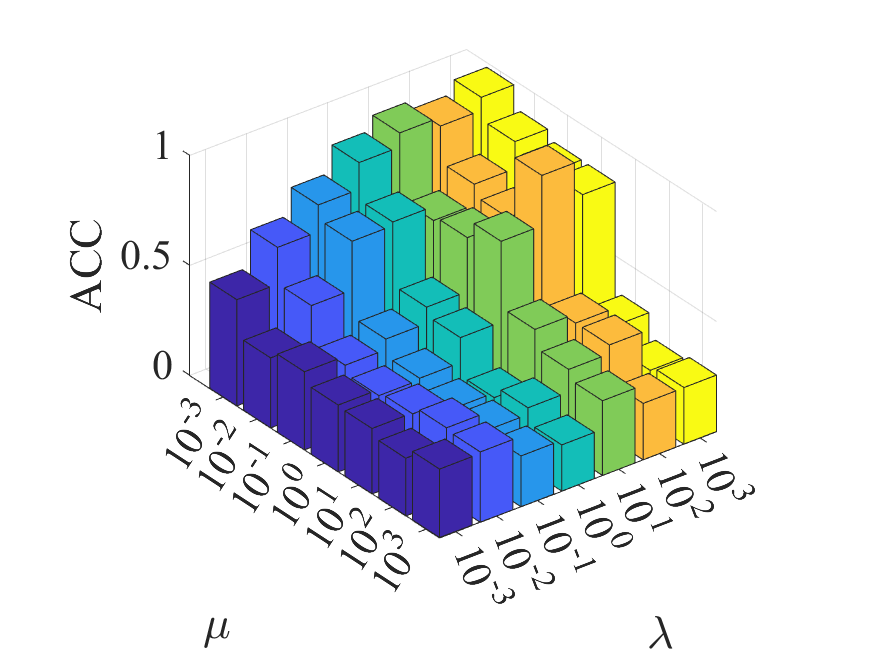}}\hspace{-6mm}
    \subfigure[~~~~~~(f) F1 (COIL20)]{
    \includegraphics[width=1.8 in]{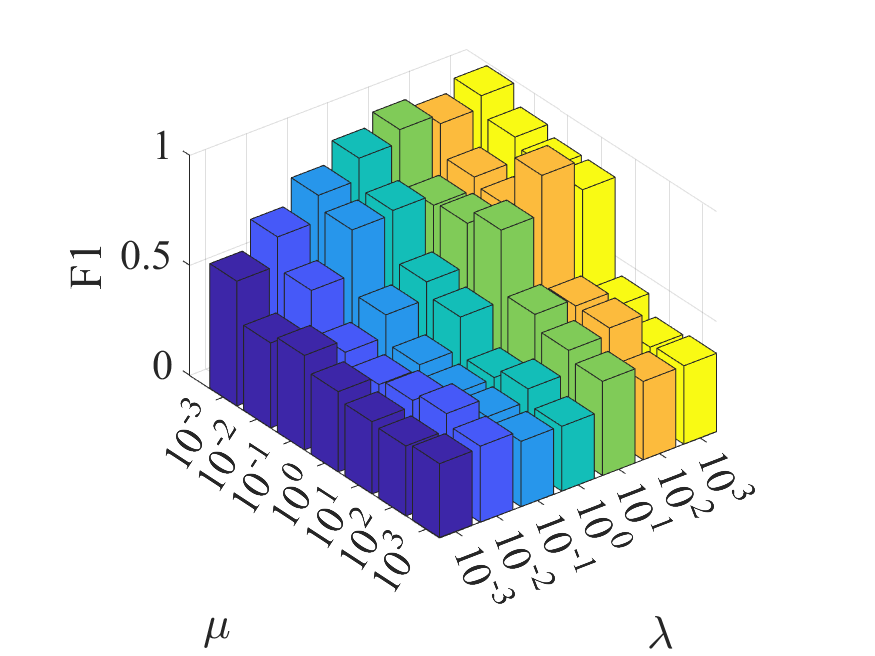}}\hspace{-6mm}
    \subfigure[~~~~~~(g) NMI (COIL20)]{
    \includegraphics[width=1.8 in]{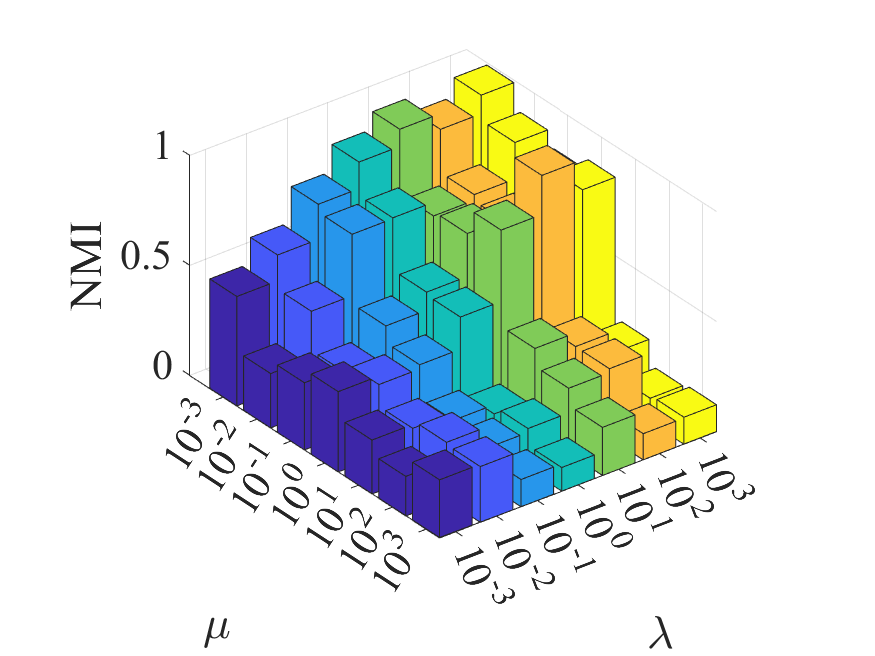}}\hspace{-6mm}
    \subfigure[~~~~~~(h) PUR (COIL20)]{
    \includegraphics[width=1.8 in]{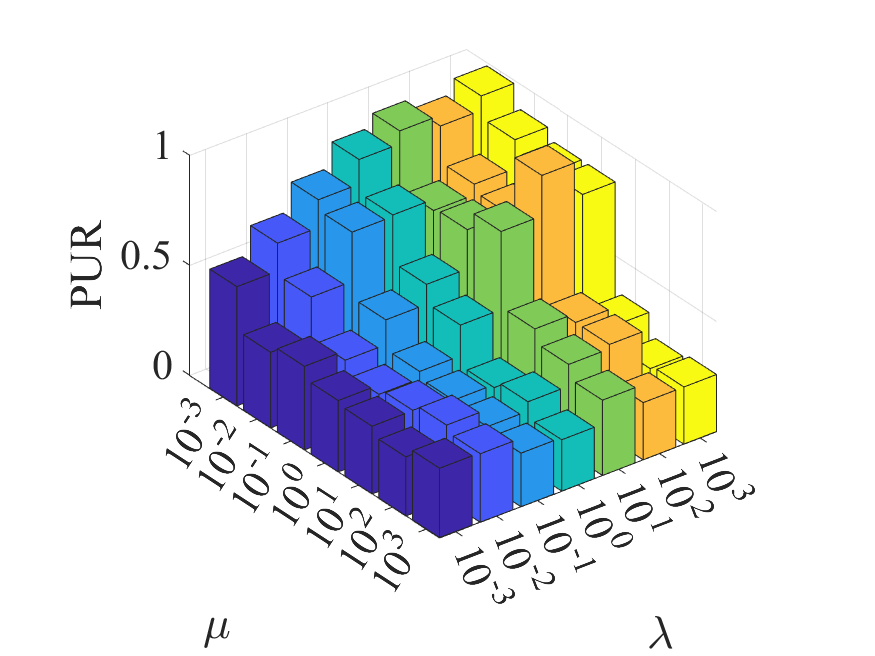}}\hspace{-6mm}
    \subfigure[~~~~~~(i) ACC (USPS)]{\label{usps}
    \includegraphics[width=1.8 in]{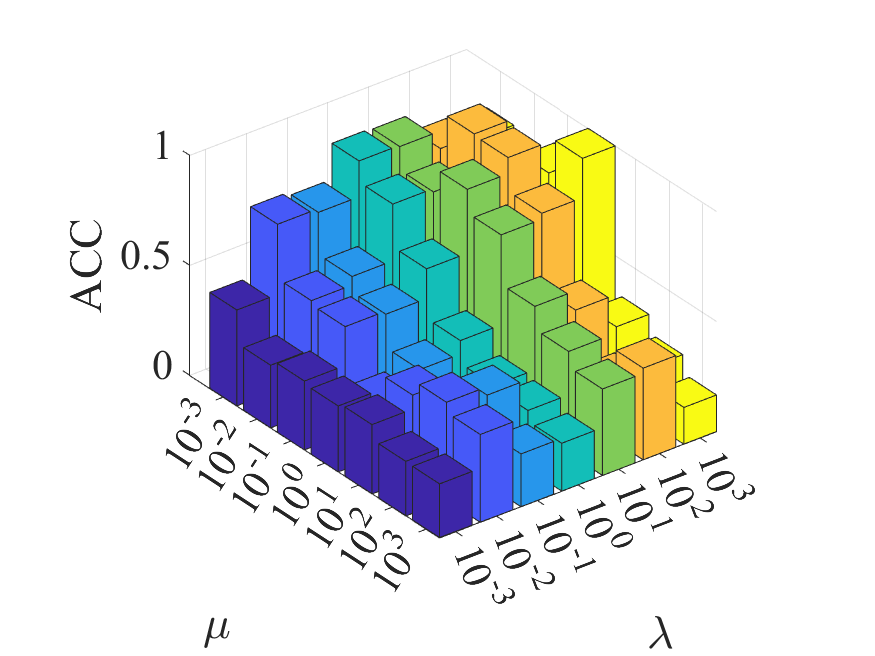}}\hspace{-6mm}
    \subfigure[~~~~~~(j) F1 (USPS)]{
    \includegraphics[width=1.8 in]{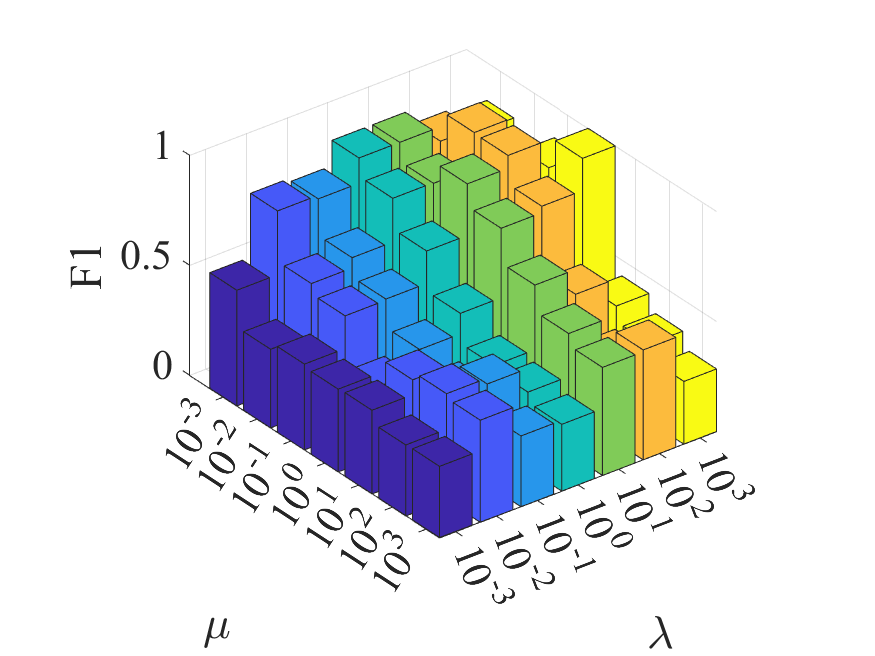}}\hspace{-6mm}
    \subfigure[~~~~~~(k) NMI (USPS)]{
    \includegraphics[width=1.8 in]{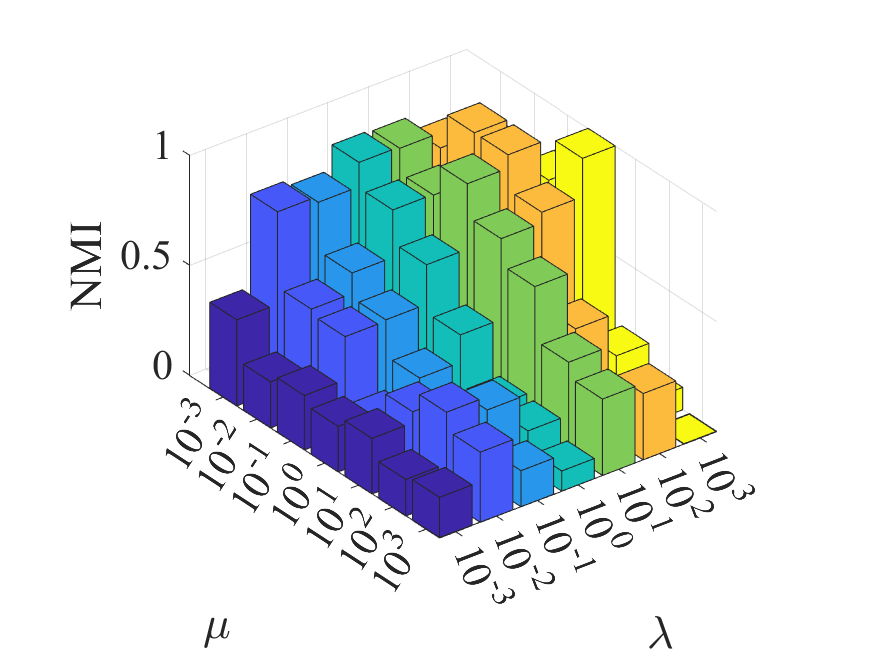}}\hspace{-6mm}
    \subfigure[~~~~~~(l) PUR (USPS)]{
    \includegraphics[width=1.8 in]{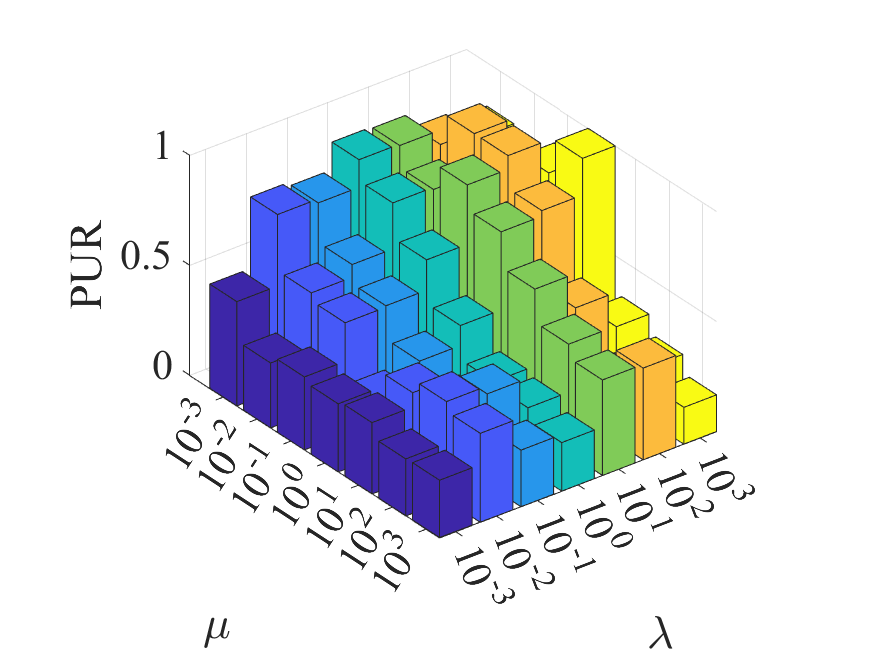}}\hspace{-6mm}
  \caption{Effects of parameters $p$ and $\lambda$ on the ACC, F1, PUR,  and NMI results.}\label{zxtclean}
   \label{heatmap}
\end{figure*}

\begin{figure}[t]
	\centering
    \vskip -0.5cm
	\includegraphics[width=3.2 in]{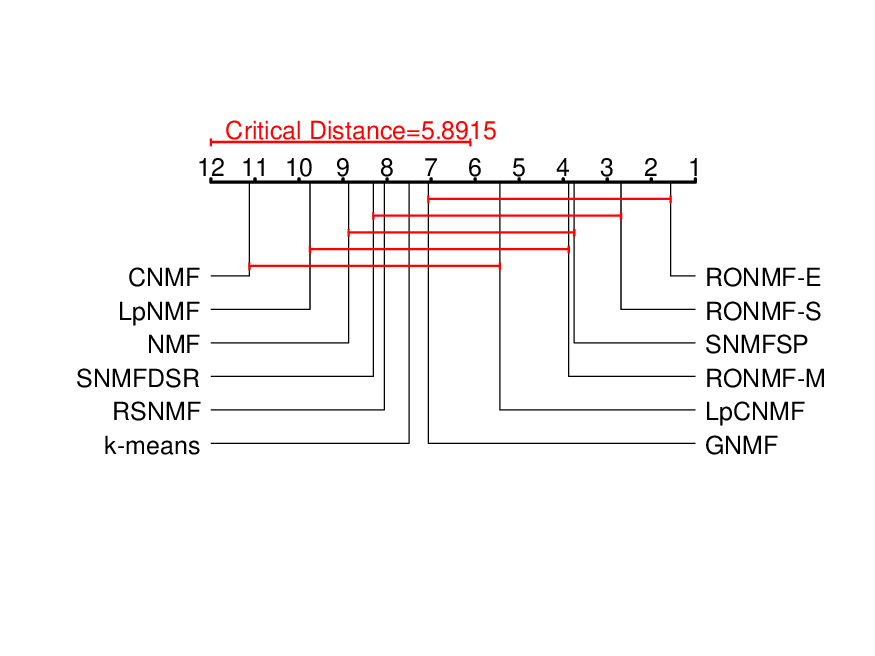}
	\vskip -1.8cm
	\caption{Post-hoc Nemenyi test of all compared methods.}
	\label{post}
\end{figure}

\subsection{Evaluation Indicators}

In the experiment, the performance of clustering is assessed using four indicators: accuracy (ACC), F1-score (F1), normalized mutual information (NMI) \cite{wu2018pairwise}, and purity (PUR),  which are defined below.

\begin{itemize}
\item
ACC quantifies the correctness of clustering by evaluating the alignment between predicted labels and true cluster labels. It is mathematically expressed as
\begin{equation} \label{acc}
\textrm{ACC}=\frac {1}{N}\sum _{i=1}^{N}\delta ({\varphi}(k_{i}),\mathrm {\varphi}(t_{i})),
\end{equation}
where $k_i$ and $t_i$ are the predicted and true cluster labels of the $i$-th data point, respectively, and
\begin{equation} \label{acc}
\delta ({\varphi}(k_{i}),\mathrm {\varphi}(t_{i})) = \left\{\begin{matrix}   1 , &\textrm{if}~{\varphi}(k_{i})= {\varphi}(t_{i}),\\    0 , &\textrm{if}~{\varphi}(k_{i})\neq {\varphi}(t_{i}).
\end{matrix}\right.
\end{equation}

\item
F1 is the harmonic mean of Precision and Recall, calculated using the following formula
\begin{equation}\label{fsc}
\textrm{F1}=\frac{2\times \textrm{Precision}\times \textrm{Recall}}{\textrm{Precision} + \textrm{Recall}},
\end{equation}
where Precision measures the ability to correctly identify negative instances, and Recall measures the ability to correctly identify positive instances.

\item
NMI is employed in clustering tasks to assess the similarity between clustering results and true labels, with values ranging from 0 to 1, where higher values indicate better alignment.
It is given by
\begin{equation}\label{nmi}
	\textrm{NMI}=\frac{\textrm{MI}(K,T)}{\max[H(K),H(T)]},
\end{equation}
where $\textrm{MI}(K; T)$ is the mutual information between the clustering results and true labels, and $H(K)$ and $H(T)$ are the entropies of the clustering results and true labels, respectively.

\item
PUR is a simple metric by calculating the proportion of correctly assigned instances to the total number of instances.
It is defined as
\begin{equation}\label{pur}
\textrm{PUR} =\frac{1}{N} \sum_{t \in T} \max_{k \in K} \left | T \cap K \right | ,
\end{equation}
where $T$ and $K$ denote the true and predicted clusters, respectively, and $N$ is the total number of samples.

\end{itemize}

\begin{table*}[ht!]
\centering
\caption{Clustering (mean±std) results comparison under Poisson noise and Salt pepper noise ($k=6$). The best and second-best results are marked in \red{red} and \blue{blue}, respectively.}\label{tabnoise1}
	\setlength{\tabcolsep}{1.25pt}
		\scalebox{0.98}{
\renewcommand{\arraystretch}{1.1}
\begin{tabular}{|c|c|ccccccccc|ccc|}
\hline
\multicolumn{2}{|c|}{Poisson} & \makecell{ k-means \\ \cite{liu2008reducing}} & \makecell{ NMF \\ \cite{lee1999learning}} & \makecell{ GNMF \\ \cite{cai2010graph}} & \makecell{ RSNMF \\ \cite{li2017robust}} & \makecell{ SNMFDSR \\ \cite{jia2019semi}} & \makecell{ CNMF \\ \cite{liu2011constrained} }&\makecell{  LpNMF \\ \cite{lan2020label}} & \makecell{ LpCNMF \\ \cite{liu2023constrained}} & \makecell{ SNMFSP \\ \cite{jing2025semi}} & RONMF-M & RONMF-S & RONMF-E \\ \hline \hline
\multirow{4}{*}[-5.5ex]{\makebox[0.05\textwidth][c]{YALE}}
& ACC & \makecell{0.230 \\ \scriptsize ($\pm$0.014)} & \makecell{0.282 \\ \scriptsize ($\pm$0.034)} & \makecell{0.279 \\ \scriptsize ($\pm$0.018)} & \makecell{0.275 \\ \scriptsize ($\pm$0.036)} & \makecell{0.348 \\ \scriptsize ($\pm$0.078)} & \makecell{0.296 \\ \scriptsize ($\pm$0.047)} & \makecell{0.306 \\ \scriptsize ($\pm$0.044)} & \makecell{0.200 \\ \scriptsize ($\pm$0.030)} & \makecell{0.374 \\ \scriptsize ($\pm$0.023)} & \makecell{\blue{0.950} \\ \scriptsize\blue{($\pm$0.081)}} & \makecell{\blue{0.950} \\ \scriptsize\blue{($\pm$0.081)}} & \makecell{\red{0.983} \\ \scriptsize\red{($\pm$0.053)}} \\
& F1 & \makecell{0.290 \\ \scriptsize ($\pm$0.014)} & \makecell{0.305 \\ \scriptsize ($\pm$0.029)} & \makecell{0.306 \\ \scriptsize ($\pm$0.018)} & \makecell{0.312 \\ \scriptsize ($\pm$0.016)} & \makecell{0.394 \\ \scriptsize ($\pm$0.050)} & \makecell{0.320 \\ \scriptsize ($\pm$0.035)} & \makecell{0.320 \\ \scriptsize ($\pm$0.028)} & \makecell{0.386 \\ \scriptsize ($\pm$0.028)} & \makecell{0.391 \\ \scriptsize ($\pm$0.011)} & \makecell{\blue{0.959} \\ \scriptsize\blue{($\pm$0.066)}} & \makecell{\blue{0.959} \\ \scriptsize\blue{($\pm$0.066)}} & \makecell{\red{0.986} \\ \scriptsize\red{($\pm$0.043)}} \\
& NMI & \makecell{0.089 \\ \scriptsize ($\pm$0.020)} & \makecell{0.116 \\ \scriptsize ($\pm$0.037)} & \makecell{0.111 \\ \scriptsize ($\pm$0.016)} & \makecell{0.138 \\ \scriptsize ($\pm$0.021)} & \makecell{0.246 \\ \scriptsize ($\pm$0.067)} & \makecell{0.178 \\ \scriptsize ($\pm$0.066)} & \makecell{0.178 \\ \scriptsize ($\pm$0.035)} & \makecell{0.109 \\ \scriptsize ($\pm$0.015)} & \makecell{0.222 \\ \scriptsize ($\pm$0.017)} & \makecell{\blue{0.964} \\ \scriptsize\blue{($\pm$0.059)}} & \makecell{\blue{0.964} \\ \scriptsize\blue{($\pm$0.059)}} & \makecell{\red{0.988} \\ \scriptsize\red{($\pm$0.038)}} \\
& PUR & \makecell{0.253 \\ \scriptsize ($\pm$0.016)} & \makecell{0.292 \\ \scriptsize ($\pm$0.034)} & \makecell{0.292 \\ \scriptsize ($\pm$0.018)} & \makecell{0.295 \\ \scriptsize ($\pm$0.021)} & \makecell{0.376 \\ \scriptsize ($\pm$0.042)} & \makecell{0.311 \\ \scriptsize ($\pm$0.031)} & \makecell{0.306 \\ \scriptsize ($\pm$0.023)} & \makecell{0.367 \\ \scriptsize ($\pm$0.026)} & \makecell{0.382 \\ \scriptsize ($\pm$0.014)} & \makecell{\blue{0.950} \\ \scriptsize\blue{($\pm$0.081)}} & \makecell{\blue{0.950} \\ \scriptsize\blue{($\pm$0.081)}} & \makecell{\red{0.983} \\ \scriptsize\red{($\pm$0.053)}} \\
\hline

\multirow{4}{*}[-5.5ex]{COIL20} & ACC & \makecell{0.700 \\ \scriptsize ($\pm$0.083)} & \makecell{0.665 \\ \scriptsize ($\pm$0.102)} & \makecell{0.843 \\ \scriptsize ($\pm$0.142)} & \makecell{0.681 \\ \scriptsize ($\pm$0.120)} & \makecell{0.706 \\ \scriptsize ($\pm$0.131)} & \makecell{0.385 \\ \scriptsize ($\pm$0.037)} & \makecell{0.668 \\ \scriptsize ($\pm$0.092)} & \makecell{0.883 \\ \scriptsize ($\pm$0.133)} & \makecell{0.953 \\ \scriptsize ($\pm$0.099)} & \makecell{\red{1.000} \\ \scriptsize\red{($\pm$0.000)}} & \makecell{\red{1.000} \\ \scriptsize\red{($\pm$0.000)}} & \makecell{\blue{0.965} \\ \scriptsize\blue{($\pm$0.074)}} \\
& F1 & \makecell{0.709 \\ \scriptsize ($\pm$0.077)} & \makecell{0.684 \\ \scriptsize ($\pm$0.088)} & \makecell{0.863 \\ \scriptsize ($\pm$0.123)} & \makecell{0.719 \\ \scriptsize ($\pm$0.105)} & \makecell{0.717 \\ \scriptsize ($\pm$0.120)} & \makecell{0.423 \\ \scriptsize ($\pm$0.031)} & \makecell{0.691 \\ \scriptsize ($\pm$0.078)} & \makecell{0.905 \\ \scriptsize ($\pm$0.110)} & \makecell{\blue{0.998} \\ \scriptsize\blue{($\pm$0.000)}} & \makecell{\red{1.000} \\ \scriptsize\red{($\pm$0.000)}} & \makecell{\red{1.000} \\ \scriptsize\red{($\pm$0.000)}} & \makecell{0.972 \\ \scriptsize ($\pm$0.059)} \\
& NMI & \makecell{0.682 \\ \scriptsize ($\pm$0.103)} & \makecell{0.644 \\ \scriptsize ($\pm$0.105)} & \makecell{0.854 \\ \scriptsize ($\pm$0.131)} & \makecell{0.685 \\ \scriptsize ($\pm$0.112)} & \makecell{0.687 \\ \scriptsize ($\pm$0.154)} & \makecell{0.262 \\ \scriptsize ($\pm$0.045)} & \makecell{0.645 \\ \scriptsize ($\pm$0.095)} & \makecell{0.884 \\ \scriptsize ($\pm$0.121)} & \makecell{\blue{0.974} \\ \scriptsize\blue{($\pm$0.054)}} & \makecell{\red{1.000} \\ \scriptsize\red{($\pm$0.000)}} & \makecell{\red{1.000} \\ \scriptsize\red{($\pm$0.000)}} & \makecell{\blue{0.974} \\ \scriptsize\blue{($\pm$0.054)}} \\
& PUR & \makecell{0.716 \\ \scriptsize ($\pm$0.086)} & \makecell{0.686 \\ \scriptsize ($\pm$0.090)} & \makecell{0.857 \\ \scriptsize ($\pm$0.133)} & \makecell{0.721 \\ \scriptsize ($\pm$0.096)} & \makecell{0.716 \\ \scriptsize ($\pm$0.123)} & \makecell{0.416 \\ \scriptsize ($\pm$0.042)} & \makecell{0.689 \\ \scriptsize ($\pm$0.081)} & \makecell{0.898 \\ \scriptsize ($\pm$0.115)} & \makecell{\blue{0.967} \\ \scriptsize \blue{($\pm$0.004)}} & \makecell{\red{1.000} \\ \scriptsize\red{($\pm$0.000)}} & \makecell{\red{1.000} \\ \scriptsize\red{($\pm$0.000)}} & \makecell{\blue{0.967} \\ \scriptsize\blue{($\pm$0.070)}} \\ \hline

\multirow{4}{*}[-5.5ex]{USPS} & ACC & \makecell{0.810 \\ \scriptsize ($\pm$0.055)} & \makecell{0.699 \\ \scriptsize ($\pm$0.097)} & \makecell{0.746 \\ \scriptsize ($\pm$0.116)} & \makecell{0.651 \\ \scriptsize ($\pm$0.046)} & \makecell{0.638 \\ \scriptsize ($\pm$0.046)} & \makecell{0.261 \\ \scriptsize ($\pm$0.041)} & \makecell{0.714 \\ \scriptsize ($\pm$0.066)} & \makecell{\red{0.943} \\ \scriptsize \red{($\pm$0.016)}} & \makecell{0.163 \\ \scriptsize ($\pm$0.007)} & \makecell{0.334 \\ \scriptsize ($\pm$0.337)} & \makecell{0.334 \\ \scriptsize ($\pm$0.337)} & \makecell{\blue{0.907} \\ \scriptsize \blue{($\pm$0.161)}} \\
& F1 & \makecell{0.810 \\ \scriptsize ($\pm$0.055)} & \makecell{0.715 \\ \scriptsize ($\pm$0.081)} & \makecell{0.788 \\ \scriptsize ($\pm$0.082)} & \makecell{0.667 \\ \scriptsize ($\pm$0.039)} & \makecell{0.660 \\ \scriptsize ($\pm$0.038)} & \makecell{0.299 \\ \scriptsize ($\pm$0.049)} & \makecell{0.714 \\ \scriptsize ($\pm$0.063)} & \makecell{\red{0.954} \\ \scriptsize \red{($\pm$0.012)}} & \makecell{0.184 \\ \scriptsize ($\pm$0.004)} & \makecell{0.421 \\ \scriptsize ($\pm$0.286)} & \makecell{0.421 \\ \scriptsize ($\pm$0.286)} & \makecell{\blue{0.935} \\ \scriptsize \blue{($\pm$0.113)}} \\
& NMI & \makecell{0.645 \\ \scriptsize ($\pm$0.060)} & \makecell{0.548 \\ \scriptsize ($\pm$0.072)} & \makecell{0.682 \\ \scriptsize ($\pm$0.077)} & \makecell{0.516 \\ \scriptsize ($\pm$0.048)} & \makecell{0.515 \\ \scriptsize ($\pm$0.057)} & \makecell{0.076 \\ \scriptsize ($\pm$0.047)} & \makecell{0.545 \\ \scriptsize ($\pm$0.055)} & \makecell{\blue{0.853} \\ \scriptsize \blue{($\pm$0.032)}} & \makecell{0.119 \\ \scriptsize ($\pm$0.002)} & \makecell{0.211 \\ \scriptsize ($\pm$0.398)} & \makecell{0.211 \\ \scriptsize ($\pm$0.398)} & \makecell{\red{0.944} \\ \scriptsize \red{($\pm$0.103)}} \\
& PUR & \makecell{0.810 \\ \scriptsize ($\pm$0.055)} & \makecell{0.713 \\ \scriptsize ($\pm$0.084)} & \makecell{0.779 \\ \scriptsize ($\pm$0.090)} & \makecell{0.668 \\ \scriptsize ($\pm$0.044)} & \makecell{0.640 \\ \scriptsize ($\pm$0.044)} & \makecell{0.275 \\ \scriptsize ($\pm$0.044)} & \makecell{0.716 \\ \scriptsize ($\pm$0.060)} & \makecell{\red{0.954} \\ \scriptsize \red{($\pm$0.013)}} & \makecell{0.187 \\ \scriptsize ($\pm$0.070)} & \makecell{0.335 \\ \scriptsize ($\pm$0.332)} & \makecell{0.335 \\ \scriptsize ($\pm$0.332)} & \makecell{\blue{0.933} \\ \scriptsize \blue{($\pm$0.117)}} \\ \hline \hline

\multicolumn{2}{|c|}{Average} & \makecell{0.654 \\ \scriptsize ($\pm$0.108)} & \makecell{0.614 \\ \scriptsize ($\pm$0.117)} & \makecell{0.736 \\ \scriptsize ($\pm$0.139)} & \makecell{0.619 \\ \scriptsize ($\pm$0.107)} & \makecell{0.629 \\ \scriptsize ($\pm$0.132)} & \makecell{0.318 \\ \scriptsize ($\pm$0.055)} & \makecell{0.620 \\ \scriptsize ($\pm$0.113)} & \makecell{\blue{0.807} \\ \scriptsize \blue{($\pm$0.130)}} & \makecell{0.641 \\ \scriptsize ($\pm$0.049)} & \makecell{0.797 \\ \scriptsize ($\pm$0.116)} & \makecell{0.797 \\ \scriptsize ($\pm$0.116)} & \makecell{\red{0.963} \\ \scriptsize \red{($\pm$0.076)}} \\  \hline \hline

\multicolumn{2}{|c|}{Salt pepper} & \makecell{ k-means \\ \cite{liu2008reducing}} & \makecell{ NMF \\ \cite{lee1999learning}} & \makecell{ GNMF \\ \cite{cai2010graph}} & \makecell{ RSNMF \\ \cite{li2017robust}} & \makecell{ SNMFDSR \\ \cite{jia2019semi}} & \makecell{ CNMF \\ \cite{liu2011constrained} }&\makecell{  LpNMF \\ \cite{lan2020label}} & \makecell{ LpCNMF \\ \cite{liu2023constrained}} & \makecell{ SNMFSP \\ \cite{jing2025semi}} & RONMF-M & RONMF-S & RONMF-E \\ \hline \hline

\multirow{4}{*}[-5.5ex]{YALE} & ACC & \makecell{0.230 \\ \scriptsize ($\pm$0.014)} & \makecell{0.282 \\ \scriptsize ($\pm$0.034)} & \makecell{0.279 \\ \scriptsize ($\pm$0.018)} & \makecell{0.275 \\ \scriptsize ($\pm$0.036)} & \makecell{0.348 \\ \scriptsize ($\pm$0.078)} & \makecell{0.296 \\ \scriptsize ($\pm$0.047)} & \makecell{0.306 \\ \scriptsize ($\pm$0.044)} & \makecell{0.200 \\ \scriptsize ($\pm$0.030)} & \makecell{\blue{0.711} \\ \scriptsize \blue{($\pm$0.073)}} & \makecell{0.397 \\ \scriptsize ($\pm$0.275)} & \makecell{0.397 \\ \scriptsize ($\pm$0.275)} & \makecell{\red{0.914} \\ \scriptsize \red{($\pm$0.146)}} \\
& F1 & \makecell{0.290 \\ \scriptsize ($\pm$0.014)} & \makecell{0.305 \\ \scriptsize ($\pm$0.029)} & \makecell{0.306 \\ \scriptsize ($\pm$0.018)} & \makecell{0.312 \\ \scriptsize ($\pm$0.016)} & \makecell{0.394 \\ \scriptsize ($\pm$0.050)} & \makecell{0.320 \\ \scriptsize ($\pm$0.035)} & \makecell{0.320 \\ \scriptsize ($\pm$0.028)} & \makecell{0.386 \\ \scriptsize ($\pm$0.028)} & \makecell{\blue{0.728} \\ \scriptsize \blue{($\pm$0.054)}} & \makecell{0.507 \\ \scriptsize ($\pm$0.238)} & \makecell{0.507 \\ \scriptsize ($\pm$0.238)} & \makecell{\red{0.941} \\ \scriptsize \red{($\pm$0.103)}} \\
& NMI & \makecell{0.089 \\ \scriptsize ($\pm$0.020)} & \makecell{0.116 \\ \scriptsize ($\pm$0.037)} & \makecell{0.111 \\ \scriptsize ($\pm$0.016)} & \makecell{0.138 \\ \scriptsize ($\pm$0.021)} & \makecell{0.246 \\ \scriptsize ($\pm$0.067)} & \makecell{0.178 \\ \scriptsize ($\pm$0.066)} & \makecell{0.178 \\ \scriptsize ($\pm$0.035)} & \makecell{0.109 \\ \scriptsize ($\pm$0.015)} & \makecell{\blue{0.599} \\ \scriptsize \blue{($\pm$0.053)}} & \makecell{0.388 \\ \scriptsize ($\pm$0.328)} & \makecell{0.388 \\ \scriptsize ($\pm$0.328)} & \makecell{\red{0.944} \\ \scriptsize \red{($\pm$0.103)}} \\
& PUR & \makecell{0.253 \\ \scriptsize ($\pm$0.016)} & \makecell{0.292 \\ \scriptsize ($\pm$0.034)} & \makecell{0.292 \\ \scriptsize ($\pm$0.018)} & \makecell{0.295 \\ \scriptsize ($\pm$0.021)} & \makecell{0.376 \\ \scriptsize ($\pm$0.042)} & \makecell{0.311 \\ \scriptsize ($\pm$0.031)} & \makecell{0.306 \\ \scriptsize ($\pm$0.023)} & \makecell{0.367 \\ \scriptsize ($\pm$0.026)} & \makecell{\blue{0.724} \\ \scriptsize \blue{($\pm$0.059)}} & \makecell{0.453 \\ \scriptsize ($\pm$0.270)} & \makecell{0.453 \\ \scriptsize ($\pm$0.270)} & \makecell{\red{0.933} \\ \scriptsize \red{($\pm$0.117)}} \\
\hline \hline

\multirow{4}{*}[-5.5ex]{COIL20} & ACC & \makecell{0.700 \\ \scriptsize ($\pm$0.083)} & \makecell{0.665 \\ \scriptsize ($\pm$0.102)} & \makecell{0.843 \\ \scriptsize ($\pm$0.142)} & \makecell{0.681 \\ \scriptsize ($\pm$0.120)} & \makecell{0.706 \\ \scriptsize ($\pm$0.131)} & \makecell{0.385 \\ \scriptsize ($\pm$0.037)} & \makecell{0.668 \\ \scriptsize ($\pm$0.092)} & \makecell{0.883 \\ \scriptsize ($\pm$0.133)} & \makecell{\red{0.998} \\ \scriptsize \red{($\pm$0.000)}} & \makecell{\blue{0.967} \\ \scriptsize \blue{($\pm$0.070)}} & \makecell{\blue{0.967} \\ \scriptsize \blue{($\pm$0.070)}} & \makecell{0.908 \\ \scriptsize ($\pm$0.159)} \\
& F1 & \makecell{0.709 \\ \scriptsize ($\pm$0.077)} & \makecell{0.684 \\ \scriptsize ($\pm$0.088)} & \makecell{0.863 \\ \scriptsize ($\pm$0.123)} & \makecell{0.719 \\ \scriptsize ($\pm$0.105)} & \makecell{0.717 \\ \scriptsize ($\pm$0.120)} & \makecell{0.423 \\ \scriptsize ($\pm$0.031)} & \makecell{0.691 \\ \scriptsize ($\pm$0.078)} & \makecell{0.905 \\ \scriptsize ($\pm$0.110)} & \makecell{\red{0.998} \\ \scriptsize \red{($\pm$0.000)}} & \makecell{\blue{0.973} \\ \scriptsize \blue{($\pm$0.058)}} & \makecell{\blue{0.973} \\ \scriptsize \blue{($\pm$0.058)}} & \makecell{0.923 \\ \scriptsize ($\pm$0.133)} \\
& NMI & \makecell{0.682 \\ \scriptsize ($\pm$0.103)} & \makecell{0.644 \\ \scriptsize ($\pm$0.105)} & \makecell{0.854 \\ \scriptsize ($\pm$0.131)} & \makecell{0.685 \\ \scriptsize ($\pm$0.112)} & \makecell{0.687 \\ \scriptsize ($\pm$0.154)} & \makecell{0.262 \\ \scriptsize ($\pm$0.045)} & \makecell{0.645 \\ \scriptsize ($\pm$0.095)} & \makecell{0.884 \\ \scriptsize ($\pm$0.121)} & \makecell{\red{0.993} \\ \scriptsize \red{($\pm$0.000)}} & \makecell{\blue{0.976} \\ \scriptsize \blue{($\pm$0.051)}} & \makecell{\blue{0.976} \\ \scriptsize \blue{($\pm$0.051)}} & \makecell{0.929 \\ \scriptsize ($\pm$0.125)} \\
& PUR & \makecell{0.716 \\ \scriptsize ($\pm$0.086)} & \makecell{0.686 \\ \scriptsize ($\pm$0.090)} & \makecell{0.857 \\ \scriptsize ($\pm$0.133)} & \makecell{0.721 \\ \scriptsize ($\pm$0.096)} & \makecell{0.716 \\ \scriptsize ($\pm$0.123)} & \makecell{0.416 \\ \scriptsize ($\pm$0.042)} & \makecell{0.689 \\ \scriptsize ($\pm$0.081)} & \makecell{0.898 \\ \scriptsize ($\pm$0.115)} & \makecell{\red{0.998} \\ \scriptsize \red{($\pm$0.000)}} & \makecell{\blue{0.967} \\ \scriptsize \blue{($\pm$0.070)}} & \makecell{\blue{0.967} \\ \scriptsize \blue{($\pm$0.070)}} & \makecell{0.917 \\ \scriptsize ($\pm$0.142)} \\
\hline
\multirow{4}{*}[-5.5ex]{USPS} & ACC & \makecell{0.810 \\ \scriptsize ($\pm$0.055)} & \makecell{0.699 \\ \scriptsize ($\pm$0.097)} & \makecell{0.746 \\ \scriptsize ($\pm$0.116)} & \makecell{0.651 \\ \scriptsize ($\pm$0.046)} & \makecell{0.638 \\ \scriptsize ($\pm$0.046)} & \makecell{0.261 \\ \scriptsize ($\pm$0.041)} & \makecell{0.714 \\ \scriptsize ($\pm$0.066)} & \makecell{\blue{0.943} \\ \scriptsize\blue{($\pm$0.016)}} & \makecell{0.172 \\ \scriptsize ($\pm$0.007)} & \makecell{0.251 \\ \scriptsize ($\pm$0.263)} & \makecell{0.251 \\ \scriptsize ($\pm$0.263)} & \makecell{\red{0.981} \\ \scriptsize\red{($\pm$0.059)}} \\
& F1 & \makecell{0.810 \\ \scriptsize ($\pm$0.055)} & \makecell{0.715 \\ \scriptsize ($\pm$0.081)} & \makecell{0.788 \\ \scriptsize ($\pm$0.082)} & \makecell{0.667 \\ \scriptsize ($\pm$0.039)} & \makecell{0.660 \\ \scriptsize ($\pm$0.038)} & \makecell{0.299 \\ \scriptsize ($\pm$0.049)} & \makecell{0.714 \\ \scriptsize ($\pm$0.063)} & \makecell{\blue{0.954} \\ \scriptsize\blue{($\pm$0.012)}} & \makecell{0.191 \\ \scriptsize ($\pm$0.007)} & \makecell{0.357 \\ \scriptsize ($\pm$0.226)} & \makecell{0.357 \\ \scriptsize ($\pm$0.226)} & \makecell{\red{0.986} \\ \scriptsize\red{($\pm$0.046)}} \\
& NMI & \makecell{0.645 \\ \scriptsize ($\pm$0.060)} & \makecell{0.548 \\ \scriptsize ($\pm$0.072)} & \makecell{0.682 \\ \scriptsize ($\pm$0.077)} & \makecell{0.516 \\ \scriptsize ($\pm$0.048)} & \makecell{0.515 \\ \scriptsize ($\pm$0.057)} & \makecell{0.076 \\ \scriptsize ($\pm$0.047)} & \makecell{0.545 \\ \scriptsize ($\pm$0.055)} & \makecell{\blue{0.853} \\ \scriptsize\blue{($\pm$0.032)}} & \makecell{0.020 \\ \scriptsize ($\pm$0.003)} & \makecell{0.102 \\ \scriptsize ($\pm$0.315)} & \makecell{0.102 \\ \scriptsize ($\pm$0.315)} & \makecell{\red{0.987} \\ \scriptsize\red{($\pm$0.041)}} \\
& PUR & \makecell{0.810 \\ \scriptsize ($\pm$0.055)} & \makecell{0.713 \\ \scriptsize ($\pm$0.084)} & \makecell{0.779 \\ \scriptsize ($\pm$0.090)} & \makecell{0.668 \\ \scriptsize ($\pm$0.044)} & \makecell{0.640 \\ \scriptsize ($\pm$0.044)} & \makecell{0.275 \\ \scriptsize ($\pm$0.044)} & \makecell{0.716 \\ \scriptsize ($\pm$0.060)} & \makecell{\blue{0.954} \\ \scriptsize\blue{($\pm$0.013)}} & \makecell{0.183 \\ \scriptsize ($\pm$0.004)} & \makecell{0.251 \\ \scriptsize ($\pm$0.263)} & \makecell{0.251 \\ \scriptsize ($\pm$0.263)} & \makecell{\red{0.983} \\ \scriptsize\red{($\pm$0.053)}} \\ \hline \hline

\multicolumn{2}{|c|}{Average} & \makecell{0.562 \\ \scriptsize ($\pm$0.053)} & \makecell{0.529 \\ \scriptsize ($\pm$0.071)} & \makecell{0.617 \\ \scriptsize ($\pm$0.080)} & \makecell{0.527 \\ \scriptsize ($\pm$0.059)} & \makecell{0.554 \\ \scriptsize ($\pm$0.079)} & \makecell{0.292 \\ \scriptsize ($\pm$0.043)} & \makecell{0.541 \\ \scriptsize ($\pm$0.060)} & \makecell{\blue{0.695} \\ \scriptsize\blue{($\pm$0.054)}} & \makecell{0.609 \\ \scriptsize ($\pm$0.022)} & \makecell{0.512 \\ \scriptsize ($\pm$0.222)} & \makecell{0.512 \\ \scriptsize ($\pm$0.222)} & \makecell{\red{0.944} \\ \scriptsize\red{($\pm$0.104)}} \\ \hline

\end{tabular}}
\end{table*}

\begin{table}[ht!]
	\centering
	\caption{Average values of the indicators obtained on COIL20 with or without regularization terms.}\label{ablation}
	\setlength{\tabcolsep}{0.8pt}
		\scalebox{0.98}{
	\begin{tabular}{|ccc|ccc|c|c|c|c|}
		\hline
		Graph & Label & $U^\top U=I$ & MCP & SCAD & ETP & ACC   & F1   & NMI    & PUR             \\ \hline \hline
		& \Checkmark & \Checkmark & \Checkmark  &  &  & 0.257   & 0.415   & 0.116  & 0.369                \\
		& \Checkmark & \Checkmark &  & \Checkmark &  & 0.282   & 0.367   & 0.148  & 0.317                \\
		& \Checkmark & \Checkmark &  &  & \Checkmark & 0.213   & 0.360   & 0.082  & 0.303                \\ \hline
		&  & \Checkmark & \Checkmark &  &  & 0.184   & 0.294   & 0.042  & 0.194                \\
		&  & \Checkmark &  & \Checkmark &  & 0.175   & 0.285   & 0.024  & 0.178                \\
		&  & \Checkmark &  &  & \Checkmark  & 0.224   & 0.081   & 0.081  & 0.231
		\\ \hline
		\Checkmark &\Checkmark &  & \Checkmark &  &  & 0.632   & 0.707   & 0.706  & 0.679              \\
		\Checkmark &\Checkmark &  &  & \Checkmark &  & 0.781   & 0.834   & 0.830  & 0.817              \\
		\Checkmark &\Checkmark &  &  &  & \Checkmark & 0.859  & 0.884   & 0.893  & 0.867              \\ \hline
		\Checkmark &\Checkmark & \Checkmark & \Checkmark &  &  & 0.847   & 0.838   & 0.823  & 0.855              \\
		\Checkmark &\Checkmark & \Checkmark &  & \Checkmark &  & 0.871   & 0.883   & 0.949  & 0.939              \\
		\Checkmark &\Checkmark & \Checkmark &  &  & \Checkmark & 0.922   & 0.947   & 0.942  & 0.941              \\ \hline
	\end{tabular}}
\end{table}

\begin{figure*}[h!]
    \makeatletter
    \renewcommand{\@thesubfigure}{\hskip\subfiglabelskip}
    \makeatother
	\centering
	\subfigure[~~~~~~(a) ACC (YALE)]{
		\includegraphics[width=1.8 in]{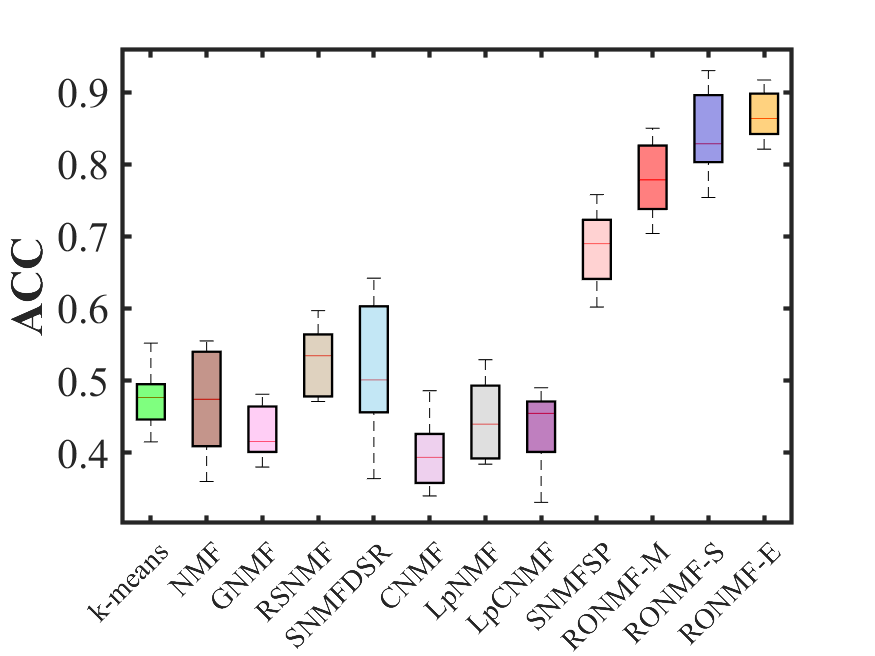}\hspace{-5mm}}
                \subfigure[~~~~~~(b) F1 (YALE)]{
		\includegraphics[width=1.8 in]{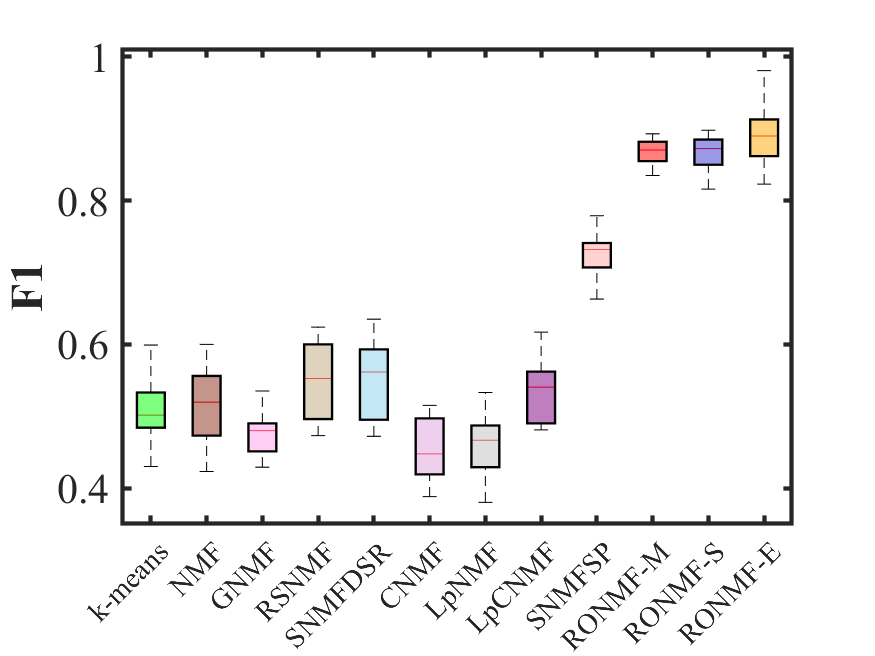}\hspace{-5mm}}
                \subfigure[~~~~~~(c) NMI (YALE)]{
		\includegraphics[width=1.8 in]{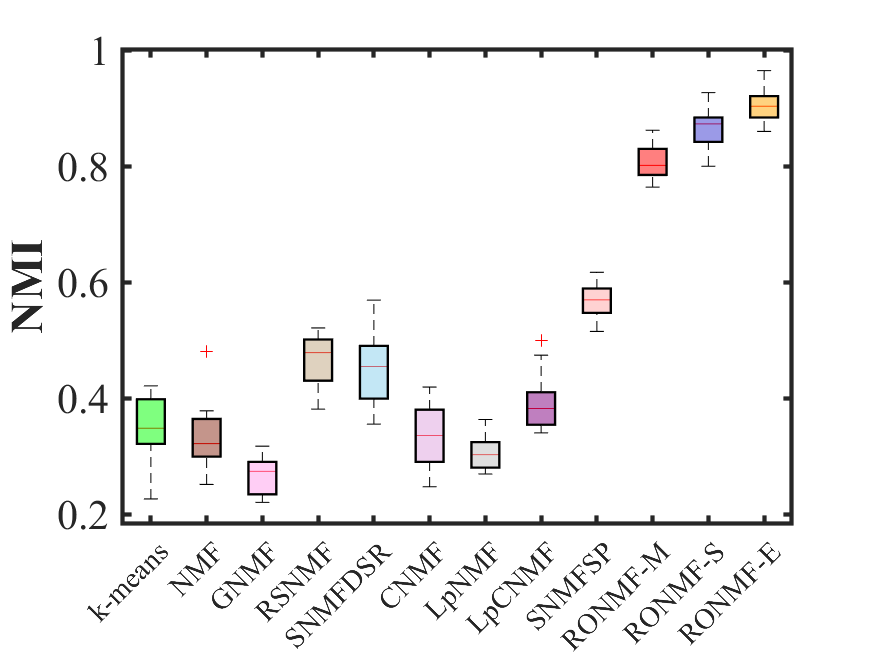}\hspace{-5mm}}
                \subfigure[~~~~~~(d) PUR (YALE)]{
		\includegraphics[width=1.8 in]{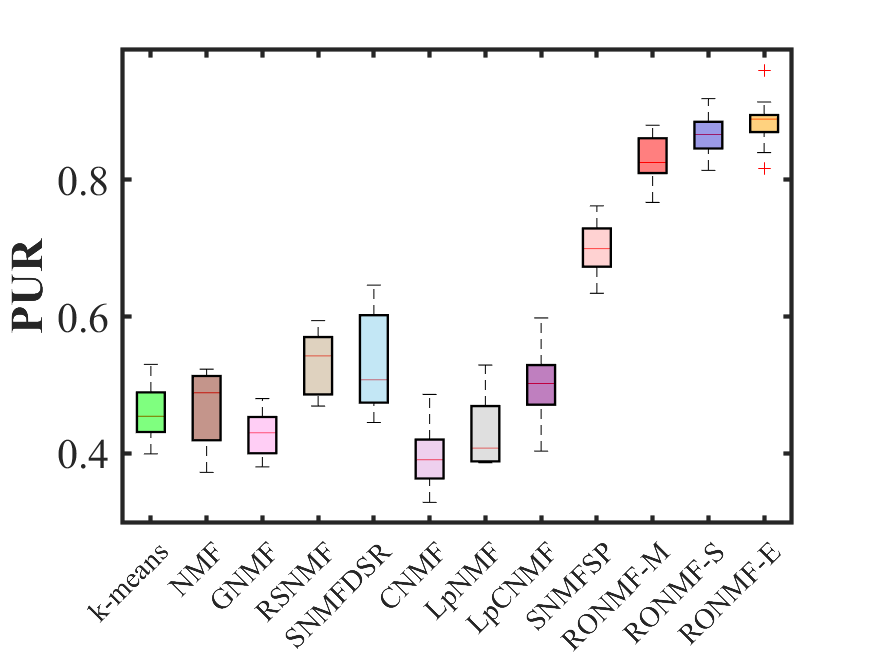}\hspace{-5mm}}
	\subfigure[~~~~~~(e) ACC (COIL20)]{
		\includegraphics[width=1.8 in]{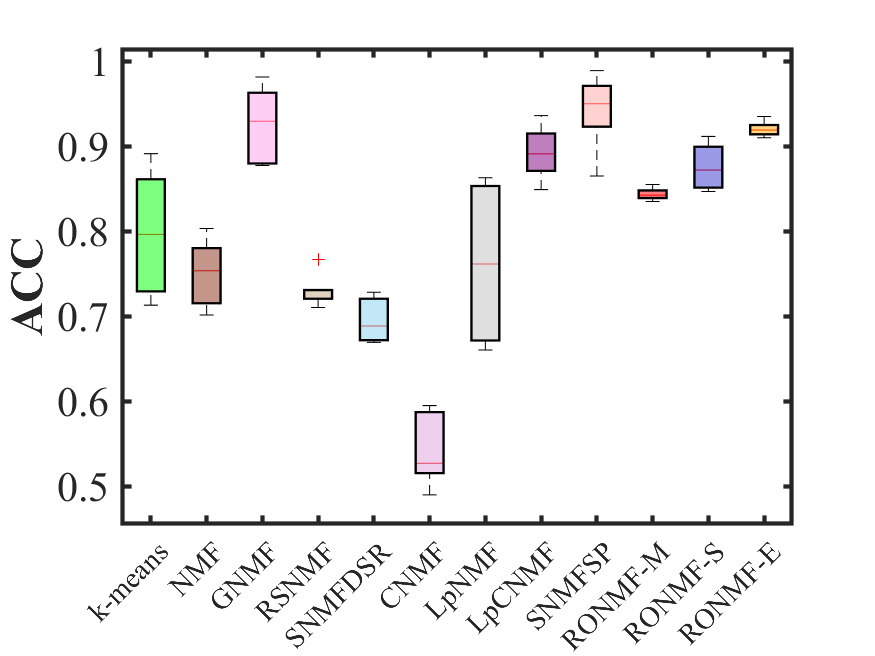}\hspace{-5mm}}
	\subfigure[~~~~~~(f) F1 (COIL20)]{
		\includegraphics[width=1.8 in]{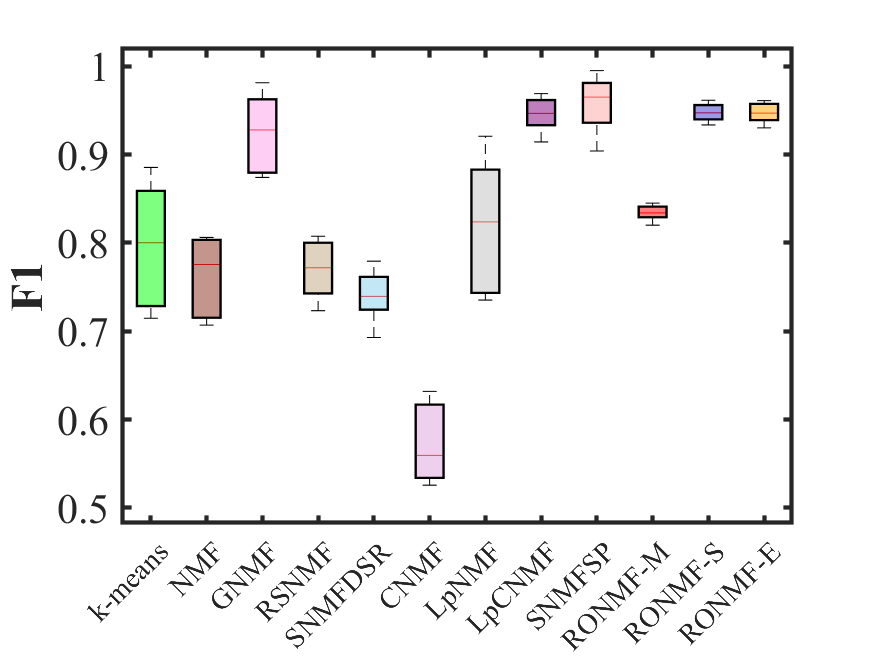}\hspace{-5mm}}
	\subfigure[~~~~~~(g) NMI (COIL20)]{
		\includegraphics[width=1.8 in]{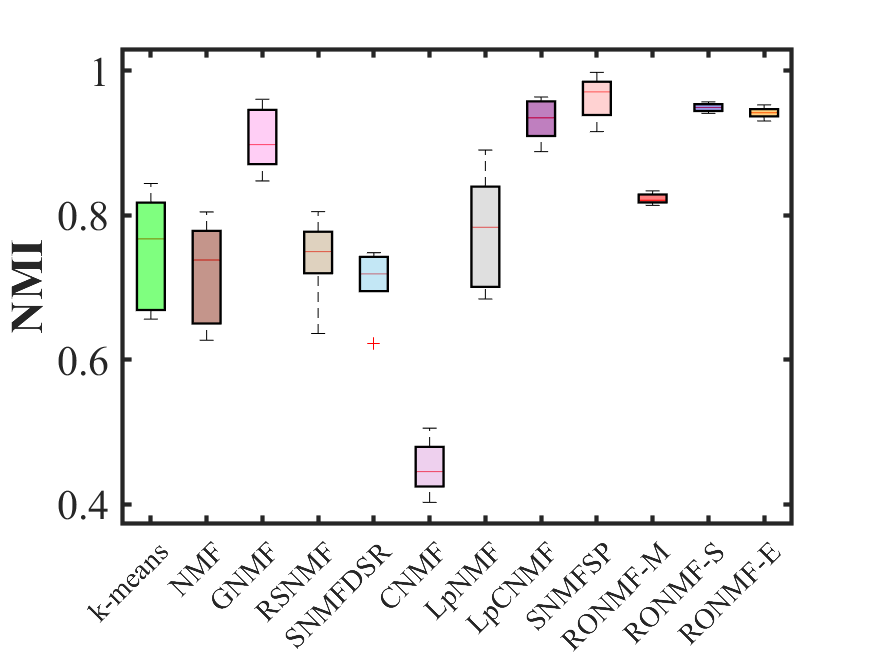}\hspace{-5mm}}
	\subfigure[~~~~~~(h) PUR (COIL20)]{
		\includegraphics[width=1.8 in]{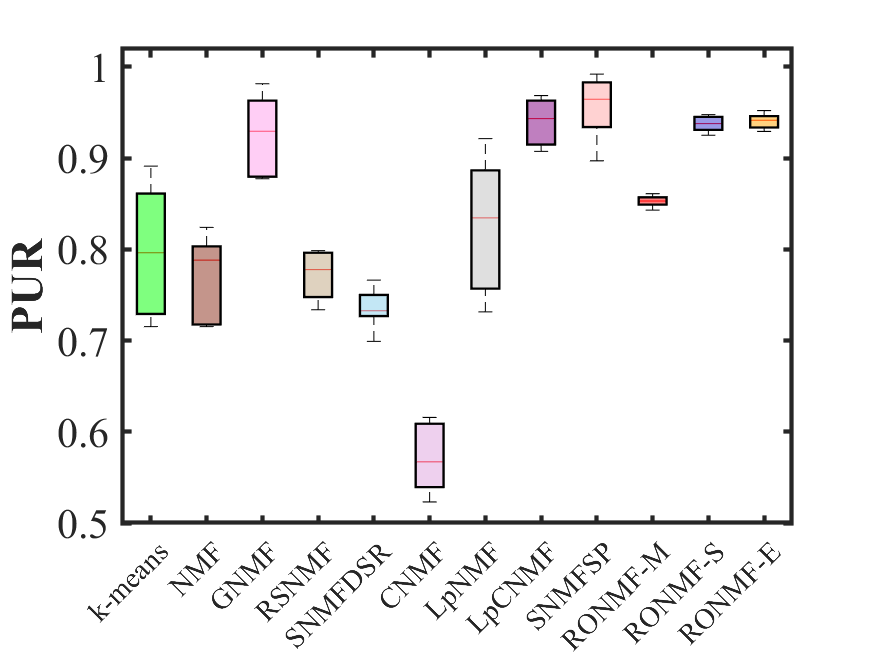}\hspace{-5mm}}
	\subfigure[~~~~~~(i) ACC (USPS)]{
		\includegraphics[width=1.8 in]{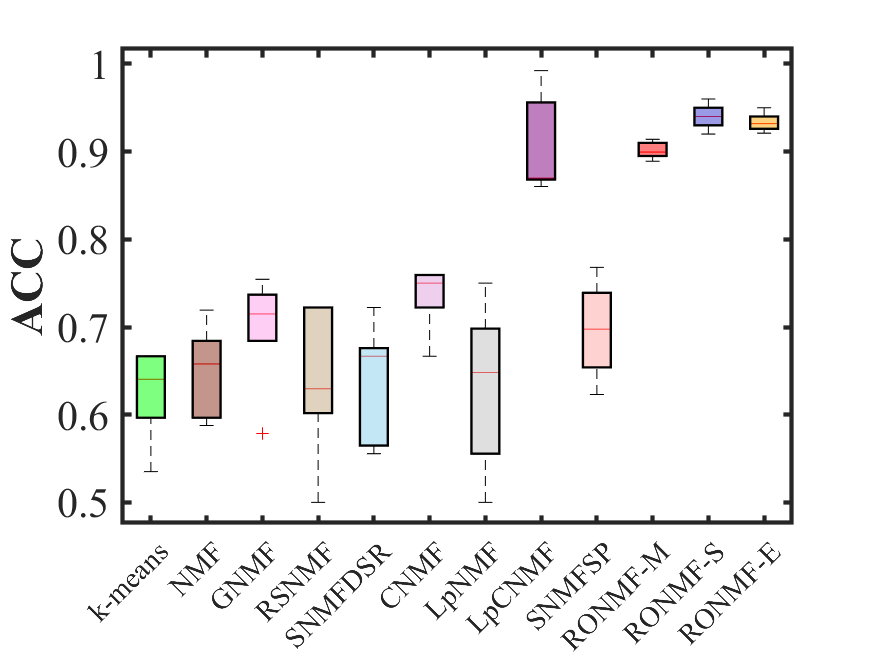}\hspace{-5mm}}
	\subfigure[~~~~~~(j) F1 (USPS)]{
		\includegraphics[width=1.8 in]{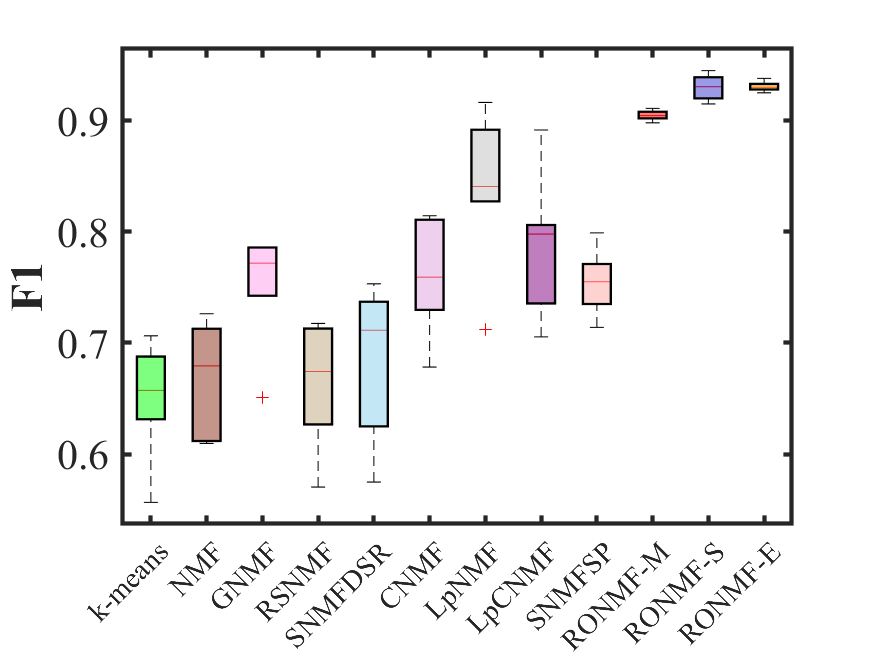}\hspace{-5mm}}
	\subfigure[~~~~~~(k) NMI (USPS)]{
		\includegraphics[width=1.8 in]{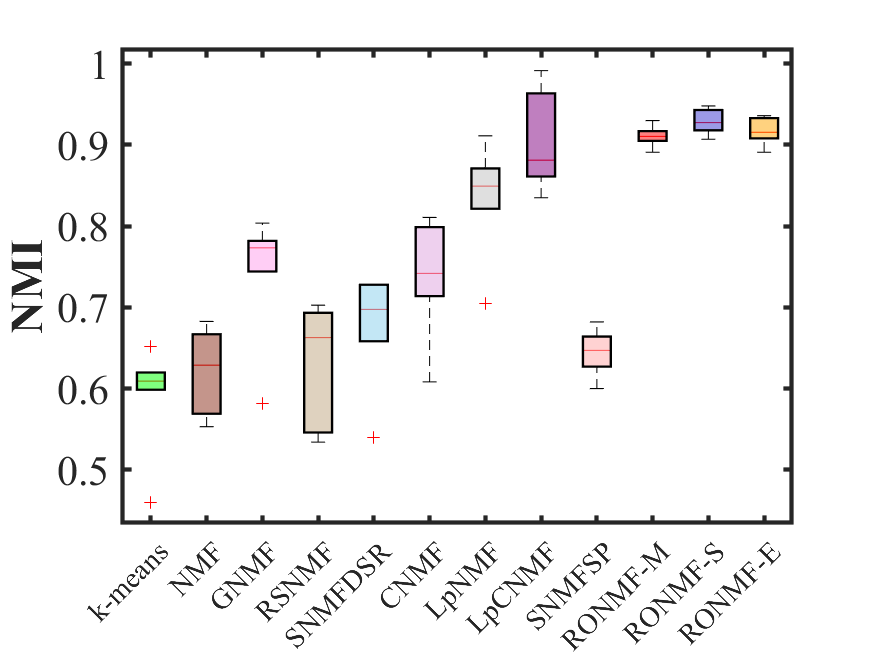}\hspace{-5mm}}
	\subfigure[~~~~~~(l) PUR (USPS)]{
		\includegraphics[width=1.8 in]{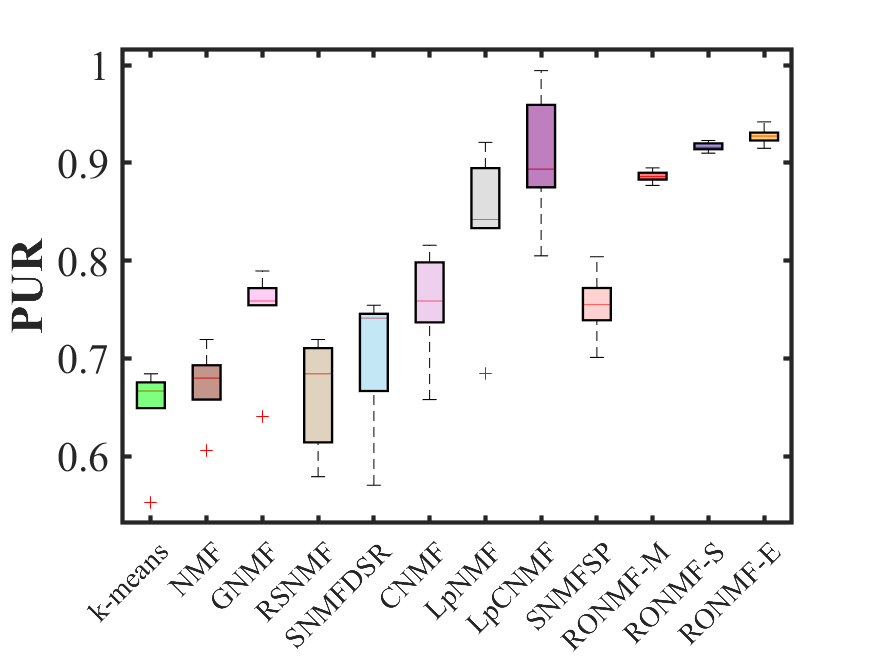}\hspace{-5mm}}
	\caption{Box plots from 10 runs  on  YALE, COIL20, and USPS  when $k=6$.}\label{box}
	\label{f3}
\end{figure*}

\begin{figure*}[ht!]
	\makeatletter
	\renewcommand{\@thesubfigure}{\hskip\subfiglabelskip}
	\makeatother
	\centering
	\subfigure[(a) k-means]{\label{tsne1}
		\includegraphics[width=1.5 in]{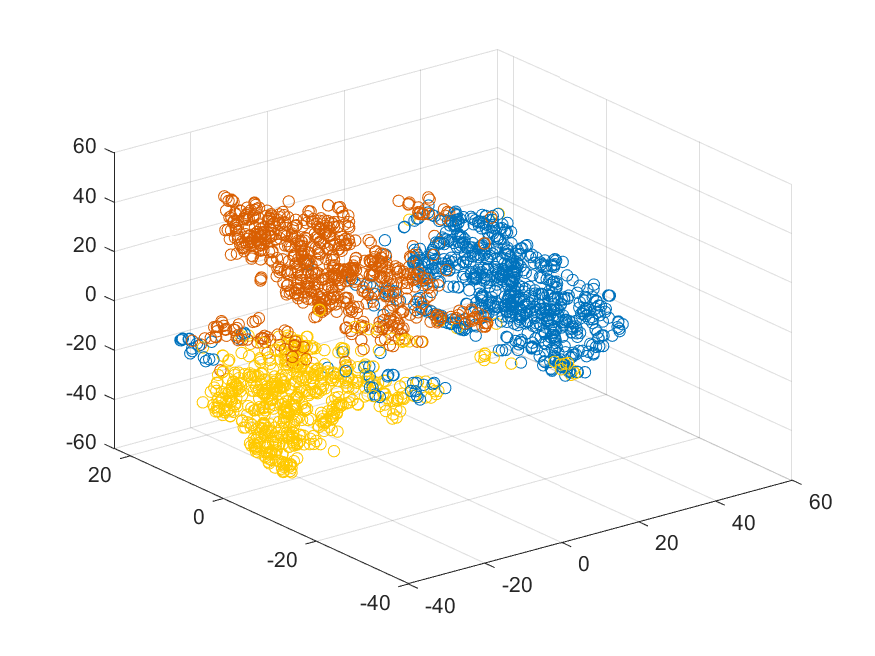}}
	\subfigure[(b) NMF]{\label{tsne2}
		\includegraphics[width=1.5 in]{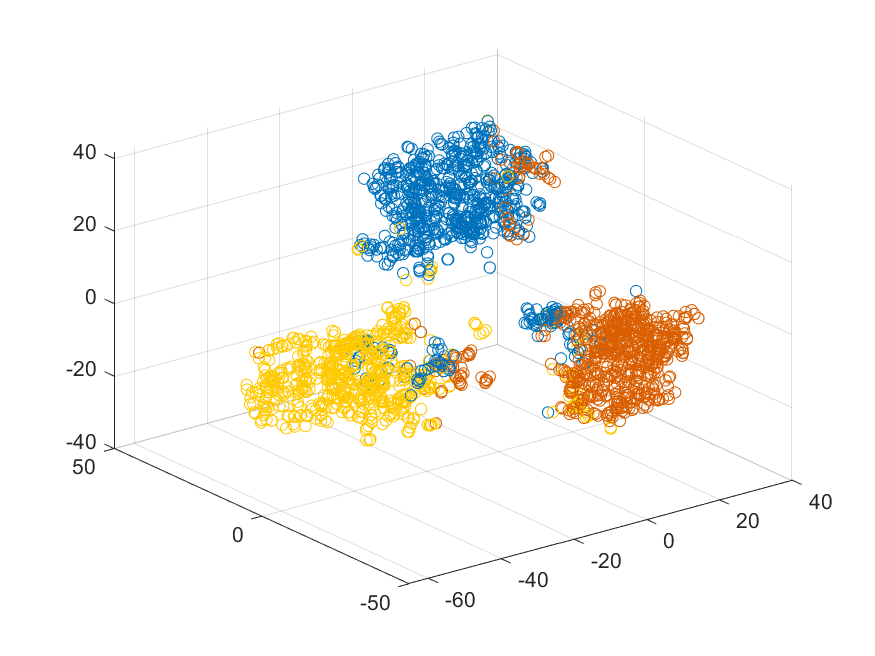}}
	\subfigure[(c) GNMF]{\label{tsne3}
		\includegraphics[width=1.5 in]{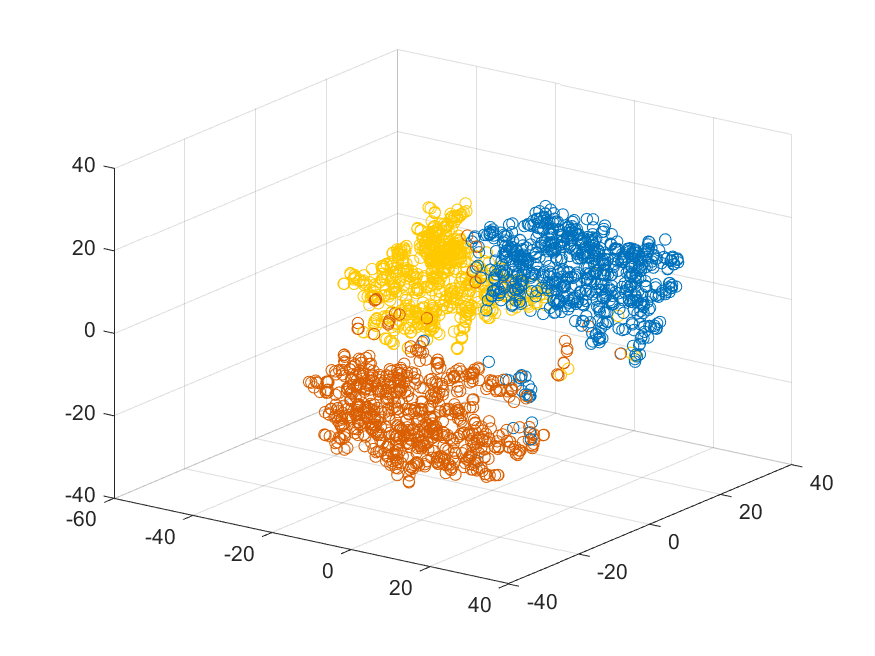}}
	\subfigure[(d) RSNMF]{\label{tsne4}
		\includegraphics[width=1.5 in]{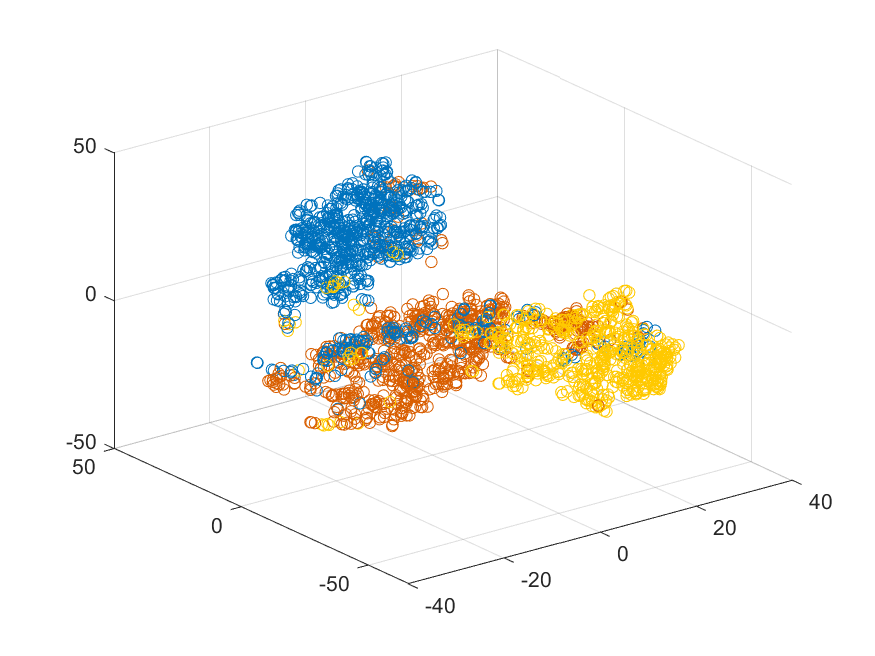}}
	\subfigure[(e) SNMFDSR]{\label{tsne5}
		\includegraphics[width=1.5 in]{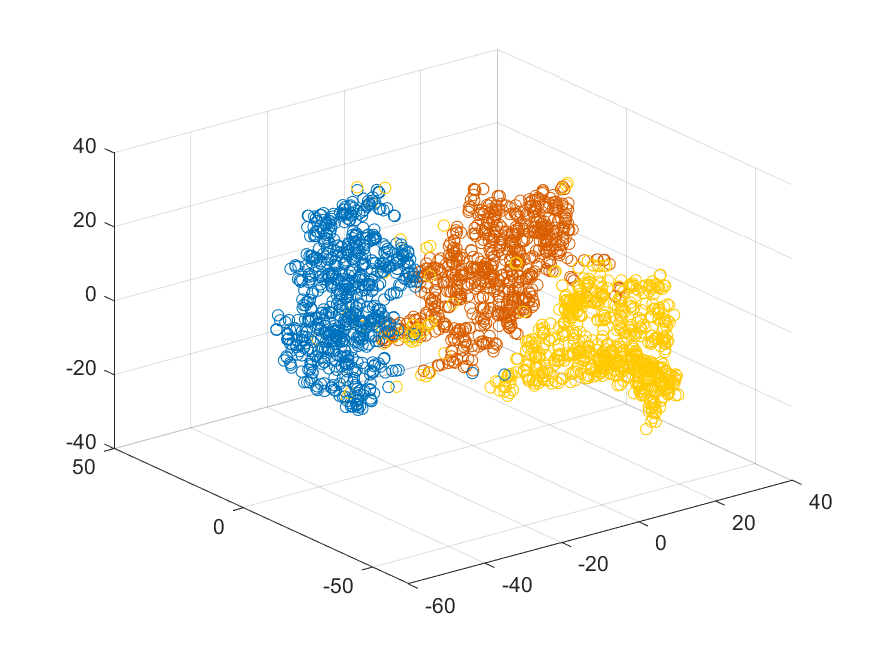}}
	\subfigure[(f) CNMF]{\label{tsne7}
		\includegraphics[width=1.5 in]{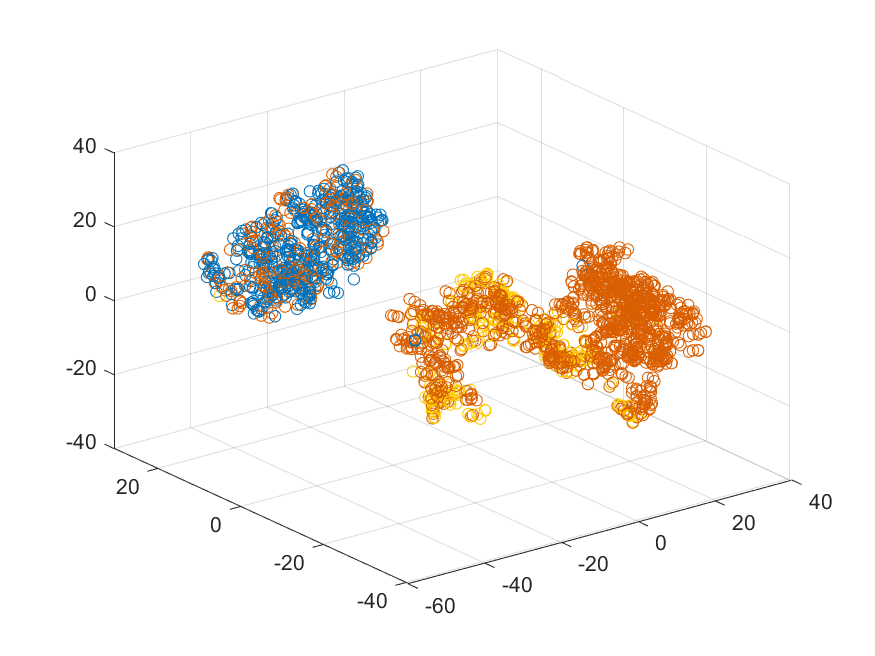}}
	\subfigure[(g) LpNMF]{\label{tsne8}
		\includegraphics[width=1.5 in]{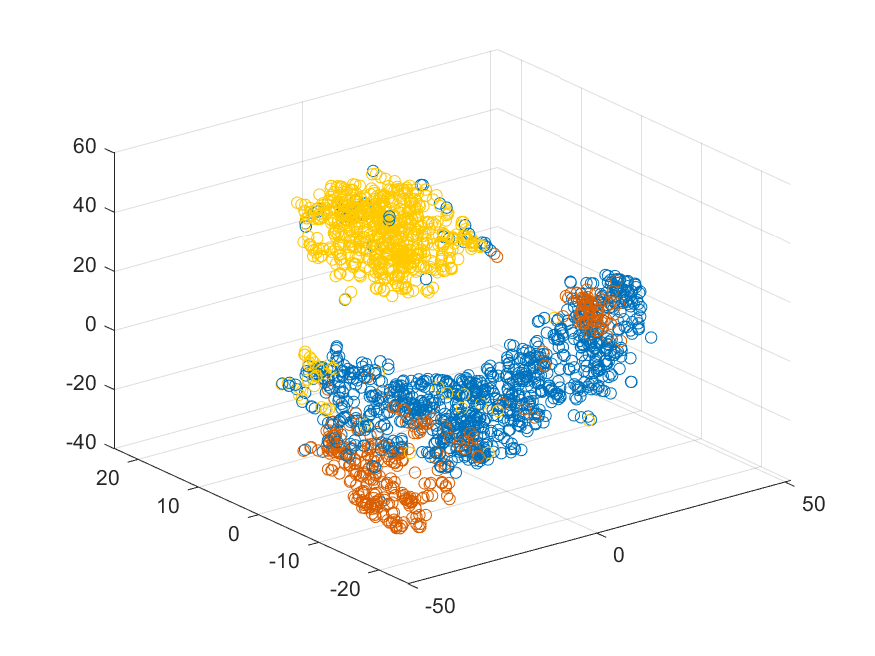}}
	\subfigure[(h) LpCNMF]{\label{tsne9}
		\includegraphics[width=1.5 in]{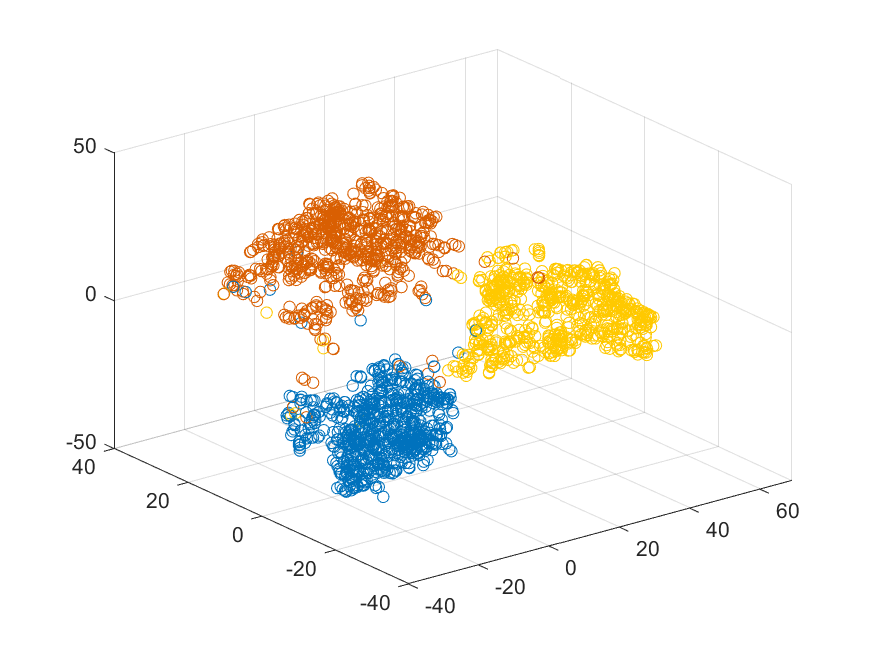}}
	\subfigure[(i) SNMFSP]{\label{tsn10}
		\includegraphics[width=1.5 in]{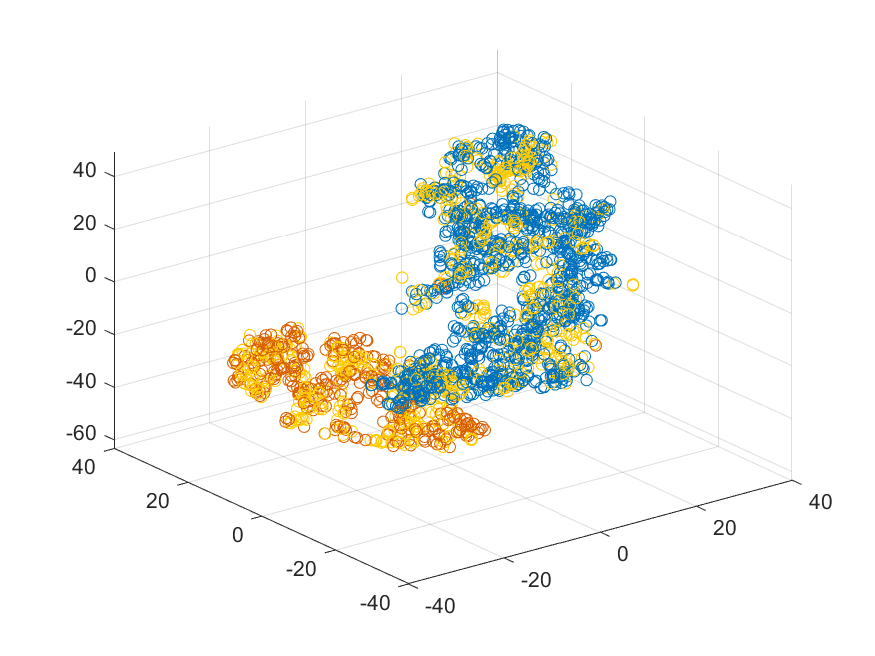}}
	\subfigure[(g) RONMF-M]{\label{tsne11}
		\includegraphics[width=1.5 in]{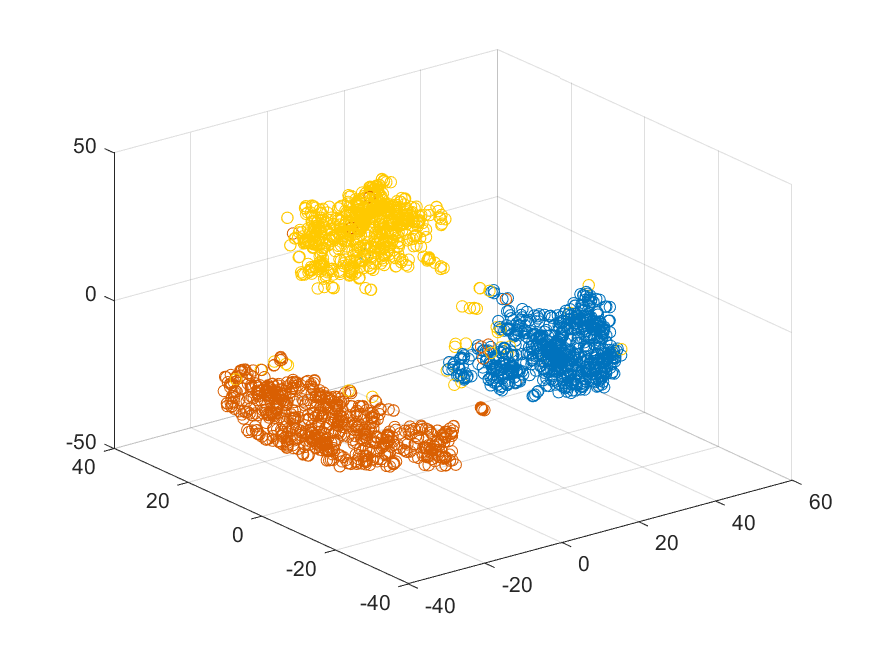}}
	\subfigure[(k) RONMF-S]{\label{tsne12}
		\includegraphics[width=1.5 in]{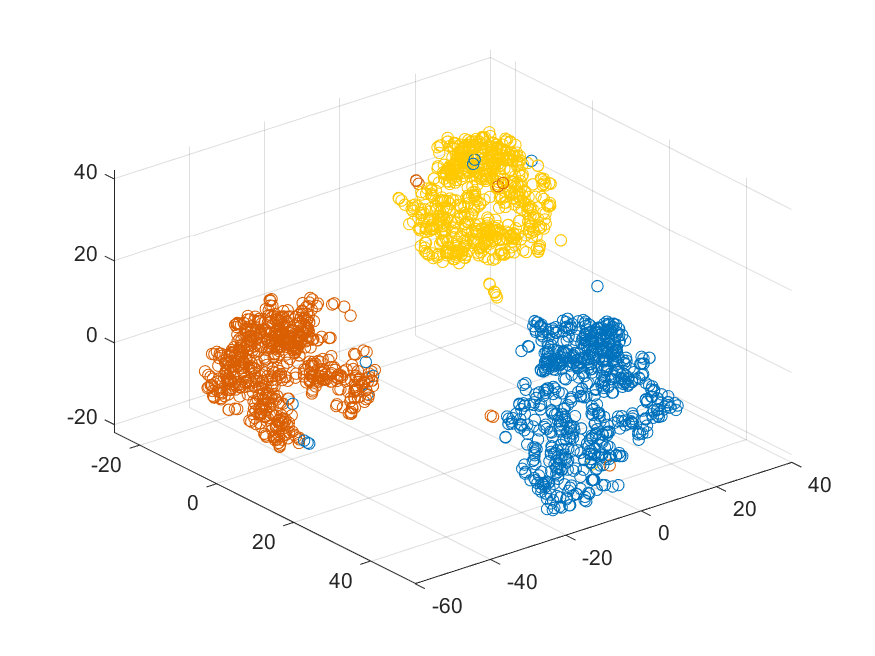}}
	\subfigure[(l) RONMF-E]{\label{tsne13}
		\includegraphics[width=1.5 in]{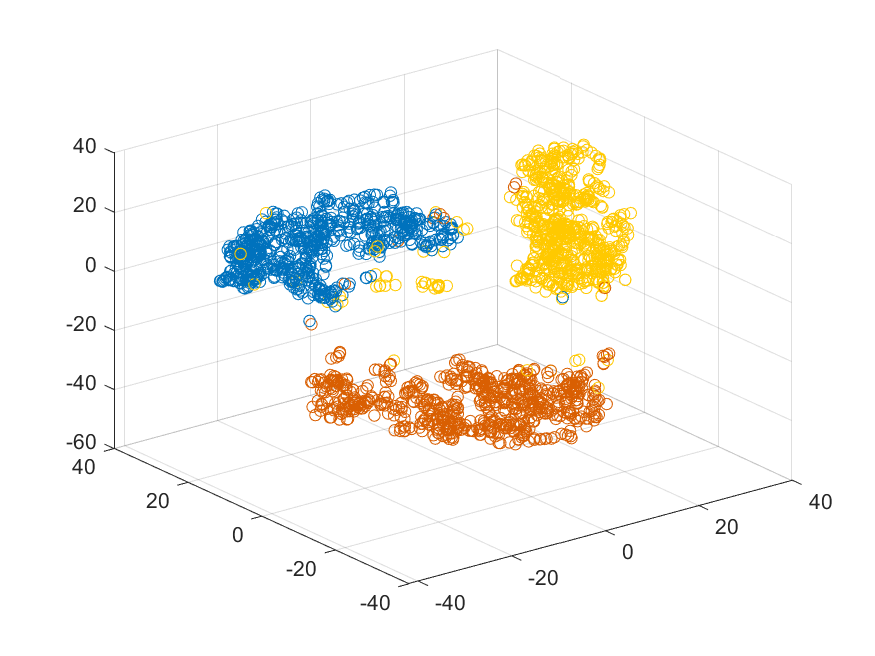}}
	\subfigure[]{\label{tsne}
		\includegraphics[width=1.6 in]{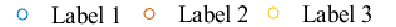}}
	\vskip-0.5cm
	\caption{Visulization of compared methods on USPS.}
	\label{tsne}
\end{figure*}

\subsection{Experimental Results}

As previously discussed, $\phi$ can choose MCP, SCAD, or ETP, as shown in Table~\ref{nonconvex}, corresponding to the methods RONMF-M, RONMF-S, and RONMF-E, respectively.
A series of experimental assessments are conducted to evaluate the effectiveness of these penalties.

Table \ref{tabclean1} presents a comparison of ACC, F1, NMI, and PUR results for all methods on the Type-\uppercase\expandafter{\romannumeral1} datasets.
The best results are highlighted in \red{red}, while the second-best results are highlighted in \blue{blue}.
Compared with LpCNMF, our proposed RONMF with structured non-convex functions demonstrates a significantly improved ability to extract face features. In particular, RONMF-E exhibits the most substantial improvements, improving ACC by 0.191 on YALE.
Table \ref{tabclean2} summarizes the clustering results of all methods on the Type-\uppercase\expandafter{\romannumeral2} datasets.
In comparison to GNMF, which also employs graph Laplacian regularization, RONMF effectively enhances the clustering accuracy by utilizing a small amount of labeled information.
When SCAD is applied to RONMF, it can be observed that the NMI is improved by 0.054 and the PUR is improved by 0.038.
Table \ref{tabclean3} presents the clustering results for the Type-\uppercase\expandafter{\romannumeral3} datasets, with RONMF demonstrating exceptional performance, particularly in F1 and NMI, outperforming other methods.
When $\phi$ is set to ETP, RONMF achieves an ACC of 0.933 and an NMI of 0.930 on MNIST, showing a relatively better improvement over SNMFSP.

Table \ref{tabclean4} presents the clustering results for the Type-\uppercase\expandafter{\romannumeral4} datasets.
The proposed RONMF algorithm has achieved relatively better performance on these two large datasets.
On the CAL101 dataset, all three penalty functions perform well, which demonstrates the effectiveness of the proposed $||\cdot||_{2, \phi}$ non-convex penalty framework.

Furthermore, Figures \ref{zxtclean1} and \ref{zxtclean2} show the clustering results when $k$ ranges from 2 to 10. It can be seen that the performance of RONMF is satisfactory compared with others, which verifies the effectiveness of our proposed method.

\subsection{Robustness Verification}
In order to evaluate the robustness of the proposed RONMF, we conduct experiments on YALE, COIL20, and USPS  with different Gaussian noise of 10\%, 30\%, 50\%, and 70\%. Figure \ref{noise} shows the evaluation results of all methods on these noisy datasets, respectively.
In most cases, the proposed RONMF demonstrates superior performance across all four evaluation metrics.
It can be concluded that our proposed method has strong anti-noise capabilities as the noise ratio varies from 10\% to 70\%.

Futhermore, Table \ref{tabnoise1} shows the clustering results after introducing Poisson noise and Salt pepper noise ($k=6$) to three datasets.
It can be found that under the same styles and levels of noise, the proposed method demonstrates its resistance to noise.
Especially on the COIL20 dataset under  Poisson noise, it has almost no negative impact on the RONMF method, which also confirms the robustness of the proposed method.

\subsection{Ablation Study}

In Table \ref{ablation}, all proposed penalties in Table \ref{nonconvex} are tested.
According to \eqref{rnmf-v}, the terms $\lambda \textrm{Tr}(A^\top LA)$ represents the image regularization component and $\mu \textrm{Tr}((A-Y)^\top S (A-Y))$ represents the label propagation component, which are set to 0 in this context to eliminate their influence on the model, respectively.
In the attempt to remove their influence, it can be observed a substantial reduction in the performance of all RONMF methods.
This also indicates that the roles of these two terms are different and imposing different parameters is necessary.
In addition, adding adding orthogonal constraints can improve the clustering performance, which shows the significance of feature selection.
It is noted that different non-convex functions have different effects, and they should be selected according to the data distribution in practical problems.

Therefore, we conclude that our proposed method can leverage the advantages of graph regularization, label propagation, orthogonal constraints, and  non-convex functions to enhance the robustness.

\subsection{Discussion}

This section provides a detailed discussion of the characteristics of our proposed method, encompassing parameter selection, statistical analysis, stability analysis, and visualization analysis.

\subsubsection{Parameter Selection}
For our proposed RONMF, the label proportion parameter \( p \) is fixed at 0.3,  while $\epsilon_1$ and $\epsilon_2$ are set to $10^{-4}$.
The parameters $\mu$ and $\lambda$ are the primary focus, with values tested as $\{10^{-3}, 10^{-2}, 10^{-1}, 1, 10^{1}, 10^{2}, 10^{3}\}$.
Cross-validation is employed to determine the optimal parameters.

The optimal values identified are $\lambda = 1000$ and $\mu = 1$.
These parameters are selected due to the potential of high values leading to overfitting.
It is observed from Figure \ref{heatmap} that the ACC and NMI of RONMF are relatively low when $\lambda$ is small, while all four indicators increase with the increment of $\lambda$.
In this end, $\lambda = 1000$ and $\mu = 1$ are chosen.
These results confirm the effectiveness of the regularized label propagation implemented by the proposed method in mitigating the significant label dependencies typically required for supervised or semi-supervised methods.

\subsubsection{Statistical Analysis}

Here, we choose the critical difference (CD) value as a metric for the post-hoc Nemenyi test to determine whether there are differences among these methods. The test results for ACC are illustrated in Figure \ref{post}. It can be observed that the proposed RONMF shows statistically significant differences compared to k-means, NMF, RSNMF, LpNMF, CNMF, and CDCF. However, there are no significant performance differences when compared to LpCNMF and SNMFSP.

\subsubsection{Stability Analysis}

In this study, box plots are used to compare the performance distribution of the proposed RONMF  with nine other methods on three datasets, YALE, COIL20, and USPS. This experiment records the experimental results of all methods in 10 tests. Figure \ref{box} reflects the results.
It can be seen that in most cases, the proposed method shows the best performance, which is characterized by smaller boxes, higher medians, and no outliers.
In fact, smaller boxes indicate smaller data fluctuations in each experiment, while higher medians indicate better average clustering performance. It further emphasizes the robustness and effectiveness of our proposed RONMF.

\subsubsection{Visualization Analysis}

This section utilizes the t-distributed stochastic neighbor embedding (t-SNE) \cite{van2008visualizing} to visualize the clustering results on USPS with three classes, as depicted in Figure \ref{tsne}.
This figure reveals how different methods handle inter-class and intra-class distances, offering insights into their clustering performance. It can be found that most methods such as LpCNMF, LpNMF, and SNMFSP show tighter intra-class clustering, reflecting their ability to preserve relative similarities among data points in high-dimensional space.
Meanwhile, the proposed RONMF variants, particularly RONMF-M and RONMF-E, exhibit better inter-class separability, highlighting their advantages in distinguishing different categories.


\section{Conclusion}\label{sec5}

In this paper, we have proposed a novel robust orthogonal NMF (RONMF) method by focusing on sparsity and robustness. Compared with existing NMF, the proposed RONMF has three differences. First, non-convex optimization is introduced to adapt to different data distributions. Second, orthogonal constraints are incorporated to effectively explore the inherent geometric structure in the data. Third, two regularization terms are considered to effectively preserve the original information of the image data while promoting the propagation of labels to similar data points. Comprehensive experiments have been conducted on different image databases, and the results show that the above modifications significantly improve the performance of traditional NMF methods. This convinces the advantages of our proposed method in image clustering.

In addition, we have noticed that in some cases, the improvement is not obvious. Therefore, integrating deep neural networks may provide a valuable opportunity for further enhancing performance.

%
%

\section*{Acknowledgments}
This work was supported in part by the National Natural Science Foundation of China under Grants 62204044 and 12371306, and in part by the State Key Laboratory of Integrated Chips and Systems under Grant SKLICS-K202302.

\bibliographystyle{elsarticle-num-names}

\bibliography{cas-refs}

\end{document}